\documentclass[lettersize,journal]{IEEEtran}

\usepackage{amsmath,amsfonts,amssymb}
\usepackage{algorithmic}
\usepackage{algorithm}
\usepackage{array}
\usepackage[caption=false,font=normalsize,labelfont=sf,textfont=sf]{subfig}
\usepackage{textcomp}
\usepackage{stfloats}
\usepackage{url}
\usepackage{verbatim}
\usepackage{graphicx}
\usepackage{cite}
\usepackage{booktabs}
\usepackage{multirow}
\usepackage{xcolor}
\usepackage{tabularx}
\usepackage{makecell}
\usepackage{cuted}
\usepackage{multirow}

\begin{document}

\title{Action QFormer: Structured Representation Shaping under Action Supervision in Vision-Language-Action Models}

\author{
Yufeng Ji$^{1,2,3,*}$,
Wenhao Tang$^{4}$,
Haoyi Niu$^{5}$,
Koushil Sreenath$^{5}$,
Yi Wu$^{4}$,
Zhongyu Li$^{3,2,*}$
\thanks{
$^{1}$Shanghai Qizhi Institute;
$^{2}$The Chinese University of Hong Kong;
$^{3}$Hong Kong Embodied AI Lab;
$^{4}$Tsinghua University;
$^{5}$University of California, Berkeley.
$^{*}$Corresponding authors:
Yufeng Ji (\texttt{jiyf@sqz.ac.cn}) and
Zhongyu Li (\texttt{zhongyuli@cuhk.edu.hk}).
}
}


\IEEEaftertitletext{%
\vspace{-3.0em}
\begin{center}
    \includegraphics[
        width=0.96\textwidth,
        trim=0 0pt 0 0pt,
        clip
    ]{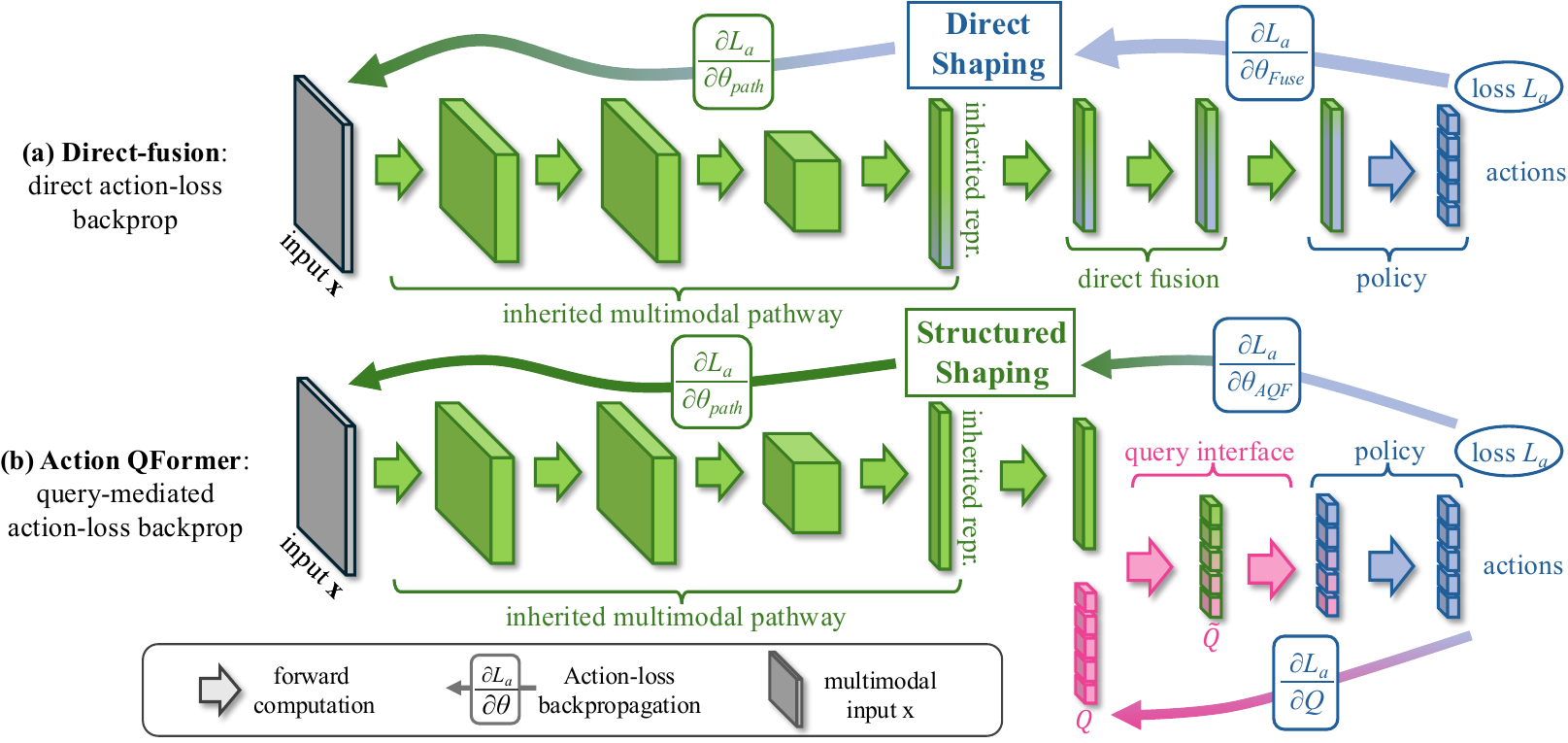}

    \vspace{1.0em}
    \refstepcounter{figure}
    \label{fig:main_figure}
    \parbox{1.0\textwidth}{%
    \footnotesize
    \linespread{0.92}\selectfont
    \textbf{Fig.~\thefigure.}
    \textbf{Direct versus structured action-loss shaping.}
    Both the direct-fusion baseline and Action QFormer use the same inherited multimodal pathway and downstream policy objective, but induce different action-loss gradient routes.
    \textbf{(a)} In direct shaping, inherited representations are fused into an action-facing representation, so action-loss gradients pass through the fusion interface and broadly act on inherited representations.
    \textbf{(b)} In structured shaping, Action QFormer inserts a trainable query interface before the policy head, giving action-loss signals an intermediate query-based route before they reach the inherited multimodal pathway.
    Gradients to $\theta_{\mathrm{path}}$, $\theta_{\mathrm{Fuse}}$, $\theta_{\mathrm{AQF}}$, and $Q$ denote action-loss updates to the inherited pathway, interface parameters, and query states.
    }
\end{center}
}
\maketitle

\begin{abstract}
Action supervision in vision-language-action (VLA) models is often treated as a downstream objective for learning action prediction.
In this paper, we study it instead as a force that shapes inherited multimodal representations.
We show that this shaping has a dual effect:
it is necessary for forming action-compatible representations, but when action supervision is applied too directly to the inherited multimodal pathway, it can also destabilize representations that support language-side processing and object grounding.
To address this tension, we introduce \textbf{Action QFormer}, a query-based action-facing interface that uses instruction-conditioned queries to reorganize inherited multimodal information into action-facing representations before downstream action generation.
In zero-shot sim-to-real navigation, Action QFormer improves average closed-loop task success from 18.8\% to 56.3\%, raises fixed-instruction action-generation correctness from 22.5\% to 75.5\%, and nearly eliminates out-of-distribution instruction generations.
Further analyses show that Action QFormer changes how action supervision shapes inherited multimodal representations, reducing broad upstream rewriting while preserving targeted and sometimes constructive action-supervised adaptation.
These results suggest that improving VLA performance requires not only stronger pretrained backbones, but also better ways of selecting and organizing inherited multimodal information while controlling how it is shaped under action supervision.
\end{abstract}


\vspace{-0.5em}

\section{Introduction}

Recent vision-language-action (VLA) models commonly build action generation on top of pretrained vision-language backbones~\cite{rt1,openvla,pi0}, enabling rapid progress by reusing rich multimodal representations for embodied control.
However, this design also exposes a structural mismatch:
the inherited multimodal pathway is asked to support both language-side processing and action-side control.
For language-side processing, the pathway must preserve rich semantic and grounding structure for instruction understanding and generation.
For action generation, it must abstract behaviorally consequential factors into stable and robust control signals.
A representation that is useful for vision-language understanding may therefore still be poorly organized for action, not because it lacks semantic information, but because it does not expose that information in an action-compatible form.

This mismatch makes action supervision not merely a downstream training signal, but a \emph{representation-shaping force}.
During action finetuning, action-loss gradients can flow through the inherited multimodal pathway and reshape it toward action-compatible organization.
The action interface is therefore not a passive readout of inherited multimodal features, but the point where action-loss gradients reshape---and can also disrupt---representations inherited from vision-language pretraining.

This gives rise to the dual effect we study:
action-supervised shaping is necessary for forming action-compatible representations, but when applied too directly to the inherited multimodal pathway, it can also destabilize representations that support language-side processing and object grounding.

This motivates the central question of the paper:
\emph{during action finetuning, how can VLA systems allow action supervision to reshape multimodal representations inherited from vision-language pretraining without destabilizing them?}
We address this question through \textbf{structured representation shaping under action supervision}: rather than letting action-loss gradients act directly on the inherited multimodal pathway, we introduce an intermediate module that mediates where action-supervised shaping occurs.

We instantiate this view with \textbf{Action QFormer}, a query-based action-facing interface inserted between the pretrained multimodal backbone and the downstream policy head.
Action QFormer uses learnable queries to reorganize inherited multimodal information before action prediction.
At the same time, this query interface also changes the backward route of action supervision, giving action-loss gradients an intermediate adaptation path before they propagate upstream to reshape inherited multimodal representations.

We study Action QFormer in zero-shot sim-to-real navigation, a challenging setting that places high demands on representation stability.
In our closed-loop perception-to-instruction-to-action pipeline, the model must generate stable intermediate instructions and execute actions sequentially under changing real-scene observations.
This makes representation instability behaviorally explicit: instruction-generation instability, weak object grounding, or directional-control errors can accumulate into trajectory divergence and final task failure.
By dissecting the full pipeline, we find that baseline degradation appears at two levels:
unstable instruction generation under real-scene visual shift, and weaker directional control and object grounding during instruction-conditioned execution.
Action QFormer improves both levels, maintaining more stable intermediate instructions in closed-loop deployment while producing stronger directional control and object-grounded execution.

Representation-level analyses further show that these behavioral gains are accompanied by more controlled action-supervised adaptation:
Action QFormer strengthens action-facing directional distinctions while reducing broad upstream rewriting and stabilizing instruction-to-visual attention.

Together, these results support a mechanism-level thesis:
\emph{action supervision reshapes multimodal representations inherited from pretrained backbones; this shaping is necessary for forming action-compatible representations, but direct shaping can cause broad upstream disruption.}
Query-based action-facing interfaces can mediate this interaction by expressing part of the action-loss update through learnable queries, optimizing the organization of multimodal information for action prediction before the update propagates back as upstream shaping.
This reframes action-interface design from the perspective of representation shaping under action supervision:
the challenge is not only to build richer visual backbones or stronger action decoders, but to control how action supervision interacts with inherited multimodal representations.

Our contributions are as follows.
First, we formulate \textbf{structured representation shaping under action supervision} as a mechanism-level problem in VLA, identifying a central tension between forming action-compatible representations and preserving multimodal representations inherited from pretraining.
Second, we introduce \textbf{Action QFormer}, a query-based action-facing interface that mediates action-supervised shaping through instruction-conditioned queries, reorganizing inherited multimodal information into action-facing representations before downstream action prediction.
Third, using zero-shot sim-to-real robot navigation as a challenging test scenario for representation stability, we show that Action QFormer improves closed-loop task success and fixed-instruction action-generation correctness, while nearly eliminating out-of-distribution instruction generations.
Finally, through mechanistic analyses, we show that Action QFormer changes how action supervision reshapes inherited multimodal representations, making adaptation more controlled, localized, and aligned with downstream action generation.

\section{Related Work}


\vspace{0.5em}
\noindent\textbf{Vision-Language-Action Models and Action Interfaces.}
Recent VLA models extend pretrained vision-language or multimodal backbones to embodied control through different action interfaces.
Representative systems include transformer policies trained on large-scale robot data in RT-1~\cite{rt1}, action-as-text token prediction in RT-2~\cite{rt2}, discrete action binning in OpenVLA~\cite{openvla}, flow-matching continuous action generation in $\pi_0$~\cite{pi0}, and discrete action tokens for scalable pretraining in $\pi_{0.5}$~\cite{pi05}.
Another line of work focuses on the action interface itself:
frequency-space action sequence tokenization in FAST~\cite{fast}, vector-quantized action tokenization in VQ-VLA~\cite{vqvla}, latent actions learned from unlabeled videos in LAPA~\cite{lapa}, task-centric latent actions for cross-embodiment transfer in UniVLA~\cite{univla}, and continuous rotational latent actions in RotVLA~\cite{rotvla}.
These approaches improve VLA scalability, action decoding, and cross-embodiment generalization by changing how actions are represented or generated, but leave how action supervision reshapes inherited multimodal representations largely unexplored.
We address this gap with an action-facing interface that structures this shaping process.

\vspace{0.5em}
\noindent\textbf{Query-Based Interfaces and Multimodal Alignment.}
Multimodal alignment has been a central mechanism for connecting visual and language representations, from contrastive image-text alignment in CLIP~\cite{radford2021clip}, align-before-fuse representation learning in ALBEF~\cite{li2021albef}, and unified understanding-generation pretraining in BLIP~\cite{li2022blip}, to visual instruction tuning in LLaVA~\cite{liu2023llava}.
Another line of work introduces intermediate query or resampling interfaces to bridge pretrained encoders and downstream language modules, including latent bottleneck processing in Perceiver~\cite{perceiver}, visual resampling for few-shot multimodal learning in Flamingo~\cite{flamingo}, learnable visual queries for frozen image-to-language transfer in BLIP-2~\cite{li2023blip2}, and instruction-conditioned visual querying in InstructBLIP~\cite{dai2023instructblip}.
Recent robotics work further studies alignment toward action-relevant structure, including visual-encoder grounding alignment for spatially aware VLA models in VEGA~\cite{vega}, dynamics-aware flow-language-image pretraining in DynaFLIP~\cite{dynaflip}, human-robot aligned representation learning for VLA pretraining in HARP-VLA~\cite{harpvla}, latent-action representation alignment in LARA~\cite{lara}, and action-grounded representation alignment for World Action Models in AGRA~\cite{agra}.
Together, these works suggest that pretrained multimodal representations should be reused through interfaces designed to match the downstream objective; for VLA, we use learnable queries to reorganize inherited image and instruction representations for downstream action prediction.

\vspace{0.5em}
\noindent\textbf{Foundation Models in Robot Navigation.}
Classical and structured learning-based navigation methods rely on explicit planning, from graph-search planning in A*\cite{astar} and dynamic local collision avoidance\cite{dwa} to planning-guided visual navigation in ViKiNG~\cite{viking}, rather than fully end-to-end action learning.
More recent methods learn scalable visual and goal-conditioned policies, including cross-embodiment navigation in GNM~\cite{gnm}, foundation visual navigation in ViNT~\cite{vint}, and goal-masked diffusion policies in NoMaD~\cite{nomad}.
Another line brings pretrained foundation models into navigation, including model composition in LM-Nav~\cite{lmnav}, video-based VLM planning in NaVid~\cite{navid}, video-based VLA navigation in Uni-NaVid~\cite{uninavid}, navigation foundation modeling in NavFoM~\cite{navfom}, and future-observation prediction in Navigation World Models~\cite{nwm}.
Together, these methods highlight that navigation depends on robust language- or goal-conditioned visual grounding across changing scenes, making it a sensitive setting for studying representation instability.
We further stress this setting through zero-shot sim-to-real navigation, deriving training data from the Habitat~\cite{habitat} ObjectNav benchmark~\cite{objectnav} and evaluating behavior in real indoor scenes.

\section{Methodology}
\label{sec:methodology}

We present the methodology for studying action-supervised representation shaping in VLA models.
We first define the terminology used throughout the paper, then specify the problem setting, direct-fusion baseline, Action QFormer interface, representation-shaping mechanism, diagnostics, and training and inference pipeline.

\vspace{0.5em}
\noindent\textbf{Terminology.}
Throughout the paper, we use \emph{inherited multimodal representations} to refer to image-side and instruction-side representations produced by the pretrained multimodal backbone.
During action finetuning, action-loss gradients can reshape these representations; we refer to this process as \emph{action-supervised representation shaping}.
We use \emph{action-facing representation} to denote the representation that directly conditions the downstream policy head.
An \emph{action-facing interface} is an intermediate module that maps inherited image-side and instruction-side representations into an action-facing representation before trajectory prediction.
Under this terminology, Action QFormer implements \emph{structured representation shaping} by mediating action-supervised adaptation through a query-based action-facing interface before action-loss gradients propagate back to inherited representations.

\vspace{0.5em} 
\noindent\textbf{Problem setup.}
We consider a single-frame navigation setting to focus on how current visual and language information is organized into an action-compatible form.
At each timestep, navigation is modeled as an instruction-conditioned action prediction problem: the model first produces an intermediate instruction $s_t$, and then predicts an $H$-step action trajectory conditioned on it.
Each timestamp-level sample contains a current-frame image observation $I_t$, a task text $g_t$, a supervised navigation instruction $s_t$, and an $H$-step future action trajectory $A_t$ in the local robot frame, where $H=8$:
\begin{equation}
\mathcal{X}_t = (I_t, g_t, s_t, A_t),
\qquad
A_t = (a_t^1, \ldots, a_t^H).
\end{equation}
Each sample is formatted as a multimodal sequence $X_t$ and processed by the pretrained multimodal backbone:
\begin{equation*}
H_t = F_{\theta}(X_t),
\qquad
H_t \in \mathbb{R}^{N_t \times d},
\end{equation*}
where $N_t$ is the sequence length and $d$ is the hidden dimension.
We define the image-side and instruction-side representations by extracting hidden states from image-observation and supervised-instruction spans:
\begin{equation*}
H_t^{I} = \{H_{t,i}\}_{i \in \mathcal{I}_t},
\qquad
H_t^{S} = \{H_{t,j}\}_{j \in \mathcal{S}_t},
\end{equation*}
where the span indices $\mathcal{I}_t$ and $\mathcal{S}_t$ are determined by boundary markers in the multimodal sequence.
Here, $H_t^{I} \in \mathbb{R}^{N_I \times d}$ and
$H_t^{S} \in \mathbb{R}^{N_S \times d}$ denote image-side and instruction-side representations inherited from the multimodal backbone.

Each action step is represented by a local relative position and a heading vector:
\begin{equation}
\begin{aligned}
a_t^h &= \left(p_t^h,\mathbf{u}_t^h\right), \qquad h=1,\ldots,H, \\
p_t^h &= (x_t^h,y_t^h), \qquad
\mathbf{u}_t^h =
\left(
\mathrm{yaw}_{dx,t}^h,
\mathrm{yaw}_{dy,t}^h
\right)
\end{aligned}
\end{equation}
where $p_t^h$ denotes local relative position and $\mathbf{u}_t^h$ denotes heading direction.

\begin{figure}[!t]
    \centering
    \includegraphics[
        width=\linewidth
    ]{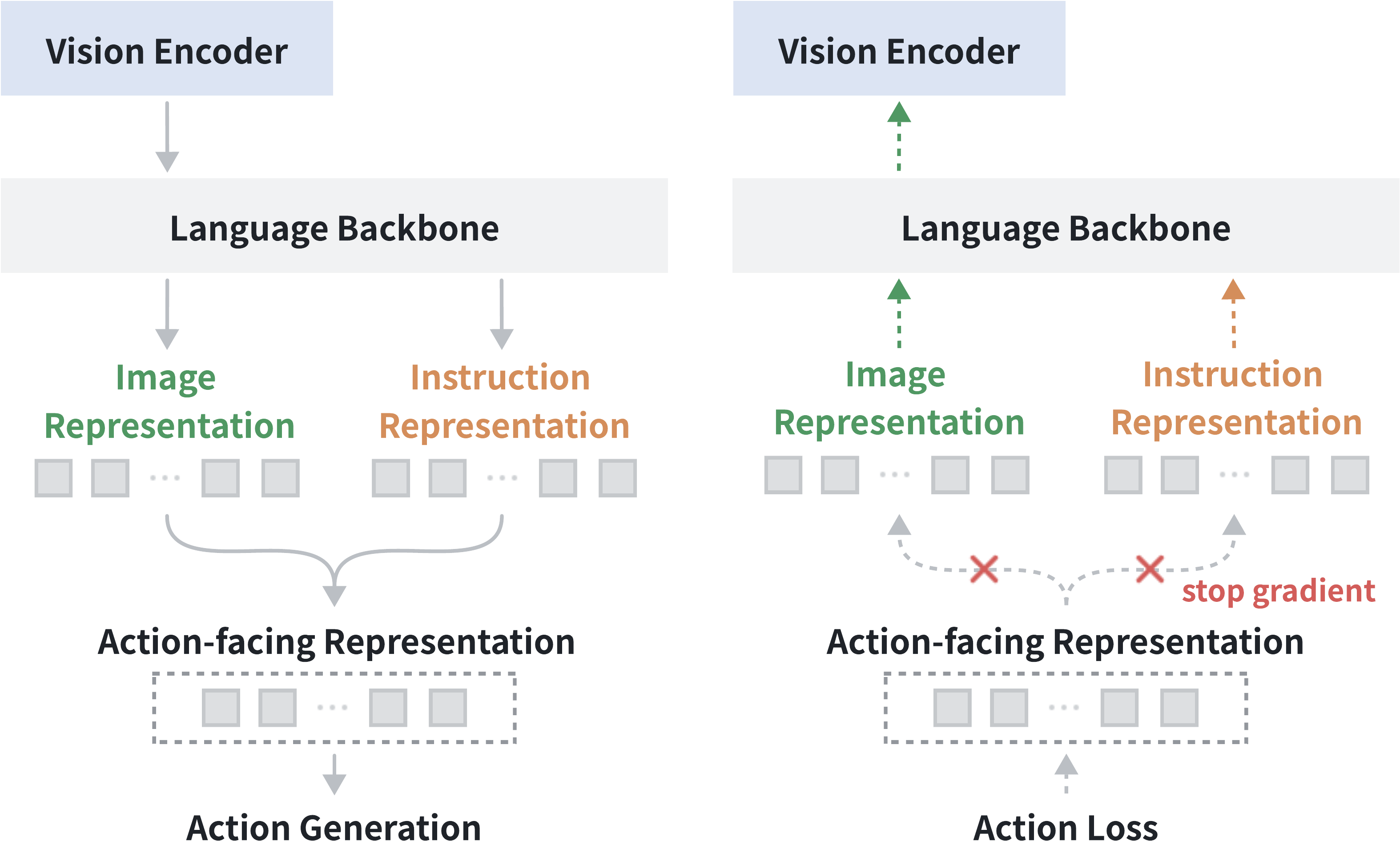}


    \begin{minipage}[t]{0.49\linewidth}
        \centering
        {\footnotesize (a) Direct-fusion baseline}
    \end{minipage}
    \hfill
    \begin{minipage}[t]{0.49\linewidth}
        \centering
        {\footnotesize (b) Selective gradient blocking}
    \end{minipage}
    
    \vspace{-0.5em}
    \caption{
    \textbf{Direct-fusion baseline as a diagnostic reference for action-facing interfaces.}
    \textbf{(a)} The baseline instantiates a common VLA interface pattern:
    image-side and instruction-side representations inherited from the multimodal backbone are directly fused into an action-facing representation for downstream policy prediction.
    \textbf{(b)} One representative stop-gradient configuration in baseline, where action-loss gradients are blocked before reaching the extracted image-side and instruction-side representations.
    }
    \label{fig:baseline_probe}
    \vspace{-1.0em}
\end{figure}

\vspace{0.5em}
\noindent\textbf{Direct-fusion baseline.} As shown in Fig.~\ref{fig:baseline_probe}(a), we first define a direct-fusion baseline to isolate a common VLA interface pattern:
inherited image-side and instruction-side representations are fused into an action-facing representation and directly exposed to action-loss optimization.
Given $H_t^I$ and $H_t^S$, the baseline implements this interface with self-attention:
\begin{equation*}
\begin{aligned}
Z_t^{\mathrm{Fuse}}
&=
\mathrm{SelfAttn}^{\mathrm{Fuse}}
\bigl(\mathrm{Concat}(H_t^I, H_t^S)\bigr), \\
z_t^{\mathrm{Fuse}}
&=
\mathrm{Pool}(Z_t^{\mathrm{Fuse}})
\equiv
\mathrm{Fuse}(H_t^I, H_t^S).
\end{aligned}
\end{equation*}

\begin{figure}[!t]
    \centering
    \includegraphics[
        width=0.85\linewidth
    ]{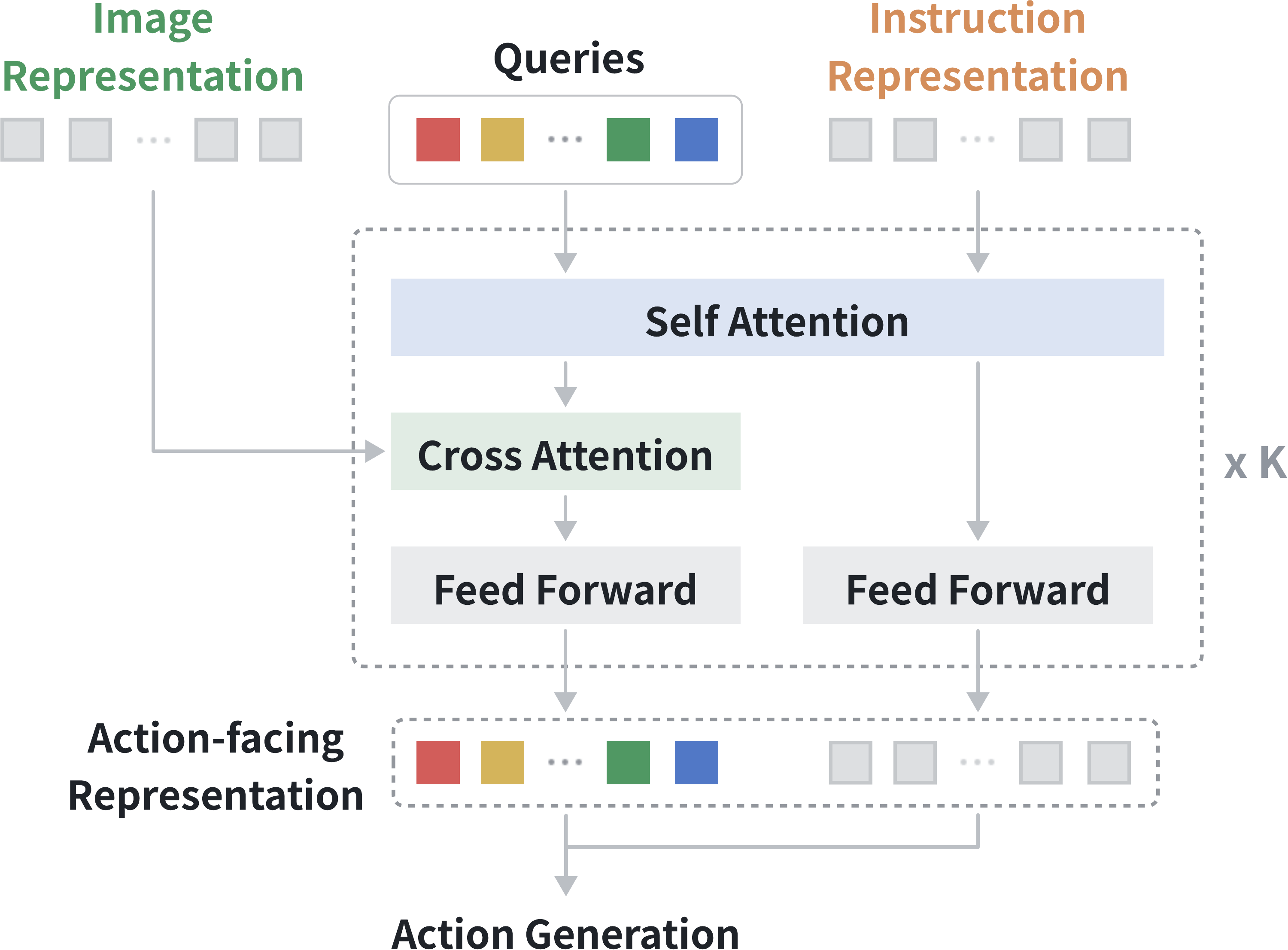}
    \vspace{-0.8em}
    \caption{
    \textbf{Action QFormer as a query-based action-facing interface.}
    Action QFormer conditions learnable queries on the instruction representation and uses the updated queries to selectively extract action-relevant visual information.
    The resulting query outputs carry instruction-conditioned visual information and are combined with the instruction-side representation to form action-facing representations for conditioning downstream action generation.
    }
    \label{fig:action_qformer_architecture}
    \vspace{-1.0em}
\end{figure}

\vspace{0.5em}
\noindent\textbf{Action QFormer architecture.}
As shown in Fig.~\ref{fig:action_qformer_architecture}, Action QFormer replaces the direct-fusion interface with a query-based intermediate interface.
It maintains a set of $M$ learnable query states:
\begin{equation*}
Q^0 =
\{q_1^0,\ldots,q_M^0\},
\qquad
Q^0 \in \mathbb{R}^{M \times d}.
\end{equation*}
Given the inherited representations $H_t^I$ and $H_t^S$, Action QFormer updates these query states through a stack of $K$ layers.
At layer $k$, the queries first absorb instruction context and then attend to visual evidence:
\begin{equation*}
\begin{aligned}
\widetilde{Q}^{k}
&=
\left[
\mathrm{SelfAttn}^{k}
\bigl(\mathrm{Concat}(Q^{k-1}, H_t^{S})\bigr)
\right]_{\mathrm{query}}, \\
Q^{k}
&=
\mathrm{CrossAttn}^{k}(\widetilde{Q}^{k}, H_t^{I}),
\qquad k=1,\ldots,K .
\end{aligned}
\end{equation*}
The self-attention update conditions the queries on instruction-side representations, while the cross-attention update lets the instruction-conditioned queries select visual evidence from image-side representations.
After $K$ layers, the final query states carry instruction-conditioned visual information and are pooled and combined with the instruction-side representation to form the final action-facing representation $z_t$ used by the policy head.
We denote the overall interface as
\begin{equation*}
z_t^{\mathrm{AQF}}
=
\mathrm{AQF}(Q^0, H_t^{S}, H_t^{I}).
\end{equation*}
Compared with direct fusion, this design changes how visual information enters action prediction:
image-side information is first selected and abstracted through instruction-conditioned queries before combined with instruction-side representations.

\vspace{0.5em}
\noindent\textbf{Representation-shaping mechanism.}
As illustrated in Fig.~\ref{fig:main_figure}, another key difference between the two interfaces is the route through which action-loss gradients propagate back to inherited representations.
Let $z_t$ denote the action-facing representation that conditions the policy head, and let $\Phi_t$ denote the intermediate output of the action-facing interface.
For direct fusion, $\Phi_t$ is the fused representation; for Action QFormer, it is the query-organized output initialized from $Q^0$.
Let $\theta_{\mathrm{path}}$ denote the trainable parameters of the inherited multimodal pathway.
For both interfaces, the upstream gradient route is:
\begin{equation*}
\begin{aligned}
\frac{\partial \mathcal{L}_{\mathrm{action}}}
{\partial \theta_{\mathrm{path}}}
&=
\frac{\partial \mathcal{L}_{\mathrm{action}}}
{\partial (H_t^I,H_t^S)}
\frac{\partial (H_t^I,H_t^S)}
{\partial \theta_{\mathrm{path}}} \\
&=
\frac{\partial \mathcal{L}_{\mathrm{action}}}
{\partial z_t}
\frac{\partial z_t}
{\partial \Phi_t}
\frac{\partial \Phi_t}
{\partial (H_t^I,H_t^S)}
\frac{\partial (H_t^I,H_t^S)}
{\partial \theta_{\mathrm{path}}}.
\end{aligned}
\end{equation*}
The chain through $z_t$ and $\Phi_t$ also determines how action-loss updates are routed through the action-facing interface.
For Action QFormer, these updates act on both the trainable query-interface parameters and the learnable query initialization:
\begin{equation*}
\begin{aligned}
\frac{\partial \mathcal{L}_{\mathrm{action}}}{\partial \theta_{\mathrm{AQF}}}
&=
\frac{\partial \mathcal{L}_{\mathrm{action}}}{\partial z_t}
\frac{\partial z_t}{\partial \Phi_t}
\frac{\partial \Phi_t}{\partial \theta_{\mathrm{AQF}}}, \\
\frac{\partial \mathcal{L}_{\mathrm{action}}}{\partial Q^0}
&=
\frac{\partial \mathcal{L}_{\mathrm{action}}}{\partial z_t}
\frac{\partial z_t}{\partial \Phi_t}
\frac{\partial \Phi_t}{\partial Q^0}.
\end{aligned}
\end{equation*}
Here, $\theta_{\mathrm{AQF}}$ denotes the trainable parameters of the Action QFormer interface, and $Q^0$ denotes the learnable query initialization, shown as $Q$ in Fig.~\ref{fig:main_figure}.
This gradient structure gives action-loss updates an additional query-specific adaptation path, so action supervision need not be expressed only through shaping the upstream pathway.
By updating the learnable queries, Action QFormer can change how the interface selects and organizes image-side and instruction-side information before gradients propagate back to the inherited pathway.
In this sense, Action QFormer changes the role of the action-facing interface from passively passing inherited representations to actively organizing their action-supervised adaptation.

\begin{table}[h]
\vspace{-1.0em}
\centering
\caption{
\textbf{Gradient settings for representation-shaping analysis.}
}
\vspace{-0.5em}
\label{tab:gradient_settings}
\scriptsize
\begin{tabular}{@{}p{0.28\linewidth}@{\hspace{0.02\linewidth}}p{0.68\linewidth}@{}}
\toprule
\textbf{Setting} & \textbf{Action-loss exposure} \\
\midrule
Action update blocked
& Action-loss gradients are stopped before reaching the extracted image-side and instruction-side representations. \\
Lang. backbone blocked
& The language backbone is blocked from action-loss updates. \\
Vision encoder frozen
& The vision encoder is frozen during action finetuning. \\
Full update
& Action-loss gradients fully propagate through the inherited multimodal pathway. \\
\bottomrule
\end{tabular}
\vspace{-1.0em}
\end{table}

\vspace{0.5em}
\noindent\textbf{Diagnostics for the dual effect.}
To diagnose how gradient-route differences affect the dual effect of action-supervised shaping, we apply stop-gradient interventions to selected parts of the backpropagation path.
For each module or route $r$, we denote the corresponding action-loss update gradient as
\begin{equation*}
\widetilde{g}_{r}
=
m_r
\frac{\partial \mathcal{L}_{\mathrm{action}}}{\partial r},
\qquad
m_r \in \{0,1\}.
\end{equation*}
Here, $m_r=1$ allows route $r$ to receive action-loss updates, while $m_r=0$ applies a stop-gradient operation on that route.

In Sec.~\ref{sec:mechanistic_analysis}, we selectively apply stop-gradient operations to gradient routes from the action-facing representation to the inherited multimodal pathway.
We first use the direct-fusion baseline to examine how action-loss gradients reshape inherited pathways when forming action-facing representations, and then apply the same diagnostics to Action QFormer to test whether its additional adaptation path reduces the disruptive effect of this shaping.
Table~\ref{tab:gradient_settings} summarizes the diagnostic gradient settings used in our analysis.
As an illustrative example, Fig.~\ref{fig:baseline_probe}(b) shows where stop-gradient operations are applied in the action-update-blocked setting for the direct-fusion baseline.

\vspace{0.5em}
\noindent\textbf{Supervision and sequence construction.}
We use the same supervision and sequence construction for both models.
Each sample provides two supervision targets:
a GPT-generated navigation instruction $s_t$ and an $H$-step future action trajectory $A_t$ in the local coordinate frame.
The GPT instruction is generated from the same observation and future trajectory using a constrained prompt, which encourages diverse phrasing while keeping the instruction spatially grounded in the future navigation target.
For backbone training, each sample is formatted as a multimodal sequence:
\begin{equation*}
X_t =
\bigl[
B_I,\ I_t,\ E_I,\ P(g_t),\ B_S,\ s_t,\ E_S
\bigr],
\end{equation*}
where $(B_I,E_I)$ and $(B_S,E_S)$ are boundary markers that define the visual and supervised-instruction spans, and $P(g_t)$ denotes the prompt prefix containing the task text.
Following the standard teacher-forcing recipe, the image observation and task context are treated as conditioning context, and language loss is applied only to the supervised instruction span:
\begin{equation*}
\mathcal{L}_{\mathrm{lang}}
=
-
\sum_{j \in \mathcal{S}_t}
\log p_{\theta}
\left(
x_{t,j}
\mid
x_{t,<j}
\right),
\end{equation*}
where $\mathcal{S}_t$ denotes the token indices of the instruction span.

\vspace{0.5em}
\noindent\textbf{Action prediction objective.}
Let $z_t$ denote the action-facing representation used by the downstream policy head:
\begin{equation*}
z_t =
\begin{cases}
z_t^{\mathrm{Fuse}}, & \text{direct-fusion baseline}, \\
z_t^{\mathrm{AQF}}, & \text{Action QFormer}.
\end{cases}
\end{equation*}
The policy head predicts the local-frame action trajectory conditioned on $z_t$.
When implemented as a conditional diffusion policy, the policy head is trained with a denoising objective.
Given a ground-truth action chunk $A_t$, a diffusion timestep $\tau$, and Gaussian noise $\epsilon$, the noised action chunk is
\begin{equation*}
A_t^{\tau}
=
\sqrt{\bar{\alpha}_{\tau}} A_t
+
\sqrt{1-\bar{\alpha}_{\tau}} \epsilon .
\end{equation*}
The policy head predicts the noise conditioned on the noised action chunk, timestep, and action-facing representation:
\begin{equation*}
\widehat{\epsilon}
=
\epsilon_{\psi}(A_t^{\tau}, \tau, z_t).
\end{equation*}
The action-prediction loss is
\begin{equation*}
\mathcal{L}_{\mathrm{action}}
=
\mathbb{E}_{\tau,\epsilon}
\left[
\left\|
\epsilon -
\epsilon_{\psi}(A_t^{\tau}, \tau, z_t)
\right\|_2^2
\right].
\end{equation*}
During training, we keep the downstream training formulation identical, so that the comparison isolates how the action-facing representation $z_t$ is formed.

\vspace{0.5em}
\noindent\textbf{Overall training objective.}
Training combines instruction-generation supervision and local-frame action-prediction supervision.
The overall objective is controlled by a scalar language-loss weight $r \in [0,1]$:
\begin{equation*}
\mathcal{L} =
\begin{cases}
\mathcal{L}_{\mathrm{action}}, & r = 0, \\
\mathcal{L}_{\mathrm{lang}}, & r = 1, \\
(1-r)\mathcal{L}_{\mathrm{action}} + r\mathcal{L}_{\mathrm{lang}}, & 0 < r < 1.
\end{cases}
\end{equation*}
This design makes it easy to adjust the training emphasis:
when $r=1$, the action branch is skipped and training is language-only;
when $r=0$, the language loss is ignored and training is action-only;
and when $0<r<1$, both branches are evaluated and combined by interpolation.
In the main training runs, $r$ is cosine-scheduled from a high language-side weight to a lower value, giving stronger weight to instruction-generation supervision early in training and gradually increasing the contribution of action-prediction supervision.

\vspace{0.5em}
\noindent\textbf{Inference.}
At inference time, the model is deployed in a closed-loop perception-to-instruction-to-action pipeline.
Given the current real-scene observation $I_t$ and task text $g_t$, the multimodal backbone first generates an intermediate instruction $\widehat{s}_t$.
The resulting sequence is processed by the backbone, and the boundary markers are used to extract image-side and instruction-side representations.
The action interface, either direct fusion or Action QFormer, maps these parsed representations to an action-facing representation $z_t$.
The policy head then samples an $H$-step local-frame action trajectory:
\begin{equation*}
\widehat{A}_t
\sim
\pi_{\psi}(\cdot \mid z_t).
\end{equation*}
The predicted trajectory is executed by the controller, and the resulting new observation is used for the next closed-loop step.

\section{Experiments}

\begin{figure*}[!t]
    \centering

    \begin{minipage}[t]{0.49\textwidth}
        \centering
        {\footnotesize\textbf{(a) Far-End Target Grounding Consistency}}\\[0.25em]
        \includegraphics[width=\linewidth]{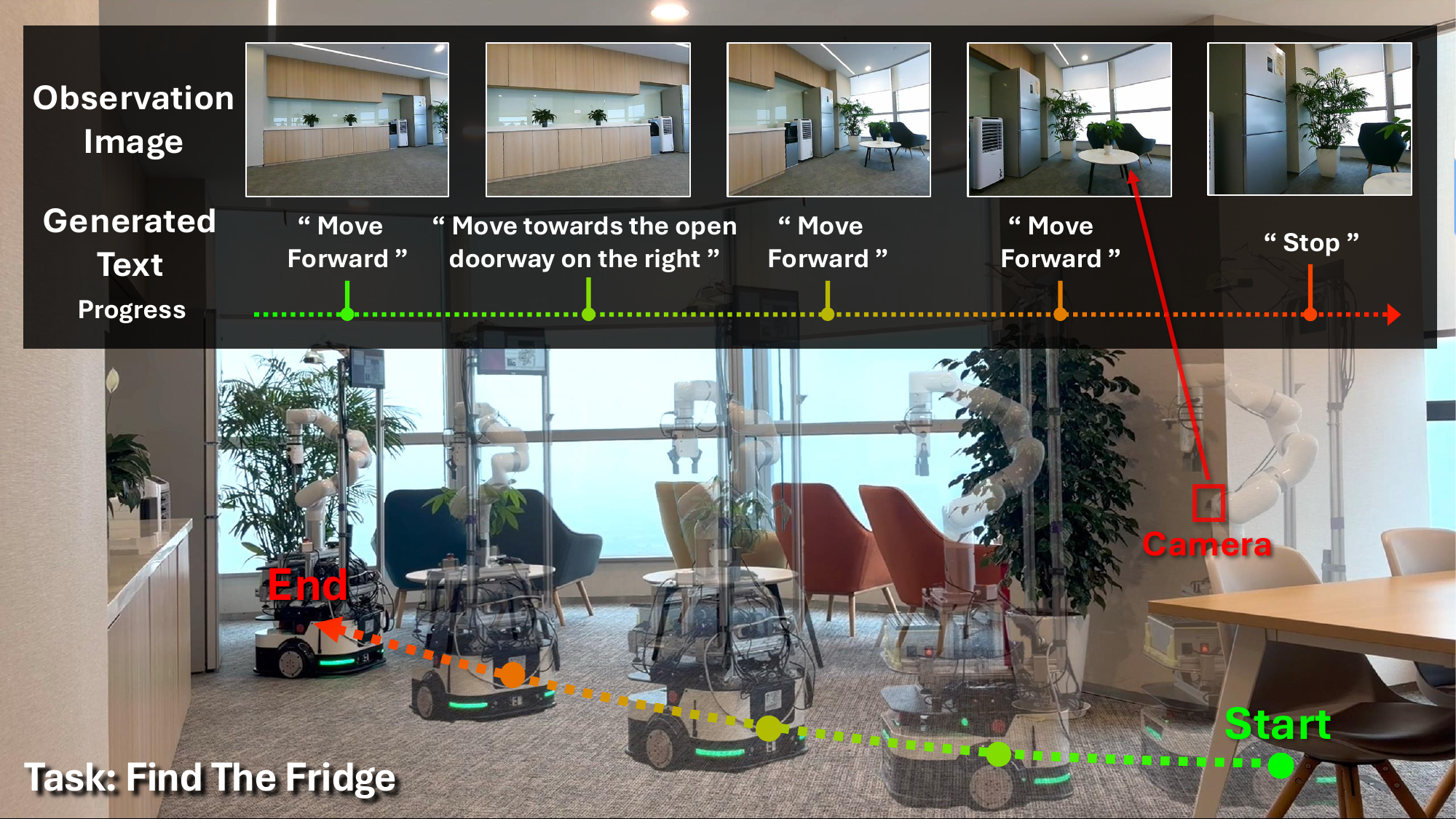}
    \end{minipage}
    \hfill
    \begin{minipage}[t]{0.49\textwidth}
        \centering
        {\footnotesize\textbf{(b) Obstacle-Aware Target Grounding}}\\[0.25em]
        \includegraphics[width=\linewidth]{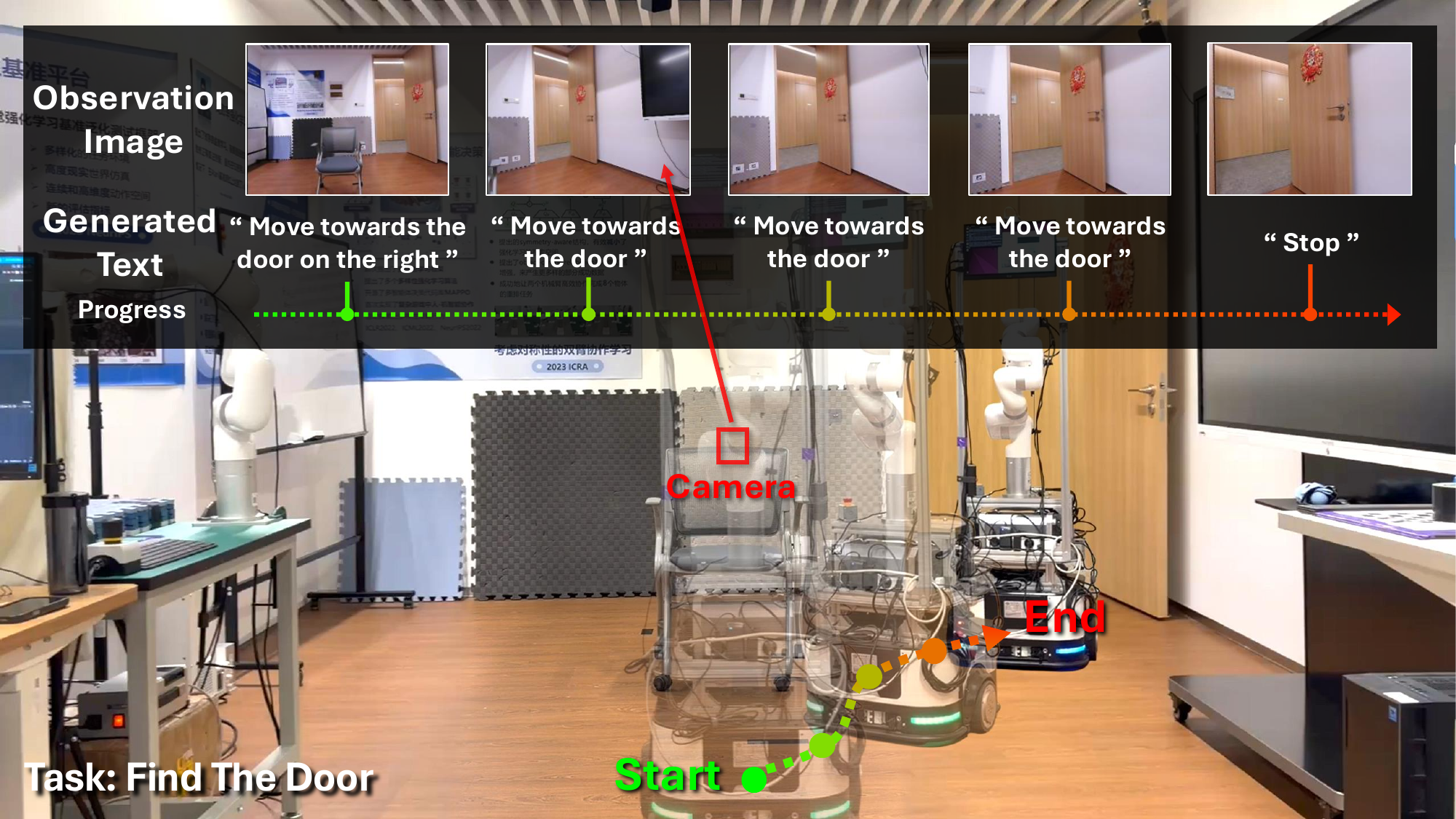}
    \end{minipage}

    \vspace{0.25em}

    \begin{minipage}[t]{0.49\textwidth}
        \centering
        {\footnotesize\textbf{(c) Weak Initial-Cue Directional Inference}}\\[0.25em]
        \includegraphics[width=\linewidth]{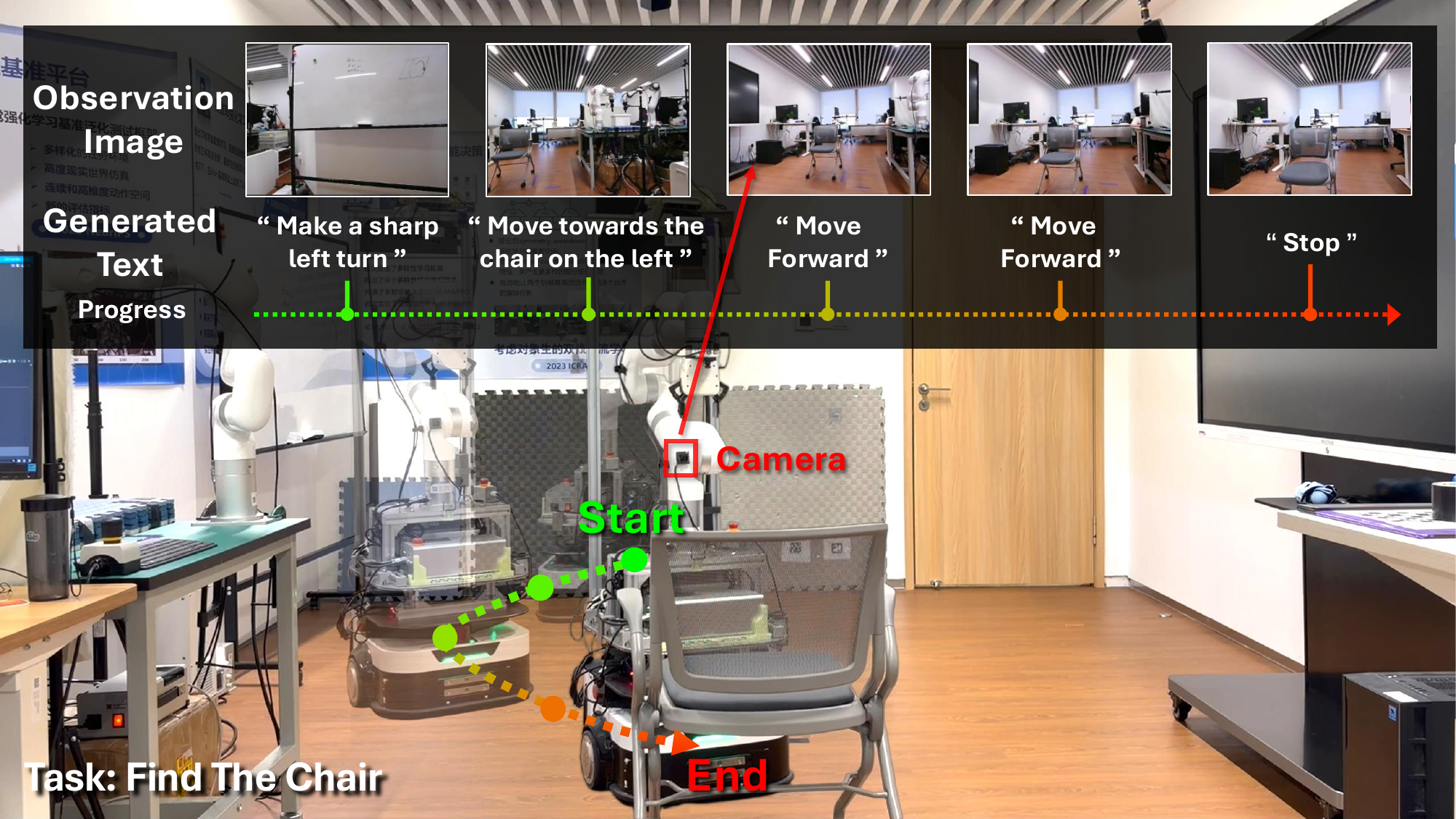}
    \end{minipage}
    \hfill
    \begin{minipage}[t]{0.49\textwidth}
        \centering
        {\footnotesize\textbf{(d) Sharp-Turn Recovery under Visual OOD}}\\[0.25em]
        \includegraphics[width=\linewidth]{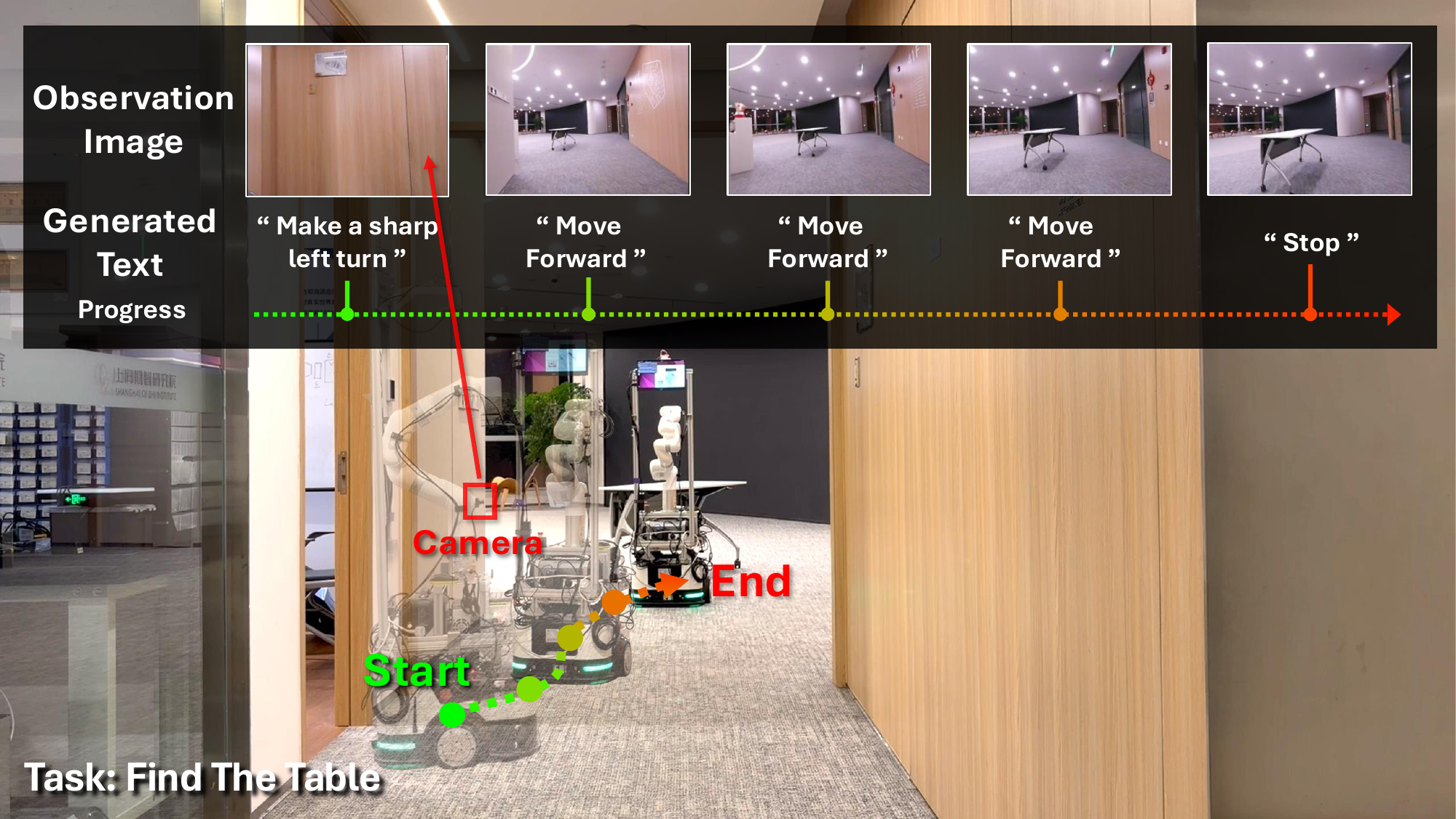}
    \end{minipage}

    \caption{
    \textbf{Four closed-loop real-scene test scenarios (Sec.~\ref{sec:closed_loop_eval}).}
    Each panel shows a third-person rollout overlay together with selected first-person observations and generated instructions along the timeline.
    The four scenarios stress complementary aspects of closed-loop stability:
    \textbf{(a)} far-end target grounding consistency tests whether a distant target near the edge of the field of view remains grounded despite early directional errors;
    \textbf{(b)} obstacle-aware target grounding tests whether target-oriented guidance remains executable in cluttered, collision-prone local geometry;
    \textbf{(c)} weak initial-cue directional inference tests whether the model can recover from weak directional evidence while resisting visually salient distractors;
    and \textbf{(d)} sharp-turn recovery under visual OOD tests whether the instruction-to-action loop remains executable after an abrupt out-of-distribution visual transition.
    }
    \label{fig:closed_loop_scenarios}
    \vspace{-1.0em}
\end{figure*}

\subsection{Experimental Overview}

We evaluate Action QFormer in zero-shot sim-to-real goal-conditioned navigation.
Following the ObjectNav formulation, the goal is specified by a target object category rather than a fixed coordinate, requiring the model to ground the target in the current observation and translate this grounding into executable motion.
In our setting, the VLA model receives a real-scene observation and a language goal, generates an intermediate instruction, and predicts a local-frame action trajectory to move toward the goal while avoiding obstacles.

All models are trained in simulation and directly deployed in real scenes without additional finetuning.
This zero-shot sim-to-real setting exposes the pipeline to large visual variation, making it a sensitive testbed for representation-level instability:
disruptions in inherited grounding or directional representations can surface as invalid or unstable intermediate instructions, or as weakened instruction-following ability in downstream action generation.
Closed-loop execution further increases the difficulty, as early instruction or action errors can compound into target loss, trajectory drift, wrong-direction commitment, collisions, or final task failure.

We organize the experiments at two levels.
First, closed-loop evaluation tests the full perception-to-instruction-to-action pipeline as the model repeatedly generates instructions and actions from changing real-scene observations.
Second, fixed-instruction action generation diagnoses where Action QFormer improves instruction-conditioned execution under the same observations and instructions.
Together, these evaluations test whether Action QFormer improves zero-shot real-scene behavior by stabilizing instruction generation, strengthening instruction-conditioned action execution, or both.

\begin{table*}[!t]
    \centering
    \caption{
    \textbf{Full perception-to-instruction-to-action results in zero-shot sim-to-real closed-loop navigation (Sec.~\ref{sec:closed_loop_eval}).}
    }
    \vspace{-0.5em}
    \label{tab:sim2real_full_pct}
    \footnotesize
    \setlength{\tabcolsep}{3.2pt}
    \renewcommand{\arraystretch}{1.08}
    \begin{tabular}{@{}>{\centering\arraybackslash}m{0.12\textwidth}lccccc|cc@{}}
        \toprule
        & & \multicolumn{5}{c|}{Without provided instructions} & \multicolumn{2}{c}{Given instructions} \\
        \cmidrule(r){3-7} \cmidrule(r){8-9}
        \multicolumn{1}{c}{Model}
        & \multicolumn{1}{c}{Scenario}
        & \makecell{Instruction\\Direction $\uparrow$}
        & \makecell{Instruction\\OOD Rate $\downarrow$}
        & \makecell{Action\\Direction $\uparrow$}
        & \makecell{Average\\Collisions $\downarrow$}
        & \makecell{Task\\Success $\uparrow$}
        & \makecell{Average\\Collisions $\downarrow$}
        & \makecell{Task\\Success $\uparrow$} \\
        \midrule
        \multirow{4}{*}{\makecell[c]{Direct Fusion}}
        & Fridge (far target) & 80.3\% & 100.0\% & 65.6\% & 0.50 & 25.0\% & 0.25 & 75.0\% \\
        & Door (obstacle) & 75.0\% & 97.2\% & 72.2\% & 1.75 & 25.0\% & 1.75 & 87.5\% \\
        & Chair (turn-around) & 46.4\% & 100.0\% & 53.6\% & 0.62 & 25.0\% & 0.75 & 25.0\% \\
        & Table (sharp turn) & 22.9\% & 100.0\% & 16.7\% & 0.75 & 0.0\% & 0.62 & 25.0\% \\
        \midrule
        \multirow{4}{*}{\makecell[c]{Action QFormer}}
        & Fridge (far target) & 95.9\% & \textbf{0.0\%} & 93.2\% & 0.62 & \textbf{37.5\%} & 0.00 & \textbf{100.0\%} \\
        & Door (obstacle) & 97.6\% & \textbf{0.0\%} & 90.5\% & 1.50 & \textbf{62.5\%} & 0.38 & \textbf{87.5\%} \\
        & Chair (turn-around) & 58.8\% & \textbf{0.0\%} & 82.4\% & 0.50 & \textbf{62.5\%} & 0.12 & \textbf{87.5\%} \\
        & Table (sharp turn) & 100.0\% & \textbf{0.0\%} & 93.0\% & 0.38 & \textbf{62.5\%} & 0.12 & \textbf{87.5\%} \\
        \bottomrule
    \end{tabular}
    \vspace{-1.0em}
\end{table*}

\subsection{Zero-shot Sim-to-Real Closed-Loop Experiment}
\label{sec:closed_loop_eval}

We first evaluate the full perception-to-instruction-to-action loop in zero-shot sim-to-real ObjectNav.
At each step, the model observes the current real scene, generates an intermediate instruction, and predicts a local-frame action trajectory conditioned on that instruction.
Because each predicted action changes the next observation, this closed-loop setting provides a stress test of whether instruction generation and action execution remain stable under real-scene visual shift.

To localize failures within the full pipeline, we report metrics at all three levels: instruction-side quality, action-side quality, and final rollout outcomes.
Instruction-side quality is measured by instruction-direction correctness and out-of-distribution (OOD) instruction rate.
Action-side quality is measured by action-direction correctness under the generated instruction.
We also report average collisions and final task success to summarize closed-loop execution quality.

As shown in Table~\ref{tab:sim2real_full_pct}, the direct-fusion baseline exhibits high instruction OOD frequency across all four scenarios, together with weaker instruction-direction correctness.
This indicates that the closed-loop pipeline often becomes unstable at the intermediate instruction stage.
The instability further coincides with weaker action-direction accuracy and lower final task success, suggesting that instruction-side errors can propagate into downstream action execution.
By contrast, Action QFormer nearly eliminates instruction OOD outputs, produces more stable directional instructions, and preserves stronger action-direction accuracy under the same real-scene observations.
These improvements consistently translate into higher end-to-end success across the four test scenarios.

\begin{figure*}[!t]
    \centering

    \newcommand{\rolloutimgheight}{0.115\textheight}
    \newcommand{\casecolwidth}{0.18\textwidth}
    \newcommand{\framecolwidth}{0.195\textwidth}
    \newcommand{\frametitleheight}{1.4em}
    \newcommand{\tagheight}{2.65em}
    \newcommand{\rowblockheight}{0.158\textheight}

    \newcommand{\steptitle}[1]{%
        \parbox[c][\frametitleheight][c]{\linewidth}{\centering\scriptsize\textbf{#1}}\\[0.15em]
    }
    \newcommand{\nosteptitle}{%
        \vspace{\frametitleheight}\vspace{0.15em}
    }
    \newcommand{\casecell}[2]{%
        \parbox[c][\rowblockheight][c]{\casecolwidth}{%
            \raggedright\footnotesize
            \textbf{Task: #1}\\
            #2
        }%
    }
    \newcommand{\framecell}[3]{%
        \parbox[c][\rowblockheight][c]{\framecolwidth}{%
            \includegraphics[width=\linewidth,height=\rolloutimgheight,keepaspectratio]{#2}\\[-0.35em]
            \parbox[t][\tagheight][t]{0.96\linewidth}{%
                \raggedright
                {\scriptsize\textbf{Output:}}\\[0.15em]
                {\fontsize{6.2pt}{6.8pt}\selectfont
                \ttfamily
                #3
                \par
                }
            }%
        }%
    }

    \casecell{find the chair}{consecutive instruction OOD outputs $\rightarrow$ target leaves view $\rightarrow$ wrong object commitment.}
    \hfill
    \framecell{\steptitle{Step 1}}
        {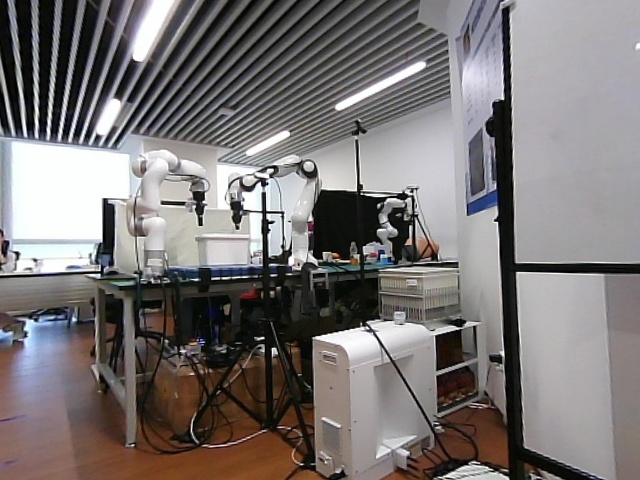}
        {<|im\_start|> <|im\_start|> move towards the left side of the table}
    \hfill
    \framecell{\steptitle{Step 2}}
        {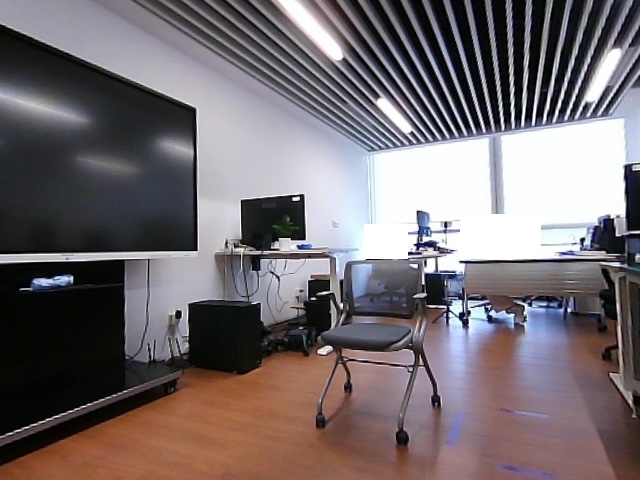}
        {<|im\_start|> <|im\_start|> <|im\_start|> ...}
    \hfill
    \framecell{\steptitle{Step 3}}
        {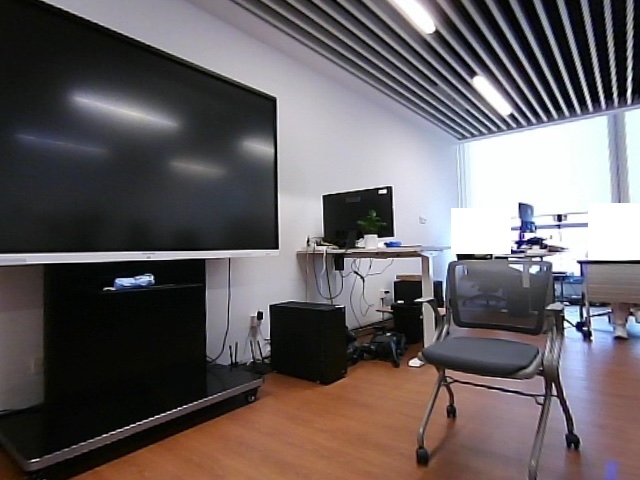}
        {<|im\_start|> <|im\_start|> <|im\_start|> ...}
    \hfill
    \framecell{\steptitle{Step 4}}
        {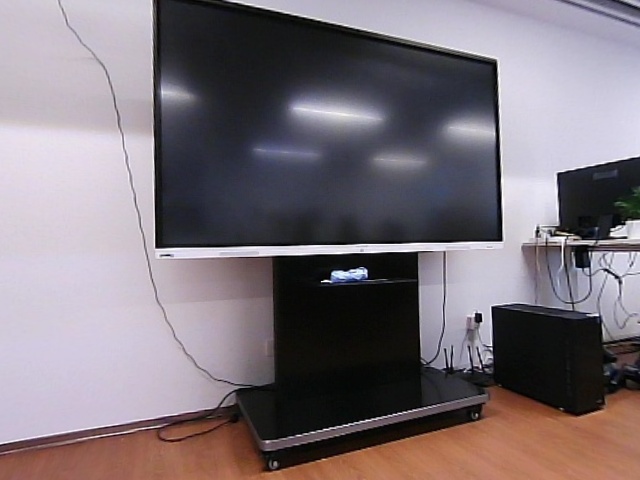}
        {<|im\_start|> <|im\_start|> move towards the TV stand}

    \vspace{0.10em}

    \casecell{find the fridge}{Early instruction--execution mismatch $\rightarrow$ target leaves view $\rightarrow$ wrong direction commitment.}
    \hfill
    \framecell{\nosteptitle}
        {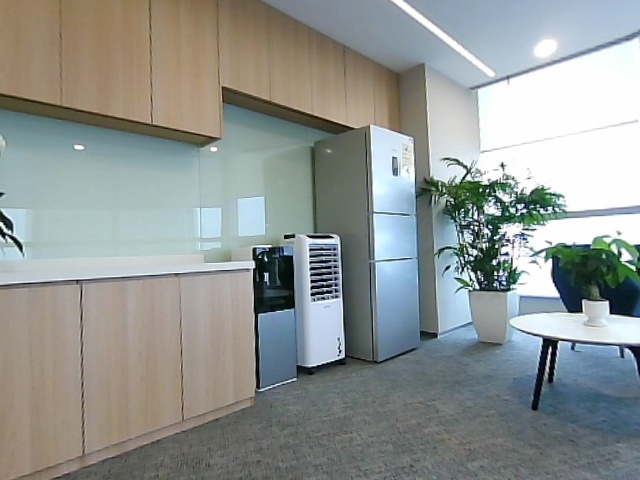}
        {<|im\_start|> <|im\_start|> arc right}
    \hfill
    \framecell{\nosteptitle}
        {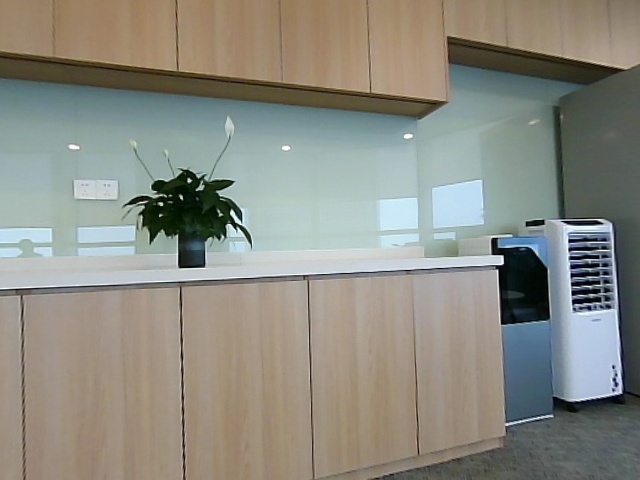}
        {<|im\_start|> veer slightly right}
    \hfill
    \framecell{\nosteptitle}
        {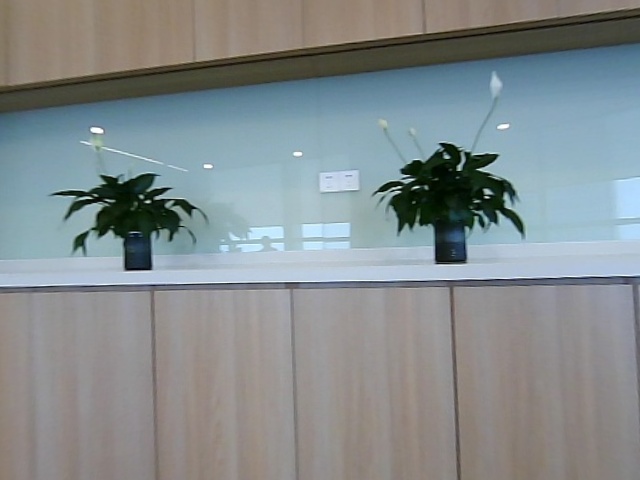}
        {<tool\_call> <tool\_call> <tool\_call> ...}
    \hfill
    \framecell{\nosteptitle}
        {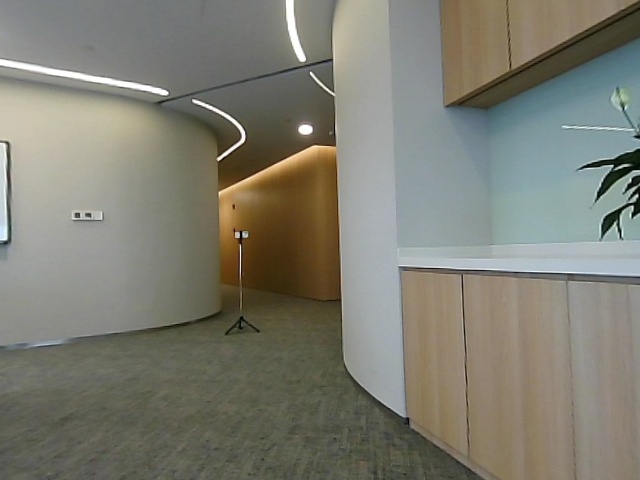}
        {<|im\_start|> move forward}
    \vspace{-1.0em}
    \caption{
    \textbf{Failure progression patterns in representative closed-loop rollouts (Sec.~\ref{sec:closed_loop_eval}).}
    Each row shows one real-scene closed-loop rollout of the direct-fusion baseline, with selected frames arranged from left to right.
    The left column summarizes the task and dominant failure chain, while the compact tags below each frame preserve the key generated instruction fragment.
    The two examples illustrate complementary failure modes:
    consecutive instruction OOD outputs followed by wrong-object commitment, and early instruction--execution mismatch followed by target loss and wrong-direction commitment.
    }
    \label{fig:app_failure_progression}
    \vspace{-1.0em}
\end{figure*}

\vspace{0.5em}
\noindent\textbf{Scenario-level Failure Progression.}
We next examine representative full-chain rollouts of the direct-fusion baseline to understand how instruction-side instability develops into closed-loop failure.
As shown in Fig.~\ref{fig:closed_loop_scenarios}, the real-scene scenarios contain visual variation, including distant targets, distractors, cluttered local geometry, weak initial directional cues, and abrupt viewpoint changes.
Under these conditions, the baseline instruction-generation stage can become unstable, producing non-executable instruction OOD outputs, direction switches, target mismatches, or wrong-object guidance.
Fig.~\ref{fig:app_failure_progression} shows how such instruction-side errors compound through closed-loop execution:
an initially recoverable trajectory can collapse after a distractor or unfamiliar visual cue enters view, and an early instruction–execution mismatch can push the target out of view before the rollout commits to the wrong direction.

\vspace{0.5em}
\noindent\textbf{Controlling for Instruction Instability.}
To test how much of the closed-loop degradation comes from unstable intermediate instructions, we evaluate the same scenarios again with given instructions, while keeping the real-scene observations and downstream action-generation problem unchanged.
As shown in Table~\ref{tab:sim2real_full_pct}, the gap becomes substantially smaller in this controlled setting.
The direct-fusion baseline recovers much of its performance once unstable intermediate instructions are removed, indicating that instruction instability under real-scene visual shift is a major source of behavioral degradation.
At the same time, Action QFormer still retains an advantage in scenarios with large real-scene visual variation, most notably the \emph{Weak Initial-Cue Directional Inference} case.
Even with fixed intermediate instructions, this case requires the model to infer and maintain the correct local direction across changing observations from a cluttered scene and an unfamiliar oblique viewpoint, rather than simply execute a memorized motion pattern.
This indicates that Action QFormer’s gains are not explained by instruction quality alone, but also reflect stronger visually grounded instruction-conditioned action execution.

The closed-loop results, together with the given-instruction control, also clarify the boundary of Action QFormer’s improvement.
The query-based interface improves stability under real-scene visual variation, reducing instruction OOD outputs and strengthening instruction-to-action alignment.
However, it does not by itself solve obstacle-aware instruction planning:
when safe execution requires explicitly routing around nearby obstacles, failures can still occur if the model does not select a suitable grounding target to describe the avoidance behavior.

\begin{figure}[!t]
    \vspace{-0.5em}
    \centering
    \includegraphics[width=\columnwidth]{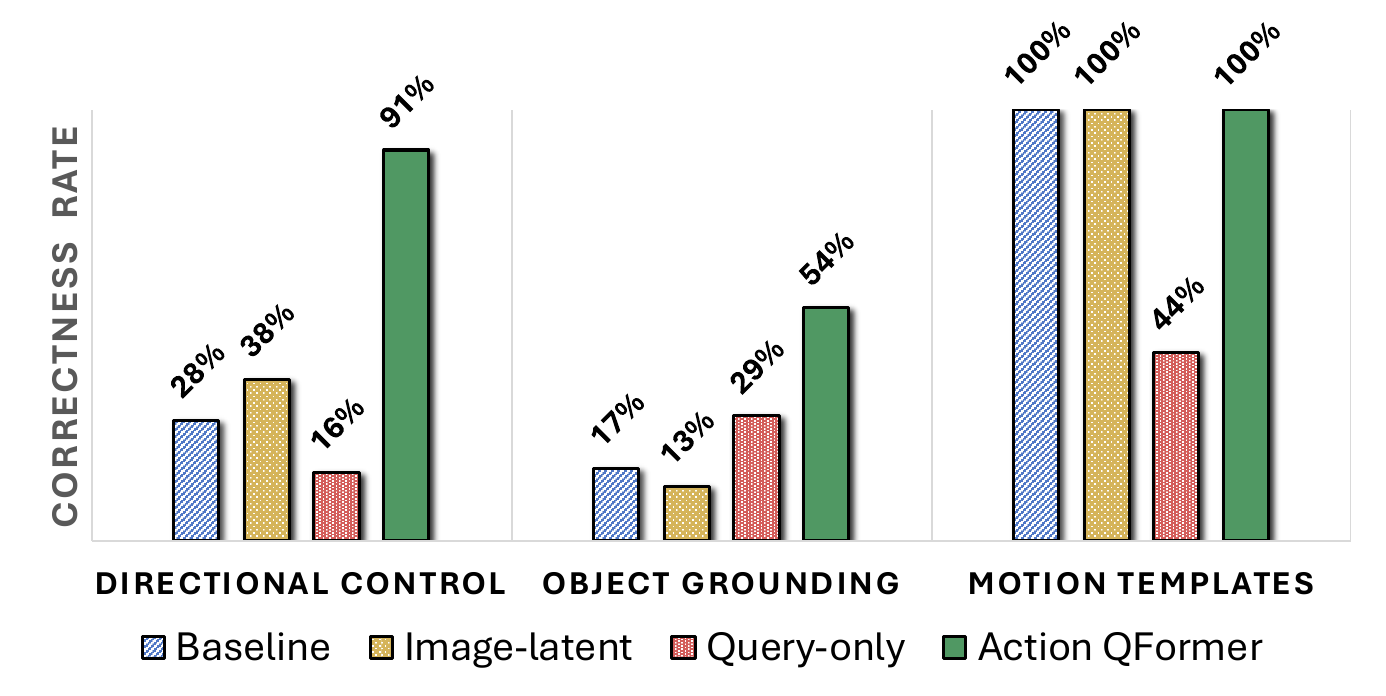}
    \vspace{-2.5em}
    \caption{
    \textbf{Fixed-instruction action generation across scenario families (Sec.~\ref{sec:fixed_instruction_eval}).}
    Family-level correctness under fixed-instruction probes on representative real-scene observations.
    Action QFormer improves most strongly on directional control and object grounding, while gains are smaller on motion-template instructions that require less perceptual disambiguation.
    }
    \label{fig:openloop_action_main}
    \vspace{-1.0em}
\end{figure}

\begin{figure*}[!t]
    \vspace{-1.0em}
    \centering
    \includegraphics[
        width=0.88\textwidth
    ]{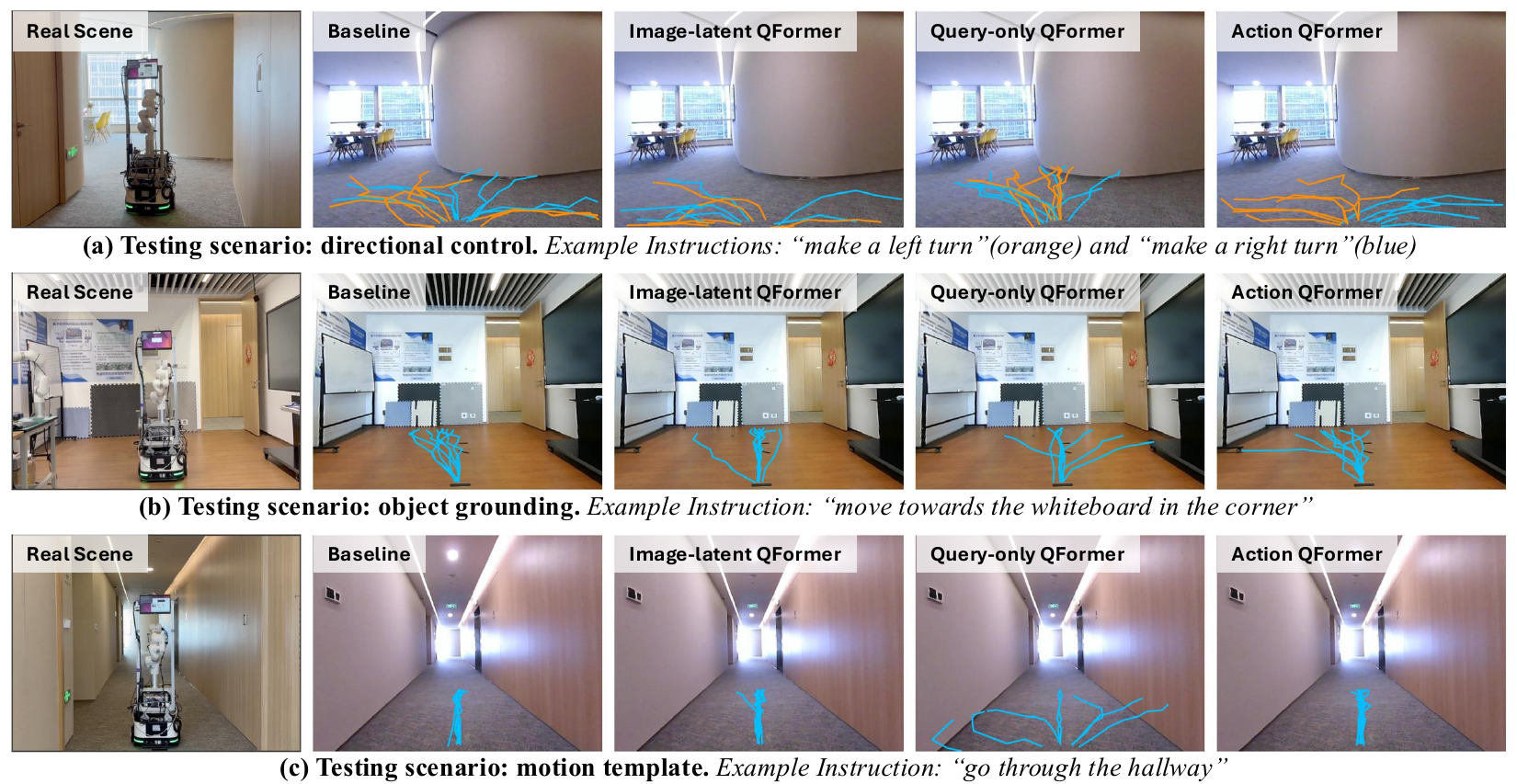}
    \vspace{-1.0em}
    \caption{
    \textbf{Representative qualitative comparisons for fixed-instruction action generation (Sec.~\ref{sec:fixed_instruction_eval}).}
    Rows correspond to the three probe families:
    \textbf{(a)} directional control,
    \textbf{(b)} object grounding,
    and \textbf{(c)} motion-template following.
    In each row, the leftmost image shows the third-person scene context, and the remaining columns show action predictions generated from the same first-person observation and fixed instruction.
    The prediction columns compare, from left to right, the \textbf{Direct-fusion Baseline}, \textbf{Image-latent Action QFormer}, \textbf{Query-only Action QFormer}, and \textbf{Action QFormer}.
    Thin trajectories show 8 stochastic inference samples from the same model, and the bold trajectory denotes the selected representative action trajectory used for comparison.
    }
    \label{fig:action_generation_structural_qualitative}
    \vspace{-1.0em}
\end{figure*}

\subsection{Fixed-Instruction Action Generation}
\label{sec:fixed_instruction_eval}

Next, we evaluate fixed-instruction action generation on representative real-scene observations from challenging closed-loop cases.
By providing the same observation and intermediate instruction to each model, this setting controls for instruction-generation instability and isolates whether the model can produce the intended local action.

We organize testing scenarios into three families:
\emph{directional control}, which tests whether the model follows left- or right-oriented local motion;
\emph{object grounding}, which tests whether the model moves toward the instructed visual target rather than a distractor or irrelevant object;
and \emph{motion templates}, which test whether the model follows simple instruction-specified motion patterns that depend less on visual disambiguation.

As shown in Fig.~\ref{fig:openloop_action_main}, family-level correctness reveals a clear pattern:
once instruction-generation instability is controlled, Action QFormer shows its strongest advantage on \emph{directional control}, a smaller but still substantial gain on \emph{object grounding}, and a much smaller gap on \emph{motion templates}, where the baseline remains comparatively competitive.
Fig.~\ref{fig:action_generation_structural_qualitative} provides representative qualitative comparisons from each family across model variants.
This indicates that the main action-side benefit of Action QFormer is not generic trajectory smoothing, but a more reliable action-facing representation when behavior depends on directional disambiguation or target-directed motion.

\vspace{0.5em}
\noindent\textbf{Interface ablations.}
We next ablate two parts of the Action QFormer interface to test where this fixed-instruction advantage comes from.
The first ablation replaces the default visual source of Action QFormer---the image embedding output directly by the vision encoder before language-backbone fusion—with the \emph{image latent}, i.e., the image-side representation after it has been processed by the language backbone.
The image-latent variant retains part of the competence of the full model, but loses much of its distinctive advantage, especially on directional control and object grounding.
This suggests that the image latent still contains useful visual information, but may no longer expose control-relevant spatial information in a form sufficient to support the full Action QFormer benefit.

The second ablation removes the instruction-side representation combined with the query outputs, sending only the updated queries to downstream action generation.
This query-only variant often still produces visually plausible motion, but loses reliable instruction-following ability.
This suggests that the updated queries mainly carry instruction-conditioned visual information extracted through cross-attention, but explicit instruction context remains necessary for forming a complete action-facing representation.
Representative qualitative comparisons for both ablations are shown in Fig.~\ref{fig:action_generation_structural_qualitative}.

\vspace{0.5em}
Overall, Action QFormer achieves consistent behavioral gains in zero-shot real-scene navigation.
Compared with the baseline, it improves closed-loop stability by reducing instruction OOD outputs and strengthens fixed-instruction action generation when behavior depends on directional disambiguation or target-grounded visual evidence.

\section{Mechanistic Analysis}
\label{sec:mechanistic_analysis}

\begin{figure*}[!t]
    \centering
    \includegraphics[width=0.95\textwidth]{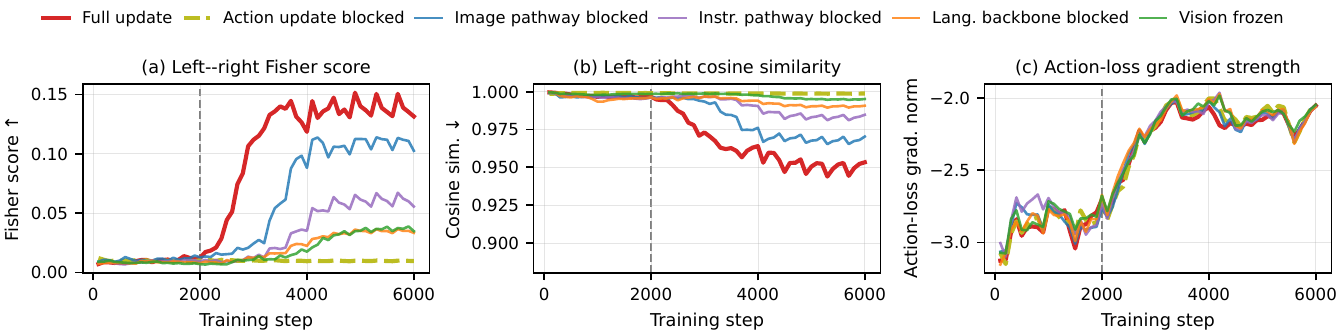}
    \vspace{-1.0em}
    \caption{
    \textbf{The left--right distinction becomes clearest when both visual and language-side pathways are shaped by action-loss gradients (Sec.~\ref{sec:directional_distinction}).}
    We train the direct-fusion baseline while selectively blocking gradients from the action loss to different parts of the multimodal pathway.
    This plot additionally includes two route-level variants that block gradients to the fused image or instruction representation.
    \textbf{(a)} Left--right Fisher score between action-facing representations; higher values indicate a clearer distinction.
    \textbf{(b)} Left--right cosine similarity between the same representations; lower values indicate a clearer distinction.
    \textbf{(c)} Action-loss gradient norm in the shared multimodal backbone, measuring how strongly the action objective updates the inherited representation pathway.
    The dotted vertical line marks the 2K training step, after which the action-loss gradient norm rises and the left--right distinction becomes more pronounced.
    Together, these trends indicate that left--right distinction emerges through joint action-loss gradient exposure of the visual and language pathways.
    }
    \label{fig:gradient_decoupling}
    \vspace{-1.0em}
\end{figure*}

We now examine the dual effect of action-supervised shaping behind the behavioral gains observed above: action-loss updates can form     behaviorally consequential distinctions, but may also induce disruptive upstream rewriting.

Across the analyses, we first use the direct-fusion baseline to expose how direct shaping acts on inherited multimodal pathways, and then use the same diagnostic settings to examine structured shaping in Action QFormer.
To isolate shaping caused by action supervision, we use the gradient settings in Table~\ref{tab:gradient_settings}.
The \emph{action-update-blocked} setting serves as an instruction-supervision-only reference, so comparisons against action-loss-exposed settings reveal the additional upstream shaping induced by action supervision.

We analyze this dual effect at three levels.
First, we examine its constructive side by studying how action-supervised shaping forms directional distinctions.
Second, we measure upstream token rewriting to assess whether the same adaptation is accompanied by disruptive rewriting of inherited multimodal representations.
Third, we analyze instruction-to-visual attention to understand how upstream rewriting affects attention to visual evidence for object grounding.

\begin{table}[!t]
    \vspace{-0.5em}
    \centering
    \caption{
    Left--right cosine similarity across interface and gradient settings.
    Lower indicates clearer left--right distinction.
    }
    \vspace{-0.5em}
    \label{tab:directional_distinction_main}
    \scriptsize
    \setlength{\tabcolsep}{3.6pt}
    \renewcommand{\arraystretch}{1.12}
    \begin{tabular}{lcccc}
        \toprule
        Gradient Setting &
        Action $\downarrow$ &
        Action Repr. $\downarrow$ &
        Instr. Repr. $\downarrow$ &
        Img. Repr. $\downarrow$ \\
        \midrule
        \multicolumn{5}{l}{\textbf{Direct-fusion baseline}} \\
        Action update blocked & 0.895 & 0.941 & 0.969 & 0.958 \\
        Vision encoder frozen & 0.894 & 0.940 & 0.971 & 0.958 \\
        Lang. backbone blocked & 0.901 & 0.953 & 0.969 & 0.969 \\
        Full update & \textbf{0.891} & \textbf{0.931} & 0.976 & 0.976 \\
        \midrule
        \multicolumn{5}{l}{\textbf{Action QFormer}} \\
        Action update blocked & 0.898 & 0.967 & 0.960 & 0.958 \\
        Vision encoder frozen & 0.924 & 0.959 & 0.958 & 0.958 \\
        Lang. backbone blocked & 0.923 & 0.979 & 0.963 & 0.963 \\
        Full update & \textbf{0.470} & \textbf{0.520} & 0.936 & 0.966 \\
        \bottomrule
    \end{tabular}
    \vspace{-1.5em}
\end{table}

\subsection{Action-facing Directional Distinction}
\label{sec:directional_distinction}

We begin with the constructive side of action-supervised shaping by using left--right directional control as a minimal diagnostic distinction.
This distinction is linguistically close but behaviorally consequential:
in a local frame, predicting a leftward trajectory instead of a rightward one corresponds to a substantially different action target.
It therefore provides a focused setting for testing whether action-loss gradients make this distinction more explicit in action-facing representations, and through which update routes.

For each diagnostic configuration in Table~\ref{tab:gradient_settings}, we extract action-facing representations from samples with explicit left or right instructions and measure their separation in representation space.
Fig.~\ref{fig:gradient_decoupling} shows that the left--right distinction emerges during direct-fusion baseline training only when action-loss updates are allowed to shape upstream multimodal representations.
Blocking either the visual pathway or the language-side pathway weakens the distinction, while blocking all action-loss updates almost prevents the separation from forming.
Moreover, under the full-update setting, the distinction emerges rapidly only in the later stage of optimization, closely aligned with the rise of action-loss gradient magnitude in the shared backbone.
Together, these trends show that the left--right distinction is not simply inherited from the pretrained representation or formed by the policy head alone.
Instead, the clearest action-facing directional distinction appears when visual and language-side pathways are jointly shaped by gradients from the action loss.

Having shown in the baseline that action-supervised shaping is critical in forming directional distinctions, we now use the same diagnostic settings to test whether Action QFormer preserves this constructive effect.
Table~\ref{tab:directional_distinction_main} reports left--right cosine similarity at four locations: visual representation, instruction representation, action-facing representation, and final action output.
Consistent with the gradient-blocking result above, the full-update setting produces the clearest left–right distinction among the representative gradient settings.
Under this setting, Action QFormer sharpens the distinction in both \texttt{Action Repr.} and \texttt{Action}.
By contrast, the differences in \texttt{Img. Repr.} and \texttt{Instr. Repr.} between the two interfaces are much smaller.
This suggests that Action QFormer does not simply inherit a stronger upstream left--right distinction, but reorganizes visual and language information into a clearer action-facing directional distinction.

\begin{figure*}[!t]
    \centering
    \includegraphics[
        width=0.95\textwidth
    ]{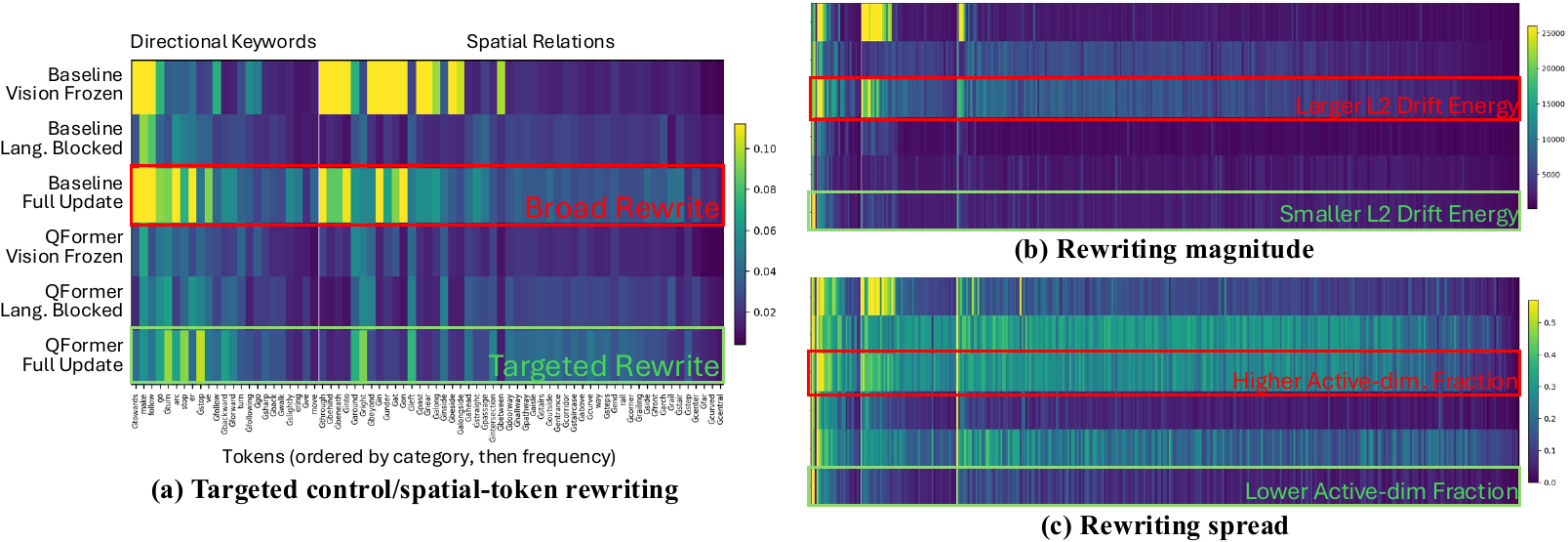}
    \vspace{-1.0em}
    \caption{
    \textbf{Action QFormer reduces upstream token rewriting while preserving targeted adaptation (Sec.~\ref{sec:upstream_rewriting}).}
    All panels compare the same six gradient settings shown in panel~(a).
    \textbf{(a)} A control/spatial-token view measures token-level cosine distance from the action-update-blocked reference; higher values indicate stronger targeted token modification.
    The baseline broadly rewrites control-related token groups, whereas Action QFormer concentrates adaptation on a more selective subset of key tokens, including directional and motion tokens such as \emph{left}, \emph{right}, \emph{turn}, \emph{around}, and \emph{stop}.
    \textbf{(b)} Rewriting magnitude, measured by token-wise drift energy, shows that the direct-fusion baseline undergoes larger-scale token rewriting, whereas Action QFormer reduces the overall magnitude of rewriting.
    \textbf{(c)} Rewriting spread, measured by active-dimension fraction, captures how broadly rewriting extends across hidden dimensions within each token.
    }
    \label{fig:mechanistic_drift_global}
    \vspace{-1.5em}
\end{figure*}

\subsection{Reduced Upstream Rewriting with Targeted Adaptation}
\label{sec:upstream_rewriting}

We next examine the disruptive side of action-supervised shaping: whether the same action-loss updates may also induce broad upstream rewriting.
Using the \emph{action-update-blocked} model as the reference, we measure upstream token rewriting as changes in token representations under action-loss-exposed training settings.

We analyze token rewriting through two complementary aspects:
\emph{rewriting magnitude} and \emph{rewriting spread}.
For each token representation $h_t \in \mathbb{R}^d$, we quantify rewriting magnitude using \emph{token-wise drift energy}, computed as the mean squared representation change relative to the reference, $\frac{1}{d}\|h_t - h_t^{\mathrm{ref}}\|_2^2$.
We quantify rewriting spread using \emph{active-dimension fraction}, computed as the fraction of hidden dimensions with nonzero change relative to the reference.
This distinguishes broad high-dimensional rewriting from sparser, more targeted adaptation.

Fig.~\ref{fig:mechanistic_drift_global}(b) shows that Action QFormer has substantially lower rewriting magnitude than the direct-fusion baseline, especially under the full-update setting.
This indicates that the action-facing query interface reduces the overall strength of upstream rewriting across token representations.
The rewriting-spread view in Fig.~\ref{fig:mechanistic_drift_global}(c) further shows that this reduction is not only about magnitude:
the direct-fusion baseline rewrites many tokens across a larger fraction of hidden dimensions, whereas Action QFormer keeps rewriting substantially more confined.

However, reduced global rewriting alone does not show whether Action QFormer still preserves useful action-driven adaptation.
To test this, we zoom in on control and spatial tokens, where action-driven shaping should remain visible if the model is still forming useful action-facing distinctions.
For this targeted view, we measure token-level \emph{cosine distance} from the action-update-blocked reference, defined as $1-\cos(h_t, h_t^{\mathrm{ref}})$; higher values indicate stronger directional changes in token representations.
Fig.~\ref{fig:mechanistic_drift_global}(a) shows that Action QFormer preserves targeted rewriting on key control and spatial tokens, including \emph{left}, \emph{right}, \emph{turn}, \emph{around}, and \emph{stop}, while avoiding the broad rewriting observed in the baseline.

\begin{figure}[!t]
    \centering
    \includegraphics[
        width=\linewidth,
        trim=7.2mm 0mm 0 5mm,
        clip
    ]{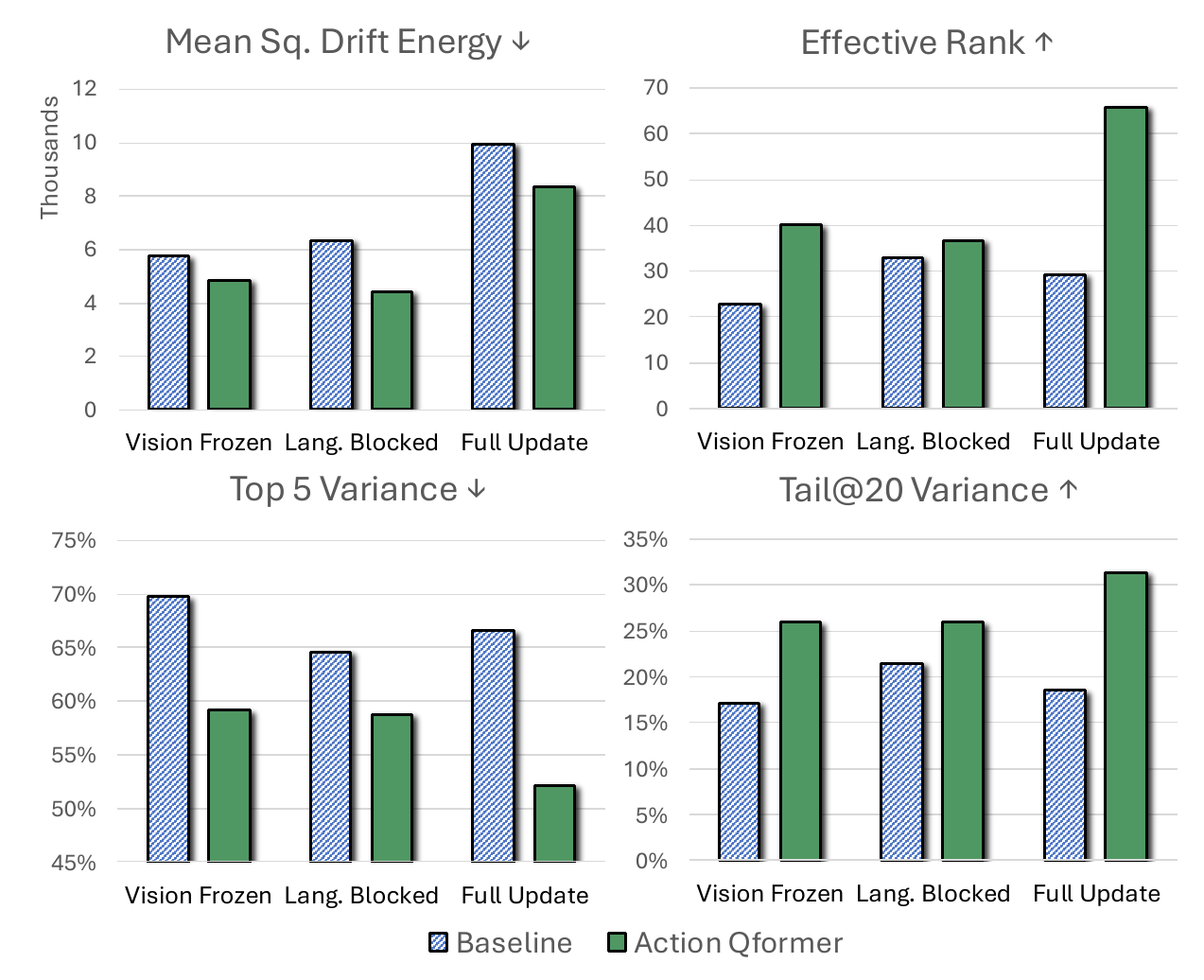}
    \vspace{-2.0em}
    \caption{
    \textbf{Pairwise subspace structure of upstream token rewriting (Sec.~\ref{sec:upstream_rewriting}).}
    \textbf{(a)} Drift energy measures the overall magnitude of representation rewriting.
    \textbf{(b)} Effective rank measures how broadly rewriting spans independent representation directions.
    \textbf{(c)} Top-5 variance concentration measures whether rewriting is dominated by a small set of principal directions.
    \textbf{(d)} Tail@20 variance measures how much drift energy remains outside the top 20 principal directions, reflecting the long-tail spread of adaptation.
    Action QFormer reduces rewriting magnitude while increasing effective rank, lowering top-5 concentration, and increasing long-tail variance, indicating a less low-rank and more distributed adaptation pattern.
    } 
    \label{fig:rewriting_subspace_stats}
    \vspace{-1.0em}
\end{figure}

\begin{figure}[!t]
    \vspace{-0.5em}
    \centering
    \includegraphics[
        width=\linewidth,
        trim=0mm 6mm 0 10mm,
        clip
    ]{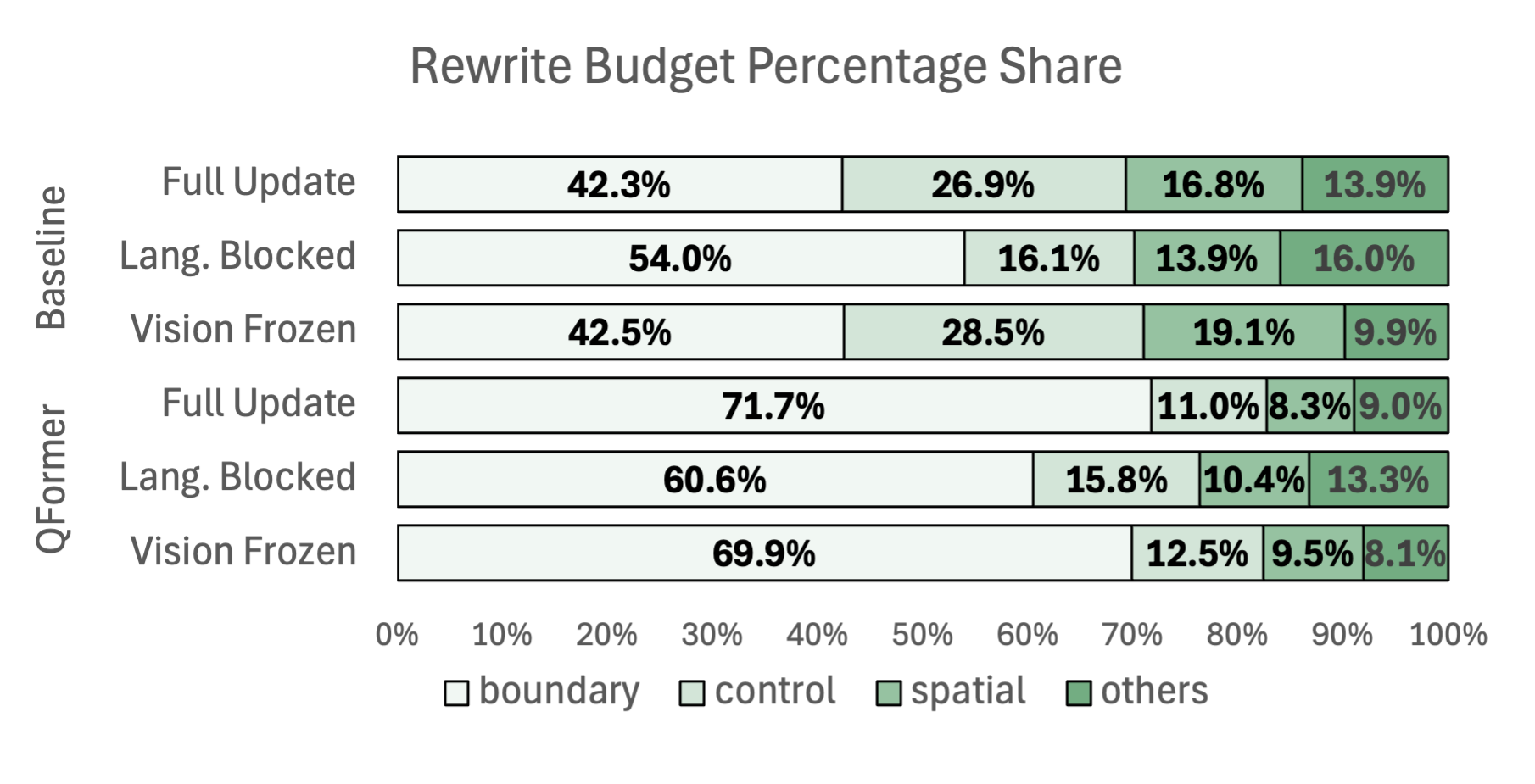}
    \vspace{-2.0em}
    \caption{
    \textbf{Token rewrite share across token groups (Sec.~\ref{sec:upstream_rewriting}).}
    We partition tokens into boundary, control, spatial, and other groups, and report the percentage share of total token rewriting assigned to each group.
    Boundary tokens denote the special tokens inserted at the beginning and end of the instruction span.
    Compared with the baseline, Action QFormer allocates a larger fraction of rewriting to boundary tokens while reducing broad allocation to control and spatial token groups.
    This indicates that Action QFormer shifts action-loss-driven shaping toward more localized boundary-level adaptation and reduces collateral rewriting over semantically meaningful content tokens.
    }
    \label{fig:token_rewrite_share}
    \vspace{-1.0em}
\end{figure}

The subspace statistics of token rewriting in Fig.~\ref{fig:rewriting_subspace_stats} further show that this effect is not merely a reduction in rewriting magnitude.
Action QFormer also increases the effective rank of the rewriting directions while lowering top-5 variance concentration and increasing Tail@20 variance, indicating that the remaining adaptation is less dominated by a few principal directions.
Fig.~\ref{fig:token_rewrite_share} shows that this change in rewriting structure is accompanied by a shift in rewrite energy allocation:
rather than broadly rewriting control and spatial tokens, Action QFormer allocates more rewriting to boundary tokens while preserving targeted adaptation on selected action-relevant tokens.

\vspace{-0.5em}
\subsection{Stable Attention with Constructive Adaptation}
\label{sec:attention_stability}

Finally, we examine whether action-supervised shaping also changes how instruction tokens attend to visual evidence.
We therefore analyze phrase-level instruction-to-visual attention maps as an observable proxy for visual grounding.
For each action-loss-exposed model, we compare its attention map against the corresponding action-update-blocked reference.

We examine this behavior from two complementary perspectives:
\emph{attention stability}, measuring alignment with the reference visual region, and \emph{constructive adaptation}, assessing whether action-supervised training sharpens attention toward visual evidence for object grounding.

\begin{figure}[!t]
    \centering
    \includegraphics[
        width=\columnwidth
    ]{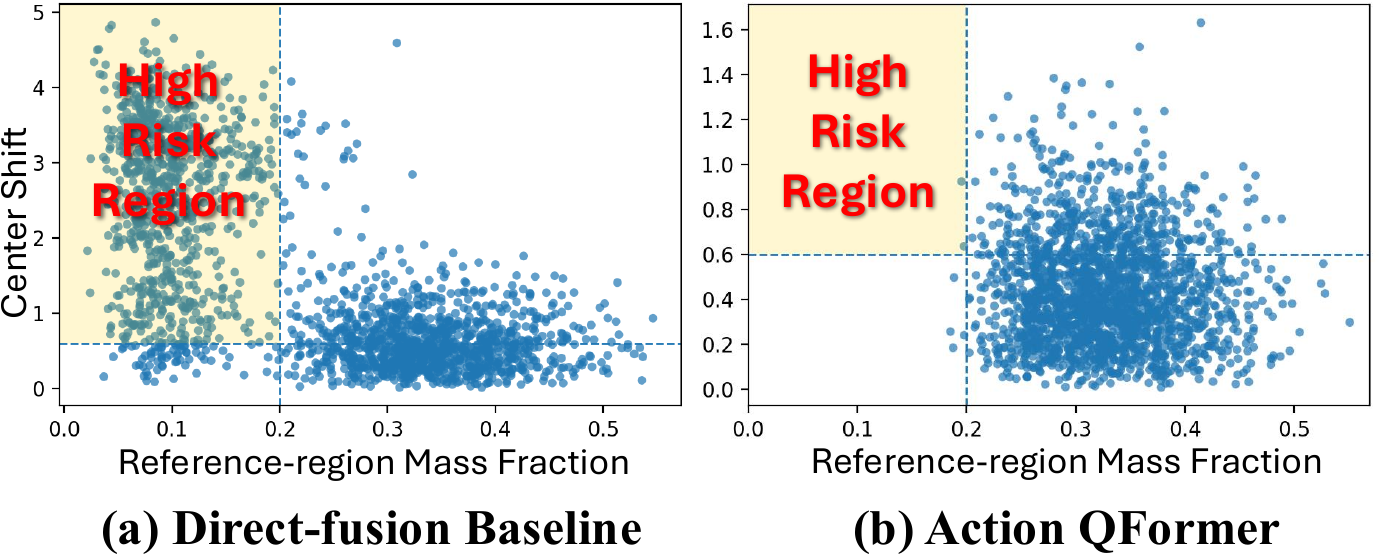}

    \vspace{-0.5em}
    \caption{
    \textbf{Action QFormer preserves more stable instruction-to-visual attention (Sec.~\ref{sec:attention_stability}).}
    Each point compares a full-update model with its action-update-blocked reference counterpart.
    Points in the lower-right region correspond to more stable attention, while the shaded upper-left region marks high-risk attention changes with low reference-region mass and large spatial drift.
    Compared with the baseline, Action QFormer produces fewer high-risk cases and remains more concentrated in the stable-attention region.
    }
    \label{fig:attention_joint_main}
    \vspace{-1.0em}
\end{figure}

\vspace{0.5em}
\noindent\textbf{Attention stability.}
We quantify attention stability using two phrase-level metrics.
Given the top-$10\%$ high-attention visual region in the action-update-blocked reference, \emph{reference-region mass} measures how much attention from the action-loss-exposed model remains inside this region, while \emph{center shift} measures the distance between the two attention centers.
Higher reference-region mass and lower center shift indicate more stable attention.
Together, these metrics capture whether attention remains on the reference visual region and whether its spatial center remains stable.

As shown in Fig.~\ref{fig:attention_joint_main}, the baseline produces many high-risk attention cases under the full-update setting, with low reference-region mass and large center shift.
Action QFormer shifts the distribution toward higher reference-region mass and smaller center shift, indicating more stable instruction-to-visual attention under the same full-update setting.
Additional gradient-blocking variants, reported in the Supplementary Materials, show the same overall stability trend, although the vision-frozen Action QFormer variant exhibits larger attention shifts, suggesting that attention stability is strongest when representations are shaped jointly through visual and language-side action-loss pathways.

\begin{figure}[!t]
    \vspace{-0.2em}
    \centering
    \includegraphics[
        width=0.80\columnwidth,
        trim=0 65pt 0 60pt,
        clip
    ]{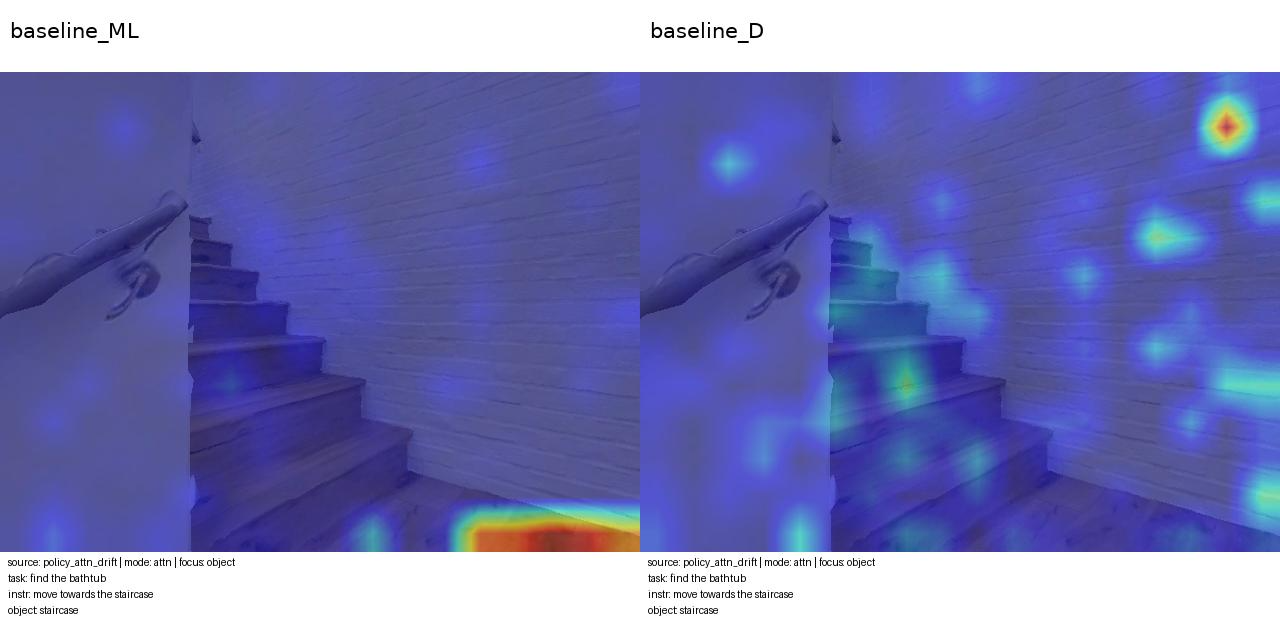}\\[-0.5em]
    {\footnotesize
    \textbf{(a) Baseline.}
    Full update vs.\ action-update-blocked reference
    }\\

    \vspace{0.1em}
    
    \includegraphics[
        width=0.80\columnwidth,
        trim=0 65pt 0 60pt,
        clip
    ]{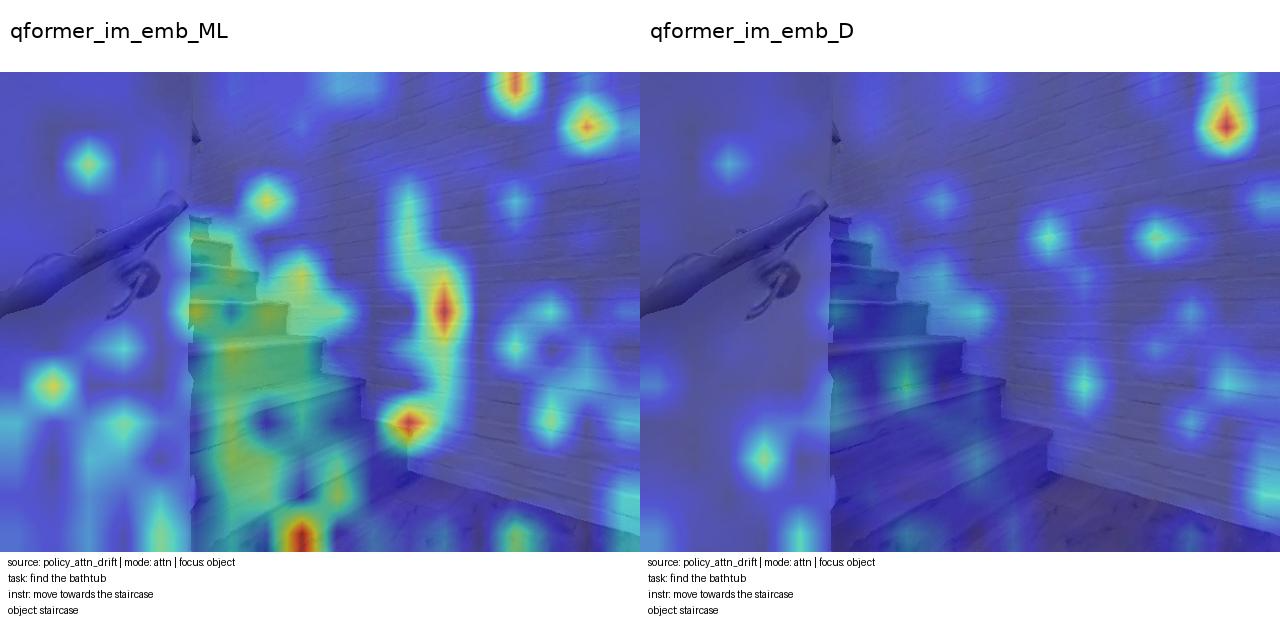}\\[-0.5em]
    {\footnotesize
    \textbf{(b) Action QFormer.}
    Full update vs.\ action-update-blocked reference
    }\\

    \vspace{-0.5em}
    \caption{
    \textbf{Action QFormer can sharpen visual attention under action-loss exposure (Sec.~\ref{sec:attention_stability}).}
    For the instruction \emph{move towards the staircase}, we compare the full-update model with the action-update-blocked reference.
    In the baseline, action-loss exposure shifts attention away from the staircase region toward an irrelevant visual-token region, whereas in Action QFormer it strengthens attention around the staircase region relative to its reference.
    }
    \label{fig:attention_focus_main}
    \vspace{-1.0em}
\end{figure}

\vspace{0.5em}
\noindent\textbf{Constructive adaptation.}
Beyond preserving attention stability, Action QFormer sometimes produces a qualitative \emph{attention focus} effect, where action-supervised training sharpens instruction-to-visual attention toward the target object region.
We do not treat this effect as a universal mechanism, but as qualitative evidence that action-supervised shaping can sometimes produce constructive upstream adaptation.
Fig.~\ref{fig:attention_focus_main} shows a representative staircase example comparing the full-update model with the action-update-blocked reference:
under the same action-loss exposure, the baseline remains weakly grounded with substantial attention mass away from the staircase, whereas Action QFormer sharpens attention toward the staircase itself.
This effect may help reduce instruction instability in closed-loop experiments:
when upstream attention remains grounded in task-relevant visual evidence, instruction generation is less likely to drift toward irrelevant regions or non-executable outputs.

\vspace{0.5em}
Taken together, these analyses characterize the dual effect of action-supervised representation shaping and show that Action QFormer changes this process toward more structured adaptation.
It preserves the constructive effect by strengthening control-relevant distinctions such as left--right directionality, while reducing the disruptive effect through more targeted token-level shaping and less broad drift over semantically meaningful content tokens.
At the attention level, it reduces high-risk attention shifts and preserves more stable instruction-to-visual grounding, consistent with the reduced instruction instability observed in closed-loop experiments.

\section{Discussion}

This work establishes structured representation shaping under action supervision as a mechanism-level problem in VLA systems.
Rather than treating action supervision only as a downstream training objective, our results suggest a different view:
when action-loss gradients propagate through inherited multimodal pathways, they reshape the representation space on which both language-side processing and action-side control depend.
This shaping is necessary for forming behaviorally consequential distinctions, such as left--right directionality, but can become disruptive when applied too directly through the inherited multimodal pathway.
Thus, action-interface design should consider not only how to decode actions from multimodal representations, but also how to allow action supervision to reshape those representations during action finetuning.

This question reflects a deeper tension in how inherited multimodal representations are used.
Language-side processing emphasizes semantic and contextual structure, whereas action generation requires stable control-relevant distinctions such as direction, target location, and small spatial changes.
Part of this tension appears in how visual evidence is attended to: the regions useful for language-side grounding are not always the same as those needed for action execution.
Action QFormer alleviates this tension by introducing a query-based action-facing interface that mediates bidirectionally between the inherited multimodal space and the action-facing representation space.
Its instruction-conditioned queries provide an action-facing route for accessing visual evidence, allowing control-relevant visual selection to form without directly disrupting inherited representations.
By concentrating part of the control-relevant learning pressure within this query interface, Action QFormer allows the remaining upstream shaping to become more selective, preserving representations and attention patterns that support language-side grounding while still permitting useful, and sometimes constructive, action-supervised adaptation.

This perspective also suggests a complementary way to understand visual robustness in VLA models.
Real-scene brittleness is often treated primarily as a domain-transfer problem, where changes in lighting, viewpoint, texture, or background appearance are expected to require more diverse real-world data.
Our results suggest another factor:
visual information must also be abstracted into the control-relevant factors needed for action.
Even when the inherited multimodal pathway contains useful visual evidence, it may still carry appearance-level variation irrelevant to action, making the downstream policy more sensitive to real-scene visual changes.
Thus, improving VLA reliability may require not only richer visual data or stronger visual backbones, but also better action-oriented abstraction of visual information.

\section{Conclusion}

We studied how action supervision reshapes inherited multimodal representations in vision-language-action models.
Starting from a direct-fusion baseline, we showed that action-loss updates can form action-facing distinctions, but can also induce broad upstream rewriting and unstable instruction-to-visual attention.
To address this tension, we introduced Action QFormer, a query-based action-facing interface that mediates how inherited multimodal information is reorganized before downstream action generation.
Across zero-shot sim-to-real navigation experiments, Action QFormer improves instruction stability, instruction-conditioned action execution, and robustness to visual variation, especially for behaviors requiring directional control and object grounding.

While this work mainly uses navigation as a challenging embodied setting for studying representation stability, future work can extend Action QFormer to manipulation, loco-manipulation, and humanoid control, where contact dynamics, whole-body coordination, and high-dimensional action spaces may place stronger demands on action-compatible multimodal representations.

More broadly, these results suggest that reliable embodied behavior depends not only on representation quality, but also on how multimodal information is selected, organized, and reshaped under action supervision.
These findings highlight action-supervised representation shaping as an important lens for understanding and improving robustness in VLA systems.

\section*{Acknowledgments}
This work was in part supported by Shanghai Qizhi Institute. This work was also in part supported by the InnoHK initiative of the Innovation and Technology Commission of the Hong Kong Special Administrative Region Government via the Hong Kong Centre for Logistics Robotics. We thank Jingzhi Cui, Yunfei Li, Shusheng Xu, and Chao Yu for insightful discussions.

\bibliographystyle{IEEEtran}
\bibliography{references_short}

\appendices
\clearpage
\newpage
\section*{Supplementary Material Overview}

This supplementary material provides additional implementation, experimental, and mechanistic analysis details.
Section~A describes training and implementation details.
Section~B provides additional experimental settings and results.
Section~C reports additional representation and attention analyses supporting the mechanism-level findings in the main paper.

\section{Training and Data Curation Details}
\label{app:training_data}

\subsection{Training Pipeline and Model Comparison}

All main experiments use the same instruction-training setup and the same ObjectNav training dataset.
At a high level, the training pipeline consists of three stages:
a pretrained vision-language backbone first produces inherited multimodal representations from the current observation and task-conditioned language input;
an intermediate interface module reorganizes these representations into a compact action-facing representation;
and a downstream policy head predicts the future navigation trajectory from the projected interface output.

In the main experiments, the pretrained multimodal backbone is Qwen2.5-VL.
The baseline and Action QFormer differ only in the intermediate interface module:
the direct-fusion baseline uses a self-attention-pooling-based latent-fuser module to fuse inherited multimodal representations into an action-facing latent, whereas Action QFormer replaces this module with a lightweight query-based interface for reorganizing inherited image-side and instruction-side features.
The projector, action target, and downstream policy architecture are otherwise kept unchanged across the main comparison.

The following subsections provide the main implementation details of this pipeline, including the training data source, language supervision, sequence construction, parsing procedure, joint language--action supervision, and optimization settings.

\subsection{Training Data Source and Action Representation}

All main experiments are trained on the same ObjectNav training dataset.
We do not mix in additional navigation datasets for the main comparison.
Each training sample is constructed at the timestamp level and contains three components:
(1) the current observation image,
(2) the future action label associated with that timestamp, and
(3) a GPT-generated navigation instruction used for instruction-generation supervision.

Importantly, the main training setting uses only the \emph{current single-frame observation} and does not include historical visual context.
This design keeps the representation problem focused:
the model must organize the currently available visual and language information into a form suitable for downstream control, without relying on temporal context accumulation.

The downstream action supervision is given as a future navigation trajectory represented in the local coordinate frame of the current timestamp.
For each sample, the model predicts the next $8$ future action steps.
Each individual action step is represented by a 4-dimensional tuple
$(x, y, \mathrm{yaw}_{dx}, \mathrm{yaw}_{dy})$,
where $(x, y)$ denotes the local relative position and $(\mathrm{yaw}_{dx}, \mathrm{yaw}_{dy})$ denotes the heading direction in vector form.

\subsection{Language Supervision and Data Curation Principle}

In addition to action supervision, each timestamp is paired with a GPT-generated navigation instruction.
The instruction labels are generated from two inputs:
(1) the current single-frame observation image, and
(2) the corresponding future 8-step action trajectory in the local frame.
The prompt is manually designed to produce exactly one concise navigation command rather than free-form text.

The GPT-based instruction generation prompt contains three key components:
(1) a role definition that asks the model to translate the current view and future trajectory into a single indoor navigation command,
(2) explicit trajectory-interpretation rules that map local-frame position changes to motion semantics such as turning, veering, backward movement, and stopping, and
(3) a constrained output space consisting of two template families:
\emph{object-oriented} templates and \emph{non-object-oriented} templates.

The object-oriented templates encourage the instruction to ground motion in visible landmarks, pathways, intersections, or doorways whenever a near-field and unambiguous goal is available.
The non-object-oriented templates are used for trajectory patterns better described by pure motion semantics, such as sharp turns, arcs, turn-around behavior, or stop commands.
This constrained generation process is not fully deterministic:
multiple valid template instantiations and lexical realizations are allowed for similar trajectories.
As a result, the generated labels retain controlled linguistic diversity while remaining grounded, concise, and geometrically consistent with the future control target.

\begin{quote}
\small
\textbf{Input:} current observation image + future 8-step local-frame trajectory. \\
\textbf{Output:} exactly one concise navigation instruction. \\
\textbf{Template families:}
\begin{itemize}
    \item \texttt{FORMAT\_OBJECT}: \emph{move towards \{goal\}}, \emph{follow \{pathway\}}, \emph{go through \{doorway\}}, \emph{go around \{obstacle\}}, ...
    \item \texttt{FORMAT\_NO\_OBJECT}: \emph{move forward}, \emph{veer slightly left/right}, \emph{turn left/right}, \emph{make a sharp left/right turn}, \emph{arc left/right}, \emph{turn around}, \emph{stop}
\end{itemize}
\end{quote}

\subsection{Sequence Construction, Parsing, and Joint Training}
\label{app:sequence_parsing_training}

This subsection explains how each timestamp-level sample is converted into a multimodal training sequence, how the parsed representations used by the intermediate interface module are extracted, and how language-side and action-side supervision are jointly applied during training.

\subsubsection{Sequence Construction and Data Flow}
Each training sample is organized around a \emph{single current-frame observation}, together with its paired language supervision and future action target.
The raw language input is first formatted through a chat-style template that encodes the navigation task context, while the image input is processed jointly through the pretrained multimodal backbone.
When special-token output mode is used, we additionally insert a short generation bridge before the target text to stabilize decoding into the structured instruction format.

If language targets are present, they are appended in teacher-forcing style, with loss applied only on the supervised output span rather than on the prompt prefix.
In parallel, action-side supervision is attached as a separate numeric target constructed from a fixed-horizon future trajectory segment in the local coordinate frame of the current timestamp.
For each sample, future positions are transformed into local-frame action targets and paired with a validity mask that distinguishes real future steps from padded terminal repetitions near the end of a trajectory.
This design keeps sequence construction unified at the sample level while preserving a clean separation between language-side and action-side supervision.

\subsubsection{Instruction Parsing and Representation Extraction}
After the multimodal backbone produces token-level hidden states, we parse them into the components used by the intermediate interface module before action generation.
In the main training setup, this parsing uses the \emph{special-token branch} exclusively.
Concretely, the sequence contains explicit boundary markers for instruction and visual spans, so that the relevant components can be extracted through a fixed and reproducible parsing rule.

The parser extracts instruction-side and image-side token groups and packages them into a standardized \texttt{parse\_pack}.
Both the baseline latent fuser and Action QFormer consume this same parsed representation interface rather than re-implementing span parsing internally.
As a result, differences between the direct-fusion baseline and Action QFormer arise from how they reorganize inherited multimodal information after parsing, rather than from different token-selection logic.

The parser also supports selective gradient detachment for chosen components, such as instruction-side or image-side representations.
These options are used in the gradient-routing analyses to control which parsed components remain trainable under action-loss updates while keeping the parsing procedure itself fixed.

\subsubsection{Joint Language and Action Supervision}
Training combines language-side supervision and action-side supervision in a single forward pass.
On the language side, the pretrained multimodal backbone is trained under teacher forcing to generate the structured instruction target sequence, with loss computed only on supervised target positions after the prompt prefix.
In our implementation, the language-side loss is taken directly from the backbone, and can be further augmented by optional format-focused and margin-based terms.
When special-token parsing is used, the format term upweights the instruction-boundary tokens, while the margin term explicitly stabilizes the prediction of the instruction-start token against competing vocabulary items.

On the action side, the intermediate interface module first converts the parsed multimodal hidden states into a compact action-facing representation, which is then projected into the policy input space and passed to the downstream diffusion policy head.
The diffusion policy head is trained with a denoising objective over future actions:
Gaussian noise is added to the ground-truth future trajectory, the model predicts the corresponding noise residual, and the resulting per-step mean-squared error is averaged under the valid-step action mask.
When progress-to-go supervision is enabled, an additional masked regression loss is applied to a lightweight progress head, and this term is added to the action denoising loss with a separate weight.

The total training objective is controlled by a scalar language-loss weight $r \in [0,1]$, implemented as \texttt{r\_loss\_weight} in the model.
Let $\mathcal{L}_{\mathrm{lang}}$ denote the language-side loss after optional format and margin terms are added, and let $\mathcal{L}_{\mathrm{action}}$ denote the action-side loss returned by the diffusion policy head.
The overall loss is then
\[
\mathcal{L} =
\begin{cases}
\mathcal{L}_{\mathrm{action}}, & r = 0, \\
\mathcal{L}_{\mathrm{lang}}, & r = 1, \\
(1-r)\,\mathcal{L}_{\mathrm{action}} + r\,\mathcal{L}_{\mathrm{lang}}, & 0 < r < 1.
\end{cases}
\]
This implementation differs slightly from a generic weighted sum:
when $r=1$, the action branch forward pass is skipped entirely and training becomes language-only;
when $r=0$, the language-side loss is ignored and training becomes action-only;
and when $0<r<1$, both branches are evaluated and combined as a convex interpolation.
This design makes the coupling between language-side supervision and action-side supervision explicit, while still allowing controlled experiments in which one branch is emphasized, suppressed, or scheduled over training.

\subsubsection{Loss Weight Scheduling}
In addition to standard optimizer-side scheduling, our training pipeline also supports \emph{loss-side scheduling} for balancing language-side and action-side supervision over training.
In the main instruction-training setup, the language-side loss weight is dynamically updated during optimization rather than kept fixed.
Concretely, the scalar \texttt{r\_loss\_weight}, which controls the interpolation between language-side and action-side supervision in the joint objective, is updated by a dedicated training callback as a function of the current global step.
The scheduling framework is defined over the total number of optimization steps and supports both linear and sigmoid-shaped schedules.
This makes it possible to gradually shift the effective emphasis between structured instruction generation and downstream action learning during training, rather than assuming that a single fixed weighting is optimal throughout.

The same callback mechanism is also available for auxiliary language-side terms.
In particular, the codebase supports dynamic scheduling for the overall format-loss weight, the start/end token weights used inside the format-focused loss, and the weight and margin value of the start-token margin objective.
In the main instruction-training configuration, these scheduling options are enabled together with language-loss scheduling, so that the contribution of format-sensitive supervision can evolve with training progress rather than remaining static.
All such updates are applied directly on the model at step boundaries through training callbacks, which keeps the optimization recipe synchronized across the direct-fusion baseline and Action QFormer variants.

\subsection{Optimization Details}

Within the common training framework described above, all main experiments use the same optimizer and training schedule family across the direct-fusion baseline and Action QFormer variants.
In particular, the main runs share the same learning rate, scheduler type, warmup ratio, gradient clipping rule, and overall training horizon.
This keeps the optimization setting aligned across model families, so that the main comparison remains focused on representation structure rather than training recipe differences.

Concretely, we use cosine learning-rate scheduling with warmup ratio $0.05$, gradient clipping with max norm $0.5$, a learning rate of $2\times10^{-4}$, and a maximum of $6000$ optimization steps.
Input images are resized with a maximum pixel budget of $200{,}000$.
The direct-fusion baseline and Action QFormer models use different batch configurations due to their different memory footprints, while keeping the same overall optimization family and training horizon.
For the main direct-fusion baseline, we use a batch configuration of $16 \times 4$.
For the main Action QFormer run, we use a batch configuration of $8 \times 8$.
The two batch configurations have the same effective batch size. In both cases, the downstream policy architecture and action supervision remain matched.
The main training hyperparameters are summarized in Table~\ref{tab:app_train_hparams}.

\begin{table}[!t]
    \centering
    \caption{Main training hyperparameters for the instruction-training setup.}
    \label{tab:app_train_hparams}
    \small
    \begin{tabular}{lc}
        \toprule
        Hyperparameter & Value \\
        \midrule
        Training dataset & ObjectNav training dataset \\
        Observation context & Single current frame only \\
        Image resizing & max pixels = 200,000 \\
        Action horizon & 8 steps \\
        Action representation & $(x, y, \mathrm{yaw}_{dx}, \mathrm{yaw}_{dy})$ \\
        Scheduler & Cosine \\
        Warmup ratio & 0.05 \\
        Learning rate & $2\times10^{-4}$ \\
        Max grad norm & 0.5 \\
        Max training steps & 6000 \\
        Direct-fusion batch setup & 16 $\times$ 4 \\
        Action QFormer batch setup & 8 $\times$ 8 \\
        Attention implementation & FlashAttention 2 \\
        \bottomrule
    \end{tabular}
\end{table}

\subsection{Action QFormer Configuration}

Unless otherwise noted, Action QFormer uses a fixed default configuration across the main experiments.
Inspired by the query-interface design of BLIP-2 and InstructBLIP, the module uses a set of learnable query tokens that interact with instruction-side representations through self-attention and extract control-relevant visual evidence from inherited image-side features through cross-attention~\cite{dai2023instructblip}.
The updated queries form compact action-facing representations that are passed to the downstream policy head together with the instruction representation.

In the main configuration, Action QFormer uses $16$ learnable query tokens and $4$ transformer layers.
We do not perform extensive module-specific hyperparameter tuning.
The downstream projector, policy head, action target, and training objective are kept matched with the direct-fusion baseline.
This choice keeps the comparison focused on the role of the structured action-facing interface itself rather than on aggressive architectural search.

\section{Experiment Details}
\label{app:experiment_details}

The raw counts corresponding to the percentage results in Sec.~\ref{sec:closed_loop_eval} are listed in Table~\ref{tab:app_closeloop_full_raw}.

\begin{table*}[!t]
    \centering
    \caption{
    Raw-count closed-loop sim-to-real results in challenging ObjectNav scenarios.
    }
    \label{tab:app_closeloop_full_raw}
    \footnotesize
    \setlength{\tabcolsep}{3.2pt}
    \renewcommand{\arraystretch}{1.08}
    \begin{tabular}{@{}>{\centering\arraybackslash}m{0.12\textwidth}lccccc|cc@{}}
        \toprule
        & & \multicolumn{5}{c|}{Without provided instructions} & \multicolumn{2}{c}{Given instructions} \\
        \cmidrule(r){3-7} \cmidrule(r){8-9}
        \multicolumn{1}{c}{Model}
        & \multicolumn{1}{c}{Scenario}
        & \makecell{Instruction\\Direction $\uparrow$}
        & \makecell{Instruction\\OOD Rate $\downarrow$}
        & \makecell{Action\\Direction $\uparrow$}
        & \makecell{Number of\\Collisions $\downarrow$}
        & \makecell{Task\\Success $\uparrow$}
        & \makecell{Number of\\Collisions $\downarrow$}
        & \makecell{Task\\Success $\uparrow$} \\
        \midrule
        \multirow{4}{*}{\makecell[c]{Direct Fusion}}
        & Fridge (far target) & 49/61 & 61/61 & 40/61 & 4 & 2/8 & 2 & 6/8 \\
        & Door (obstacle) & 27/36 & 35/36 & 26/36 & 14 & 2/8 & 14 & 7/8 \\
        & Chair (turn-around) & 13/28 & 28/28 & 15/28 & 5 & 2/8 & 6 & 2/8 \\
        & Table (sharp turn) & 11/48 & 48/48 & 8/48 & 6 & 0/8 & 5 & 2/8 \\
        \midrule
        \multirow{4}{*}{\makecell[c]{Action QFormer}}
        & Fridge (far target) & 70/73 & 0/73 & 68/73 & 5 & 3/8 & 0 & 8/8 \\
        & Door (obstacle) & 41/42 & 0/42 & 38/42 & 12 & 5/8 & 3 & 7/8 \\
        & Chair (turn-around) & 20/34 & 0/34 & 28/34 & 4 & 5/8 & 1 & 7/8 \\
        & Table (sharp turn) & 57/57 & 0/57 & 53/57 & 3 & 5/8 & 1 & 7/8 \\
        \bottomrule
    \end{tabular}
    \vspace{-1.0em}
\end{table*}

\subsection{Probe Scenes and Evaluation Rules for Action Generation}

For the action-generation probes in the main text, we use representative real-scene observations and evaluate repeated stochastic action predictions under fixed instructions.
Each image--instruction pair is sampled 8 times, and we report three complementary metrics: \textbf{correctness}, \textbf{collision}, and \textbf{quality}.
Correctness measures whether the predicted trajectory satisfies the intended control behavior of the instruction; collision measures whether the trajectory enters scene-specific collision-prone regions; and quality measures trajectory smoothness.

We use three probe families.
For \textbf{directional control}, the probe scene contains a bifurcating layout with a right-front curved wall, and correctness is defined by whether the trajectory commits to the intended left/right side under the given directional instruction.
For \textbf{object grounding}, the probe scenes contain visually specified targets embedded in cluttered indoor layouts, and correctness is defined by whether the trajectory reaches a manually specified endpoint region corresponding to the target object.
For \textbf{motion templates}, the probe scene is a hallway view, and correctness is defined by whether the predicted trajectory exhibits the intended template-compatible forward traversal behavior.

Collision is evaluated using manually specified forbidden regions that approximate wall-adjacent or obstacle-prone areas in each probe scene.
Quality is measured by a trajectory smoothness score, averaged across repeated samples for each image--instruction pair.

\begin{figure*}[!t]
    \centering

    \includegraphics[
        width=\textwidth
    ]{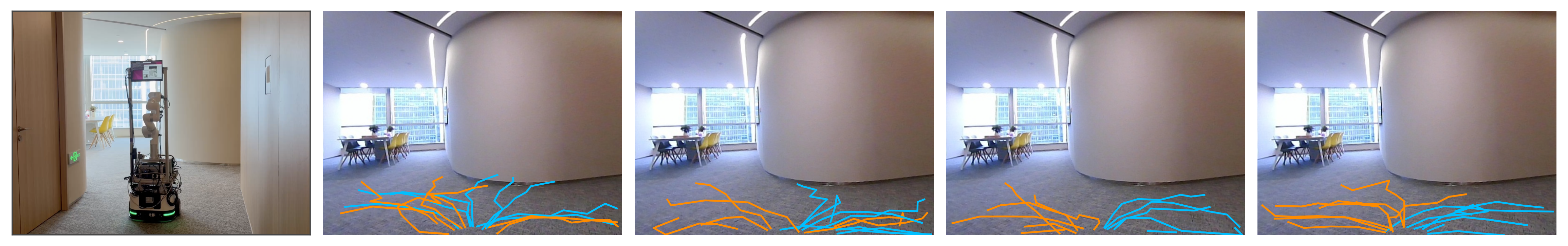}\\[-0.8em]
    {\footnotesize
    \textbf{(a) Testing scenario: directional control.}
    \textit{Example instructions: ``make a left turn'' and ``make a right turn''.}
    }

    \includegraphics[
        width=\textwidth
    ]{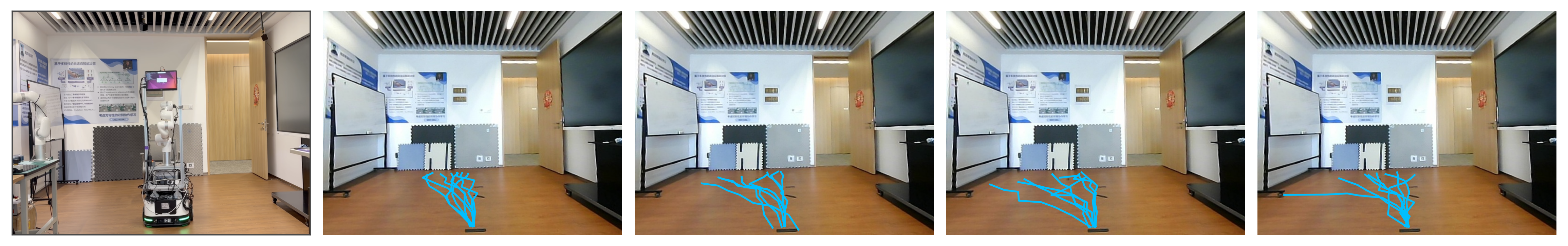}\\[-0.8em]
    {\footnotesize
    \textbf{(b) Testing scenario: object grounding.}
    \textit{Example instruction: ``move towards the whiteboard in the corner''.}
    }

    \includegraphics[
        width=\textwidth
    ]{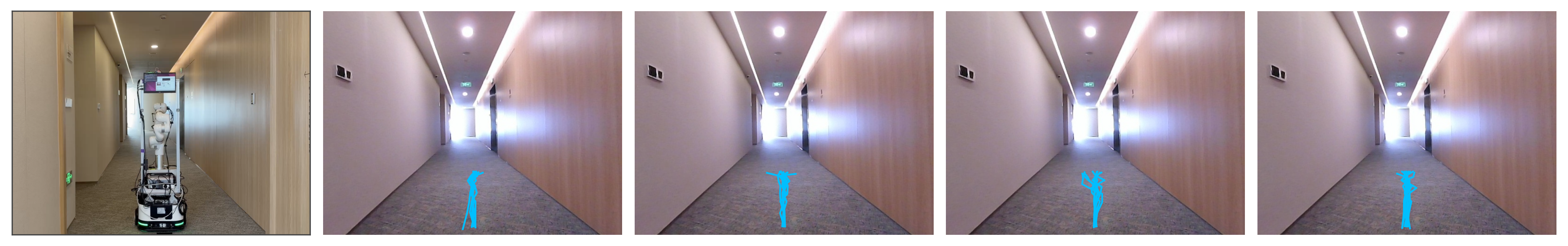}\\[-0.8em]
    {\footnotesize
    \textbf{(c) Testing scenario: motion template.}
    \textit{Example instruction: ``go through the hallway''.}
    }

    \vspace{-0.5em}
    \caption{
    \textbf{Representative qualitative comparisons for interface-capacity ablations under fixed-instruction action generation.}
    Each row shows one testing scenario, with the third-person real scene on the left and action predictions from the same first-person observation and fixed instruction in the remaining columns.
    Prediction columns compare the \textbf{Direct-fusion baseline}, \textbf{Reduced-depth Action QFormer}, \textbf{Reduced-query Action QFormer}, and \textbf{Action QFormer}.
    Thin trajectories show multiple inference samples from the same model, and the bold trajectory shows the selected action trajectory used for comparison.
    }
    \label{fig:action_generation_capacity_qualitative}
    \vspace{-0.8em}
\end{figure*}

\subsection{Additional Interface Ablations on Action Generation}

Beyond the direct-fusion baseline and the full Action QFormer, we evaluate two capacity-reduced variants:
a reduced-depth Action QFormer and a reduced-query Action QFormer.
This allows us to test whether the action-generation gain comes merely from inserting a structured intermediate interface, or whether the capacity of that interface also matters for organizing inherited multimodal information into an action-compatible representation. The corresponding qualitative comparisons are shown in Fig.~\ref{fig:action_generation_capacity_qualitative}.

Both reduced variants already improve over the direct-fusion baseline on the more diagnostic probe families, indicating that the benefit of Action QFormer does not depend on a single full-capacity configuration.
At the same time, the full Action QFormer remains strongest overall, with the clearest advantage on directional control and a consistent competitive edge on object grounding.
This suggests that a structured action-facing interface is already beneficial, but that interface capacity still affects how reliably control-relevant information is organized.

The contrast between the two reduced variants is also informative.
Both reductions retain part of the gain over the direct-fusion baseline, but the reduced-query variant remains consistently stronger than the reduced-depth variant, especially on directional control.
This suggests that, in the current navigation probe setting, the benefit of Action QFormer is more sensitive to depth reduction than to query reduction.

\section{Additional Mechanistic Analysis}
\label{app:mechanistic_analysis}

This appendix provides additional details and supporting evidence for the mechanistic analyses in Sec.~\ref{sec:mechanistic_analysis}.
The goal is to complement the main results rather than introduce separate model families or new evaluation settings.
All analyses use the same two interface families studied in the main text:
the direct-fusion baseline and Action QFormer.

We organize the analyses around how action supervision interacts with the inherited multimodal pathway.
For representation and attention comparisons, each action-loss-exposed setting is compared against an architecture-matched \textbf{action-update-blocked} reference, in which action-loss gradients are prevented from updating the inherited multimodal pathway.
Thus, the reported changes should be read as action-loss-induced changes relative to a matched reference, not as comparisons between unrelated models.

The appendix follows the same structure as the main mechanistic analysis.
We first define the gradient-routing settings used for the architecture-matched comparisons.
We then provide additional evidence for action-facing directional distinctions, upstream token rewriting and targeted adaptation, and instruction-to-visual attention stability.
Together, these analyses support the main claim that Action QFormer changes where and how action supervision reshapes inherited multimodal representations:
it preserves constructive action-supervised adaptation while reducing broad upstream rewriting and instruction-to-visual attention shifts.

\subsection{Gradient-Routing Settings}
\label{app:mechanistic_settings}

Our mechanistic analyses are organized around how action-loss gradients interact with the inherited multimodal pathway.
The goal is not only to determine whether representations change under action supervision, but also to characterize \emph{how} they change:
whether behaviorally consequential distinctions become clearer,
whether upstream token rewriting is broad or targeted,
and whether instruction-to-visual attention remains stable.

\subsubsection{Interface Families}
We compare two interface families throughout the mechanistic analyses.
The \textbf{direct-fusion baseline} uses the minimal shared-context interface described in Sec.~\ref{sec:methodology}, in which inherited image-side and instruction-side representations are fused directly into an action-facing representation.
The \textbf{Action QFormer} family replaces direct fusion with a query-based action-facing interface that reorganizes inherited image-side and instruction-side representations into an action-facing representation before downstream policy prediction.
Across all variants, both families share the same pretrained Qwen2.5-VL backbone, downstream action target, and policy head; the main architectural difference lies in how the action-facing representation is formed.

\subsubsection{Gradient-Routing Settings}
Within each interface family, we use four descriptive gradient-routing settings.
These settings differ in which parts of the inherited multimodal pathway are exposed to action-loss gradients.

\textbf{Action-update-blocked} prevents action-loss gradients from updating the inherited multimodal pathway and serves as the reference setting for representation and attention comparisons.
\textbf{Vision-encoder-frozen} keeps the vision encoder fixed while allowing action-loss gradients to update the language backbone and downstream interface modules.
\textbf{Language-backbone-blocked} prevents action-loss gradients from updating the language backbone while allowing the vision encoder and downstream interface modules to remain exposed to the action objective.
\textbf{Full-update} allows action-loss gradients to update both the language-side and vision-side pathways, corresponding to the strongest action-loss exposure among the settings considered.

Together, these settings let us separate two effects:
whether action-loss exposure is needed to form behaviorally consequential distinctions, and whether stronger exposure also induces broader upstream rewriting or less stable instruction-to-visual attention.

\subsubsection{Pairwise Comparison Protocol}
Mechanistic quantities such as token-wise rewriting, rewriting-subspace statistics, and attention drift are computed by comparing each action-loss-exposed setting against the \textbf{action-update-blocked} reference within the same interface family.
Each comparison therefore measures how a given gradient-routing setting changes representations or attention relative to an architecture-matched reference in which upstream action-loss updates are blocked.

We consider three main comparisons:
\textbf{language-backbone-blocked vs.\ action-update-blocked}, which measures changes induced when the language backbone is protected while the vision encoder and downstream interface modules remain trainable under the action objective;
\textbf{vision-encoder-frozen vs.\ action-update-blocked}, which measures changes induced when the vision encoder is fixed while the language backbone and downstream interface modules remain trainable under the action objective;
and \textbf{full-update vs.\ action-update-blocked}, which measures changes induced by joint action-loss exposure of the language-side and vision-side pathways.
These comparisons should be interpreted as \emph{changes relative to an action-update-blocked reference}, not as comparisons between independent model families.

\subsubsection{Use in Mechanistic Analyses}
Directional-distinction analysis uses these settings to test whether behaviorally consequential distinctions are formed in the action-facing representation.
Token-wise rewriting and rewriting-subspace analyses examine how strongly the inherited pathway is rewritten and whether that rewriting is broad or targeted.
Attention analyses examine whether stronger action-supervised shaping destabilizes or preserves instruction-to-visual attention.
Together, these settings provide a controlled framework for studying how Action QFormer changes where and how action supervision reshapes the inherited multimodal pathway.

\subsection{Additional Evidence for Upstream Token Rewriting and Targeted Adaptation}
\label{app:drift_targeted_adaptation}

The main text presents the primary evidence for upstream token rewriting through rewriting magnitude, active-dimension fraction, and focused control- and spatial-token visualizations.
Here we provide complementary numerical summaries and structural statistics for the same contrast between broad upstream rewriting and targeted adaptation.
These results are intended to support, rather than replace, the main visual analyses:
they summarize token-level rewriting, rewriting-subspace statistics, rewrite-share allocation across token groups, and active-dimension usage under the same architecture-matched gradient-routing comparisons.

The goal of this section is to distinguish broad upstream rewriting from targeted adaptation.
Broad upstream rewriting refers to action-loss-induced changes that spread across many tokens and hidden dimensions in the inherited multimodal pathway.
Targeted adaptation refers to a more selective pattern in which rewriting is less broadly distributed, while meaningful changes remain concentrated on selected control and spatial tokens or limited representation subspaces.

\subsubsection{Token-Level Rewriting Statistics}
\label{app:rewriting_summary_stats}

We first provide compact statistics that quantify the upstream token-rewriting trends summarized in the main text.
These statistics are supporting measurements:
they are not intended to replace the token-level visualizations in Sec.~\ref{sec:upstream_rewriting}, but instead provide complementary comparisons across gradient-routing settings.

Table~\ref{tab:app_tokenwise_stats} reports token-level cosine-similarity and $L_2$-distance statistics relative to the architecture-matched reference.
Across comparable gradient-routing settings, Action QFormer generally reduces token-level $L_2$ rewriting magnitude relative to the direct-fusion baseline, with the clearest gap appearing in the \textbf{full-update comparison}.
This is consistent with the main-text heatmaps, where the direct-fusion baseline exhibits stronger broad upstream rewriting under the strongest action-loss exposure setting.

Table~\ref{tab:app_subspace_stats} summarizes the subspace structure of token rewriting, including mean-squared drift energy, effective rank, and top-$k$ variance concentration.
In the \textbf{full-update comparison}, Action QFormer exhibits lower mean-squared drift energy than the direct-fusion baseline, together with substantially higher effective rank and lower top-$k$ variance concentration.
This suggests that action-loss-induced rewriting under Action QFormer is less dominated by a small number of principal directions.
We treat these rewriting-subspace statistics as supporting evidence for the token-level analysis:
they reinforce the view that direct fusion produces more directionally concentrated upstream rewriting, whereas Action QFormer distributes the remaining rewriting across a less concentrated subspace.

\begin{table}[!t]
    \centering
    \caption{
    Token-level cosine-similarity and $L_2$-distance statistics across pairwise gradient-routing comparisons.
    }
    \vspace{-0.5em}
    \label{tab:app_tokenwise_stats}
    \scriptsize
    \setlength{\tabcolsep}{3.5pt}
    \renewcommand{\arraystretch}{1.05}
    \begin{tabular}{llcccc}
        \toprule
        Interface & Comparison & Cosine Mean & Cosine Std & L2 Mean & L2 Std \\
        \midrule
        Direct-fusion & Vision frozen  & 0.934 & 0.089 & 65.76 & 38.15 \\
        Direct-fusion & Lang. blocked  & 0.946 & 0.067 & 74.17 & 29.32 \\
        Direct-fusion & Full update & 0.913 & 0.083 & 93.90 & 33.82 \\
        \midrule
        Action QFormer & Vision frozen  & 0.940 & 0.089 & 56.80 & 40.41 \\
        Action QFormer & Lang. blocked  & 0.950 & 0.060 & 57.03 & 34.03 \\
        Action QFormer & Full update & 0.913 & 0.128 & 75.18 & 52.15 \\
        \bottomrule
    \end{tabular}

    \vspace{0.3em}
    \begin{minipage}{0.95\linewidth}
        \scriptsize
        \emph{Note:} All comparisons are computed relative to the action-update-blocked reference within the same interface family.
    \end{minipage}
    \vspace{-1.5em}
\end{table}

\begin{table}[!t]
    \centering
    \caption{
    Rewriting-subspace statistics across pairwise gradient-routing comparisons.
    }
    \vspace{-0.5em}
    \label{tab:app_subspace_stats}
    \scriptsize
    \setlength{\tabcolsep}{3.2pt}
    \renewcommand{\arraystretch}{1.05}
    \begin{tabular}{llcccc}
        \toprule
        Interface & Comparison & Mean Sq. & Eff. Rank & Top-5 Var. & Top-10 Var. \\
        \midrule
        Direct-fusion & Vision frozen  & 5779.49 & 22.86 & 0.697 & 0.765 \\
        Direct-fusion & Lang. blocked  & 6360.11 & 33.02 & 0.646 & 0.721 \\
        Direct-fusion & Full update & 9960.78 & 29.41 & 0.666 & 0.747 \\
        \midrule
        Action QFormer & Vision frozen  & 4859.10 & 40.08 & 0.592 & 0.674 \\
        Action QFormer & Lang. blocked  & 4410.74 & 36.83 & 0.588 & 0.664 \\
        Action QFormer & Full update & 8371.07 & 65.68 & 0.521 & 0.610 \\
        \bottomrule
    \end{tabular}

    \vspace{0.3em}
    \begin{minipage}{0.95\linewidth}
        \scriptsize
        \emph{Note:} All comparisons are computed relative to the action-update-blocked reference within the same interface family.
    \end{minipage}
    \vspace{-1.5em}
\end{table}

\subsubsection{Token Rewrite Share Across Token Groups}
\label{app:token_rewrite_share}

We further analyze how total token-level rewriting is allocated across token groups.
We partition tokens into boundary, control, spatial, and other groups, and report each group’s share of the total rewriting budget in Table~\ref{tab:app_token_rewrite_share}.
This group-level view complements the focused control- and spatial-token visualization in Sec.~\ref{sec:upstream_rewriting}:
rather than showing which individual tokens are rewritten, it asks how total rewriting is distributed across semantically meaningful token groups.

In the \textbf{full-update comparison}, the direct-fusion baseline places a substantially larger fraction of the total rewriting budget on control and spatial token groups.
Action QFormer reduces this group-level allocation, while the focused heatmap in the main text shows that meaningful rewriting is still retained on a smaller subset of key control and spatial tokens.
Together, these results support the targeted-adaptation interpretation:
Action QFormer does not eliminate action-loss-induced rewriting, but reduces broad upstream rewriting while preserving more focused adaptation.

\begin{table}[!t]
    \centering
    \caption{
    Token rewrite share across token groups.
    }
    \vspace{-0.5em}
    \label{tab:app_token_rewrite_share}
    \scriptsize
    \setlength{\tabcolsep}{3.5pt}
    \renewcommand{\arraystretch}{1.05}
    \begin{tabular}{llrrrr}
        \toprule
        Interface & Comparison & Boundary & Control & Spatial & Others \\
        \midrule
        Direct-fusion & Vision frozen  & 0.425 & 0.285 & 0.191 & 0.099 \\
        Direct-fusion & Lang. blocked  & 0.540 & 0.161 & 0.139 & 0.160 \\
        Direct-fusion & Full update & 0.423 & 0.269 & 0.168 & 0.139 \\
        \midrule
        Action QFormer & Vision frozen  & 0.699 & 0.125 & 0.095 & 0.081 \\
        Action QFormer & Lang. blocked  & 0.606 & 0.158 & 0.104 & 0.133 \\
        Action QFormer & Full update & 0.717 & 0.110 & 0.083 & 0.090 \\
        \bottomrule
    \end{tabular}

    \vspace{0.3em}
    \begin{minipage}{0.95\linewidth}
        \scriptsize
        \emph{Note:} All comparisons are computed relative to the action-update-blocked reference within the same interface family.
        Each row reports the fraction of total token-wise rewriting allocated to boundary, control, spatial, and other token groups.
    \end{minipage}
    \vspace{-1.0em}
\end{table}

\subsubsection{Active-Dimension Structure}
\label{app:active_dim_structure}

The preceding analyses show that Action QFormer reduces broad upstream token rewriting.
We next ask whether this selectivity is also reflected in how rewritten dimensions are organized.
Rather than examining individual token-level heatmaps, we summarize active-dimension usage from two complementary perspectives:
how many hidden dimensions change for each token, and how many tokens are affected by each hidden dimension.

\begin{figure}[!t]
    \centering

    \includegraphics[width=\linewidth]{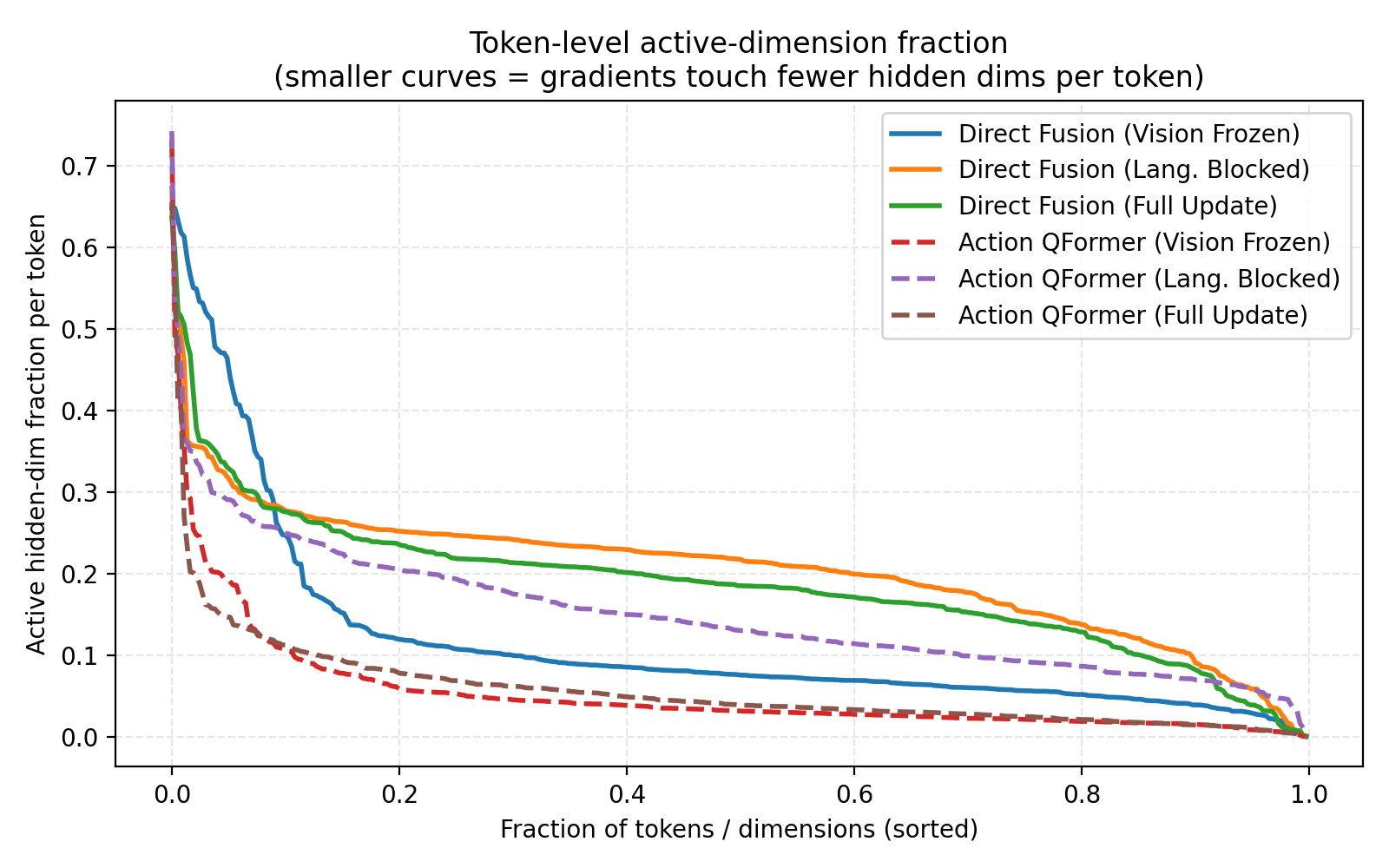}\\[-0.8em]
    {\footnotesize \textbf{(a) Token-level active-dimension fraction.}}

    \vspace{0.4em}

    \includegraphics[width=\linewidth]{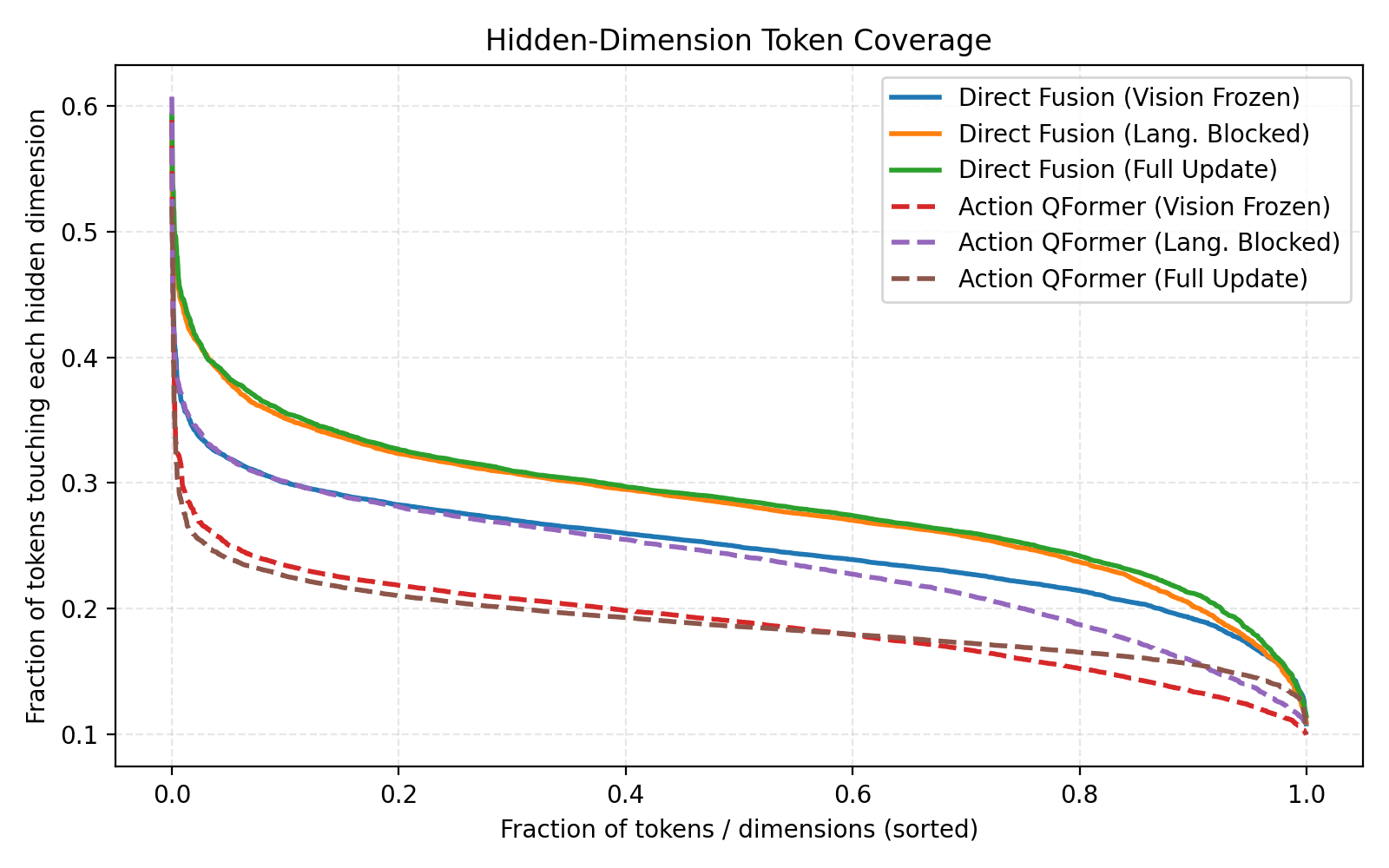}\\[-0.8em]
    {\footnotesize \textbf{(b) Hidden-dimension active-token fraction.}}

    \vspace{-0.3em}
    \caption{
    \textbf{Active-dimension structure across gradient-routing comparisons.}
    \textbf{(a)} Token-level active-dimension fraction measures how many hidden dimensions are significantly rewritten per token; lower curves indicate sparser per-token rewriting.
    \textbf{(b)} Hidden-dimension active-token fraction measures how many tokens are significantly affected by each hidden dimension; lower curves indicate that most hidden dimensions affect fewer tokens.
    }
    \label{fig:app_active_dim_structure}
    \vspace{-0.8em}
\end{figure}

Fig.~\ref{fig:app_active_dim_structure}(a) reports token-level active-dimension fractions across gradient-routing settings.
This view aggregates the token-level active-dimension patterns into sorted curves.
The direct-fusion baseline remains higher across much of the curve, indicating that many tokens are rewritten along a larger number of hidden dimensions.
Action QFormer yields noticeably lower curves, supporting the interpretation that per-token rewriting is more selective.

Fig.~\ref{fig:app_active_dim_structure}(b) presents the complementary hidden-dimension perspective.
Rather than asking how many dimensions rewrite each token, it measures how many tokens are affected by each hidden dimension.
In the direct-fusion baseline, a subset of dimensions becomes active across many tokens, consistent with more globally shared rewriting directions.
Action QFormer yields lower and more compact curves, indicating that most hidden dimensions affect fewer tokens.

Together, these views suggest that Action QFormer makes action-loss-induced rewriting more selective from both the token and hidden-dimension perspectives.
They provide additional evidence that action-supervised shaping is expressed through more localized token--dimension patterns rather than broad rewriting across the inherited multimodal pathway.

\subsection{Additional Evidence for Instruction-to-Visual Attention}
\label{app:attention_behavior}

Sec.~\ref{sec:attention_stability} reports the main instruction-to-visual attention analysis under the full-update comparison.
Here we provide complementary evidence, including the full gradient-exposure sweep, visual-source variants, representative attention-focus cases, selected per-head examples, and auxiliary alignment statistics.
The goal is to characterize two aspects of instruction-to-visual attention under action supervision:
\emph{attention stability}, where Action QFormer reduces high-risk attention shifts,
and \emph{constructive adaptation}, a qualitative and non-universal effect in which action-loss exposure sharpens attention toward task-relevant visual regions in representative cases.

Unless otherwise noted, attention is analyzed at the phrase level using maps from instruction tokens to visual tokens.
Each action-loss-exposed setting is compared against its architecture-matched action-update-blocked reference, allowing us to measure how closely its attention remains aligned with the corresponding reference pattern.

\begin{figure*}[!t]
    \centering
    \includegraphics[
        width=\textwidth,
        trim=0 0 0 120pt,
        clip
    ]{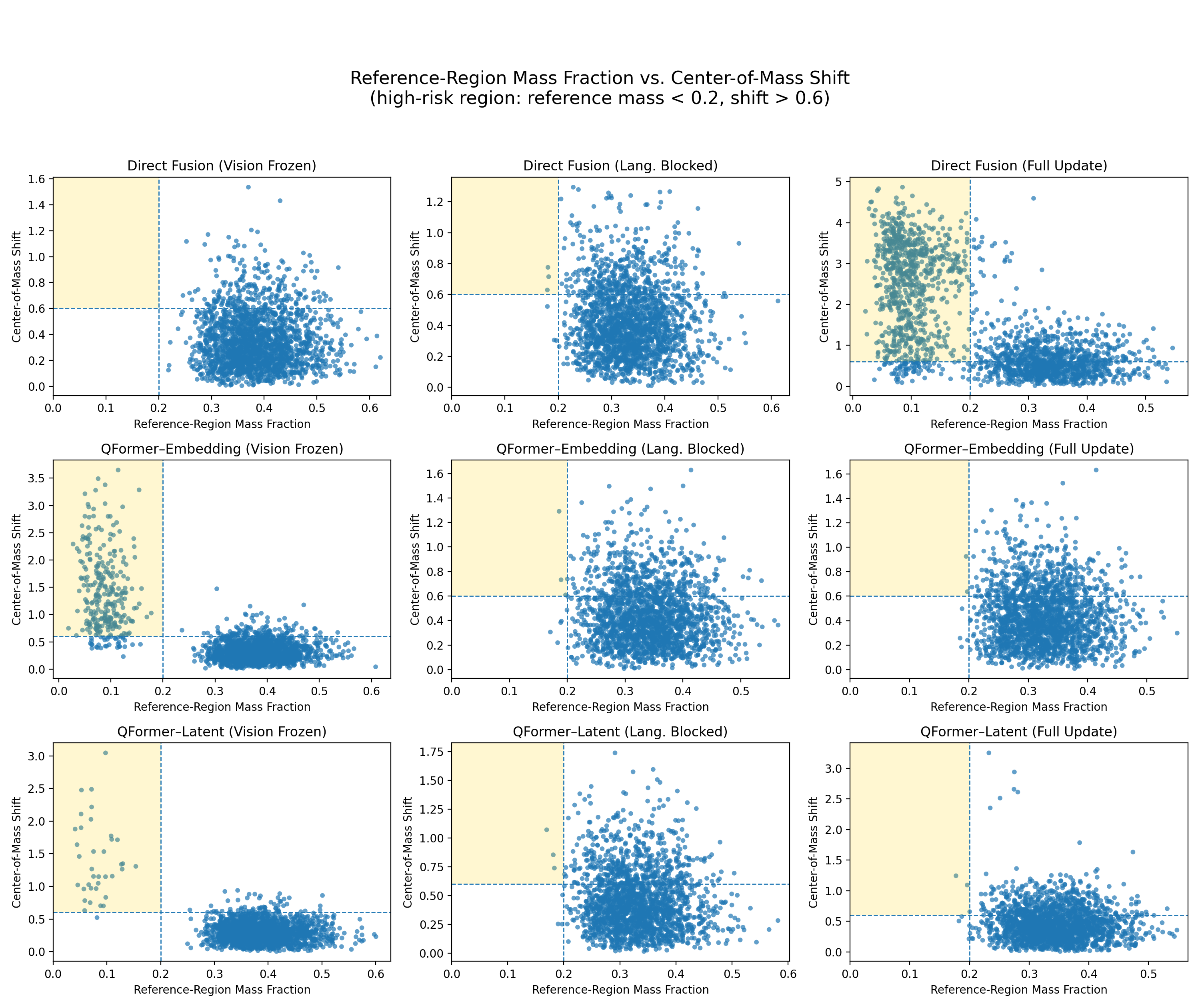}
    \vspace{-2.0em}
    \caption{
    \textbf{Full joint distribution of reference-region mass and spatial drift.}
    The x-axis measures reference-region mass, defined as the fraction of attention from the action-loss-exposed setting that remains inside the action-update-blocked reference's top-$10\%$ attention region.
    The y-axis measures center shift, defined as the center-of-mass distance between the action-loss-exposed and reference attention maps.
    The shaded region indicates high-risk attention changes with low reference-region mass and large spatial drift.
    The strongest baseline attention drift appears in the full-update comparison, whereas Action QFormer shifts the distribution toward higher reference-region mass and smaller center shifts.
    Action QFormer--Embedding uses the vision-encoder output as its visual source, whereas Action QFormer--Latent uses the image-side representation after language-backbone processing.
    }
    \label{fig:app_attention_joint_refmass_shift}
\end{figure*}

\subsubsection{Full Attention-Stability Results Across Gradient-Routing Settings}
\label{app:attention_joint}

The main text focuses on the full-update comparison, where attention drift in the direct-fusion baseline is most pronounced.
In Fig.~\ref{fig:app_attention_joint_refmass_shift}, we report the full gradient-exposure sweep together with two Action QFormer visual-source variants.
The contrast is clearest in the full-update comparison; the remaining settings provide context for how attention changes vary with action-loss exposure rather than indicating a strictly monotonic trend.

We use the same phrase-level attention-stability metrics defined in Sec.~\ref{sec:attention_stability}: reference-region mass and center shift, with low reference-region mass and large center shift indicating a high-risk attention change.
Fig.~\ref{fig:app_attention_joint_refmass_shift} shows the full joint distribution.
In the full-update comparison, the direct-fusion baseline produces many high-risk cases with low reference-region mass and large spatial drift.
Action QFormer shifts the distribution toward higher reference-region mass and smaller center shifts, and this qualitative trend remains visible for both variants.

The \textbf{vision-encoder-frozen comparison} shows a more nuanced pattern:
both Action QFormer variants still contain a noticeable number of shifted or collapsed attention cases.
One possible explanation is that exposing the language backbone and downstream interface modules to action-loss gradients while keeping the vision encoder fixed may create a mismatch between the updated instruction-side pathway and fixed visual representations.
By contrast, the \textbf{full-update comparison} allows the language backbone, vision encoder, and downstream interface modules to adapt jointly; this is also the setting in which baseline attention drift becomes most severe and Action QFormer exhibits its clearest attention-stability advantage.
We treat this explanation as an interpretive hypothesis rather than a definitive causal claim, and use the full sweep primarily to contextualize the full-update comparison in the main text.

\begin{figure*}[!t]
    \centering

    \begin{minipage}[t]{0.49\textwidth}
        \centering
        \includegraphics[
            width=\linewidth,
            trim=0 0 0 120pt,
            clip
        ]{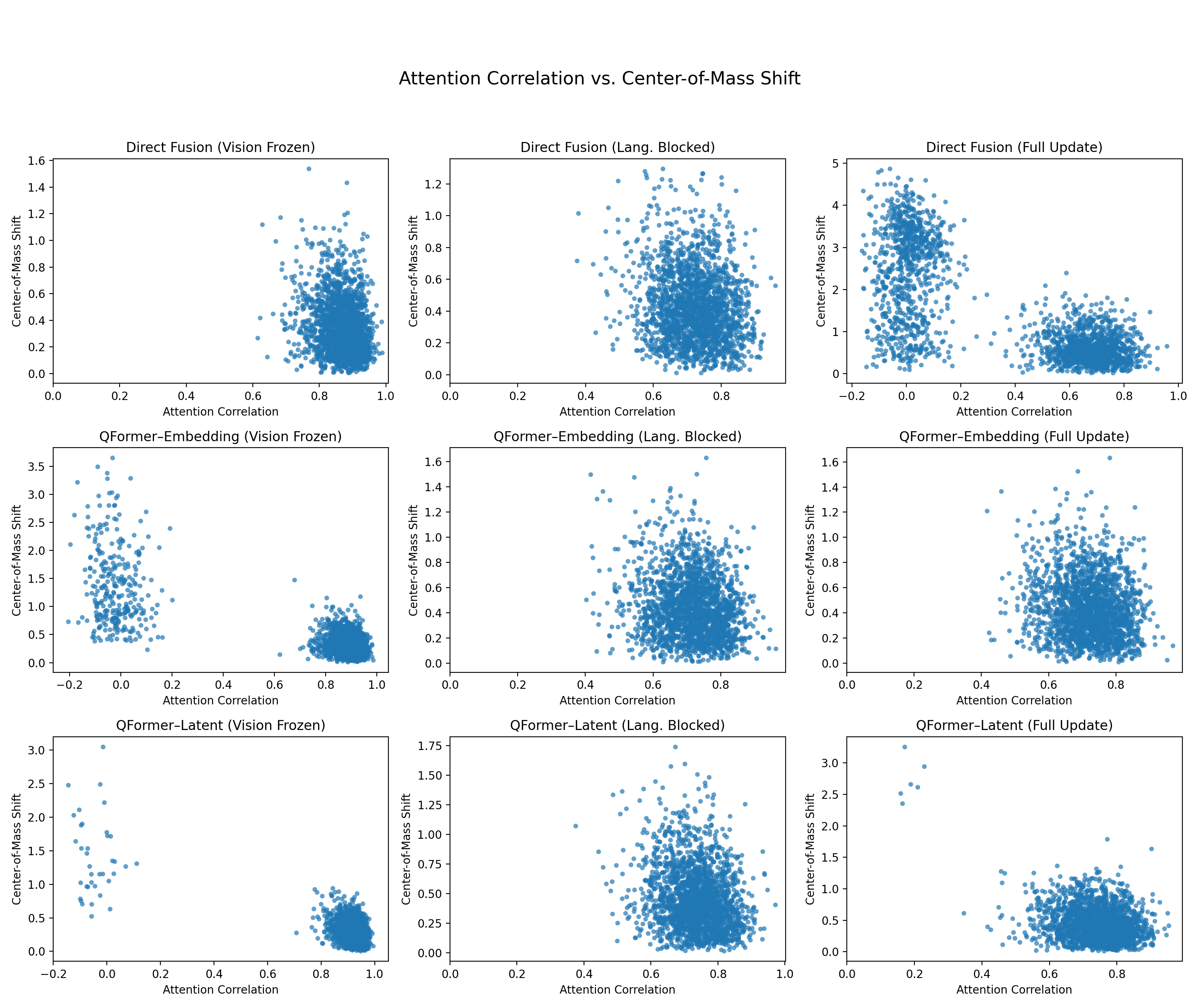}\\[-0.2em]
        {\footnotesize \textbf{(a)} \texttt{corr} vs.\ \texttt{center\_shift}}
    \end{minipage}
    \hfill
    \begin{minipage}[t]{0.49\textwidth}
        \centering
        \includegraphics[
            width=\linewidth,
            trim=0 0 0 120pt,
            clip
        ]{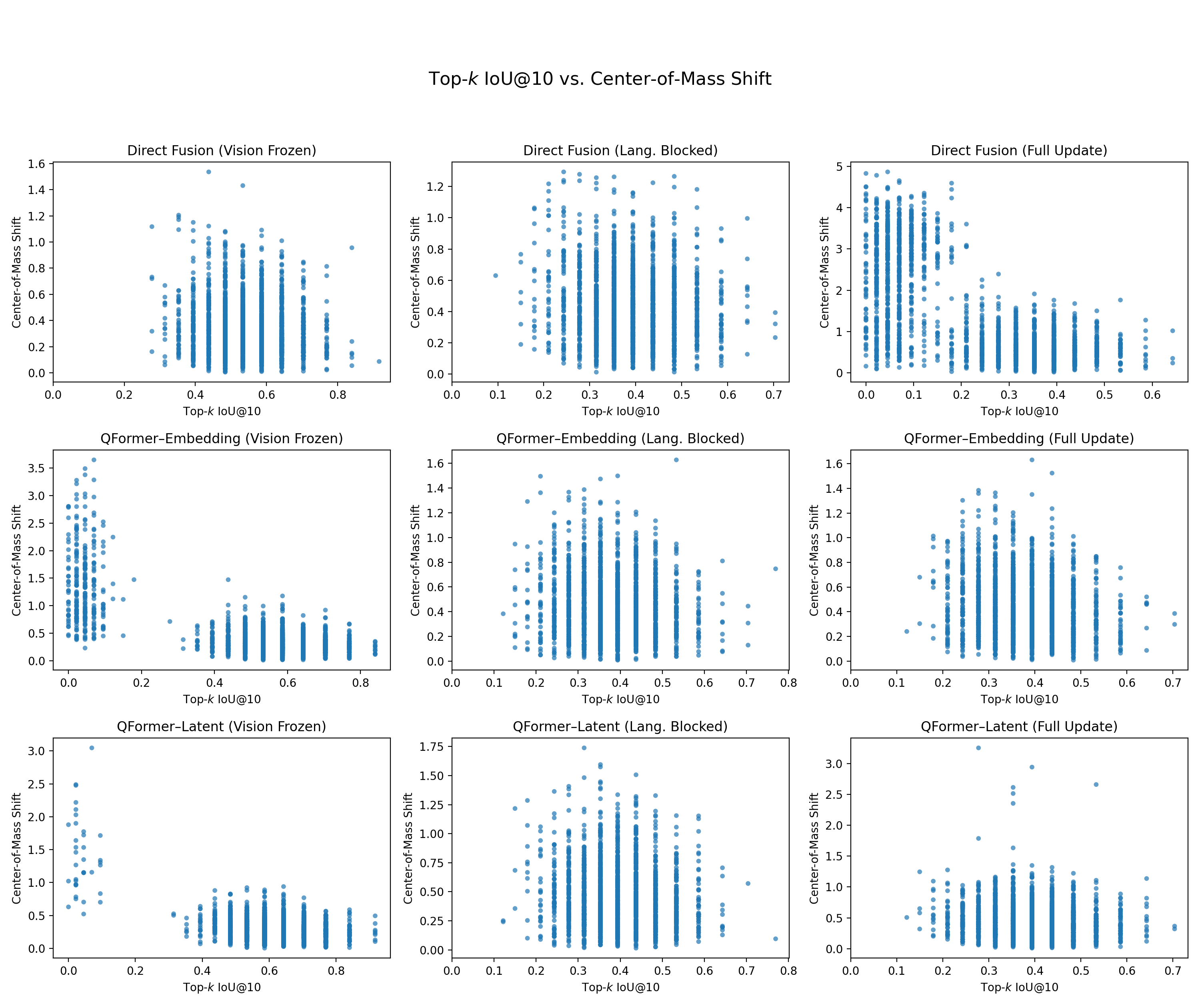}\\[-0.2em]
        {\footnotesize \textbf{(b)} \texttt{topk\_iou\_10} vs.\ \texttt{center\_shift}}
    \end{minipage}

    \caption{
    \textbf{Additional attention-alignment scatter statistics.}
    These supporting plots complement the joint analysis in Fig.~\ref{fig:app_attention_joint_refmass_shift}.
    Across the auxiliary metrics, Action QFormer generally exhibits stronger reference-pattern similarity and smaller spatial shifts than the direct-fusion baseline.
    }
    \label{fig:app_attention_supporting_scatter}
\end{figure*}

\begin{figure*}[!t]
    \centering
    \includegraphics[
        width=0.80\textwidth
    ]{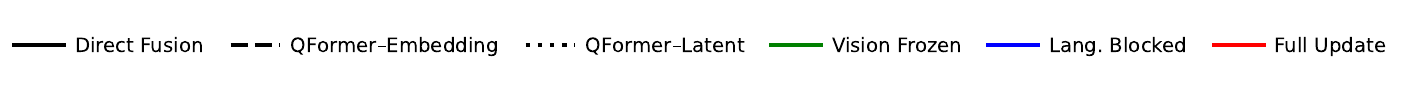}

    \vspace{-1.5em}
    {\scriptsize
        \textit{Line style denotes the interface; color denotes the gradient-routing setting.}
    }

    \begin{minipage}[t]{0.32\textwidth}
        \centering
        \includegraphics[
            width=\linewidth
        ]{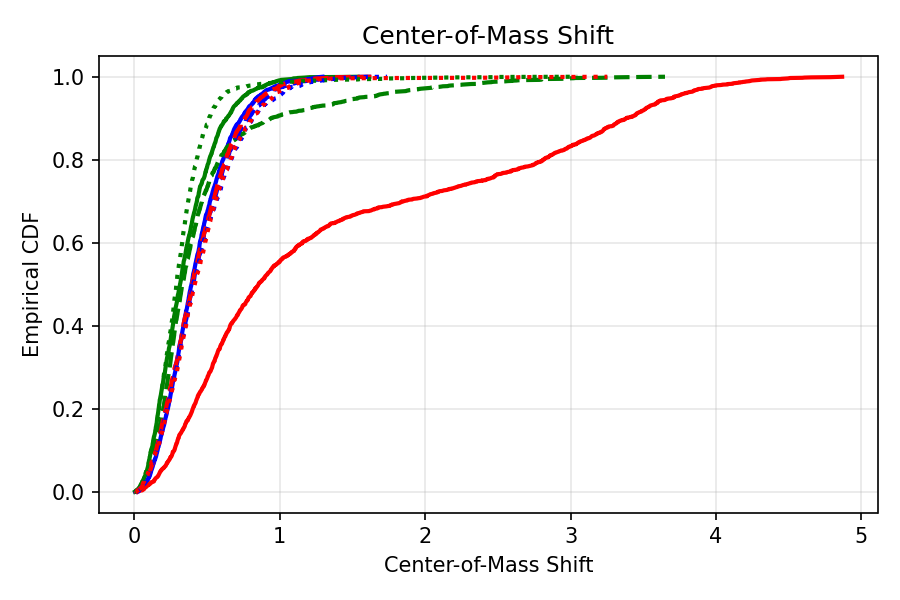}\\[-0.5em]
        {\footnotesize \textbf{(a)} Center shift CDF}
    \end{minipage}
    \hfill
    \begin{minipage}[t]{0.32\textwidth}
        \centering
        \includegraphics[
            width=\linewidth
        ]{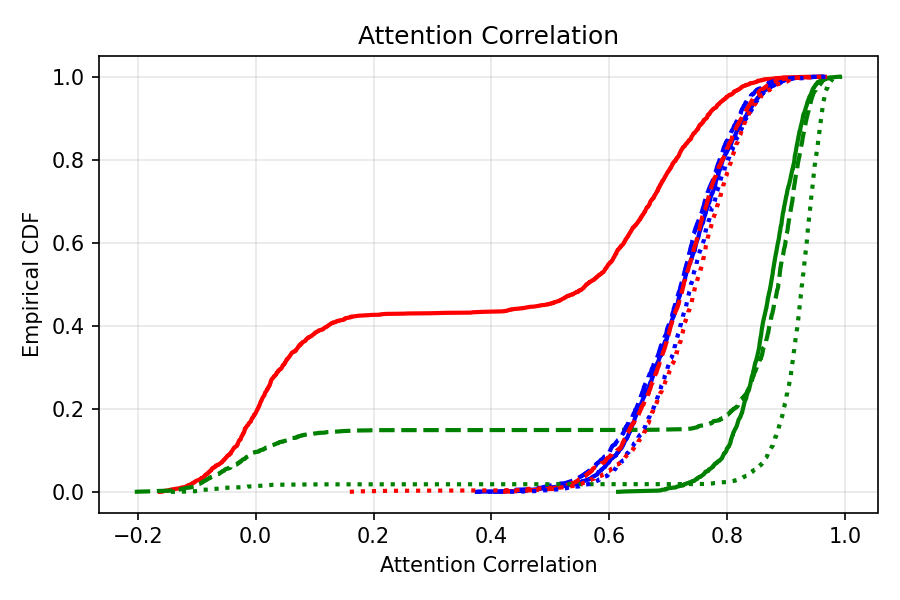}\\[-0.5em]
        {\footnotesize \textbf{(b)} Correlation CDF}
    \end{minipage}
    \hfill
    \begin{minipage}[t]{0.32\textwidth}
        \centering
        \includegraphics[
            width=\linewidth
        ]{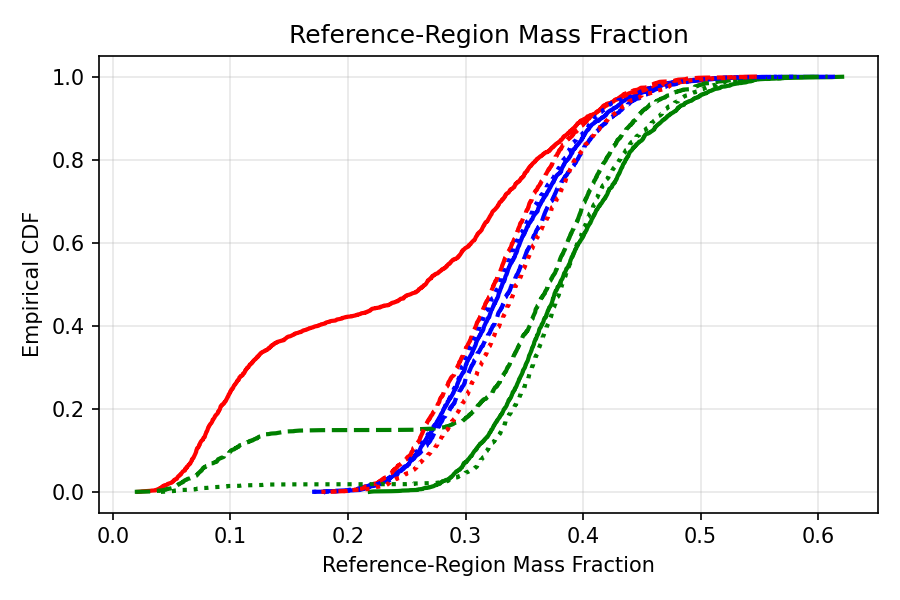}\\[-0.5em]
        {\footnotesize \textbf{(c)} Reference-region mass CDF}
    \end{minipage}

    \begin{minipage}[t]{0.32\textwidth}
        \centering
        \includegraphics[
            width=\linewidth
        ]{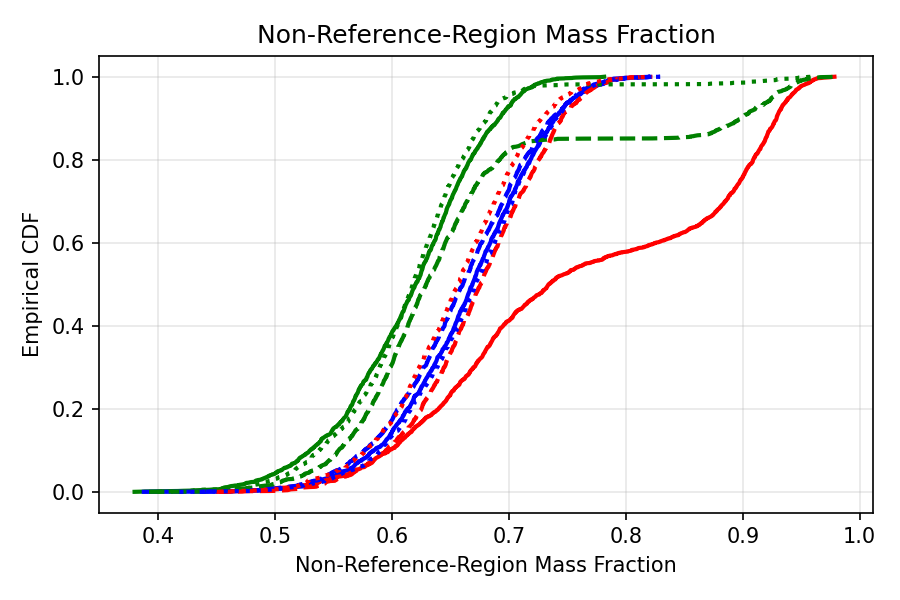}\\[-0.5em]
        {\footnotesize \textbf{(d)} Non-reference-region mass CDF}
    \end{minipage}
    \hfill
    \begin{minipage}[t]{0.32\textwidth}
        \centering
        \includegraphics[
            width=\linewidth
        ]{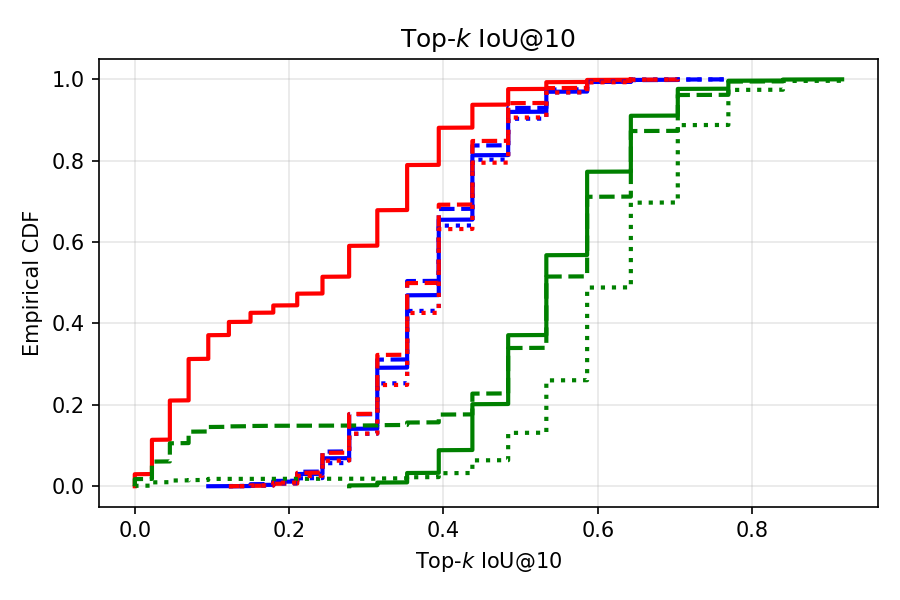}\\[-0.5em]
        {\footnotesize \textbf{(e)} Top-$k$ IoU@10 CDF}
    \end{minipage}
    \hfill
    \begin{minipage}[t]{0.32\textwidth}
        \centering
        \includegraphics[
            width=\linewidth
        ]{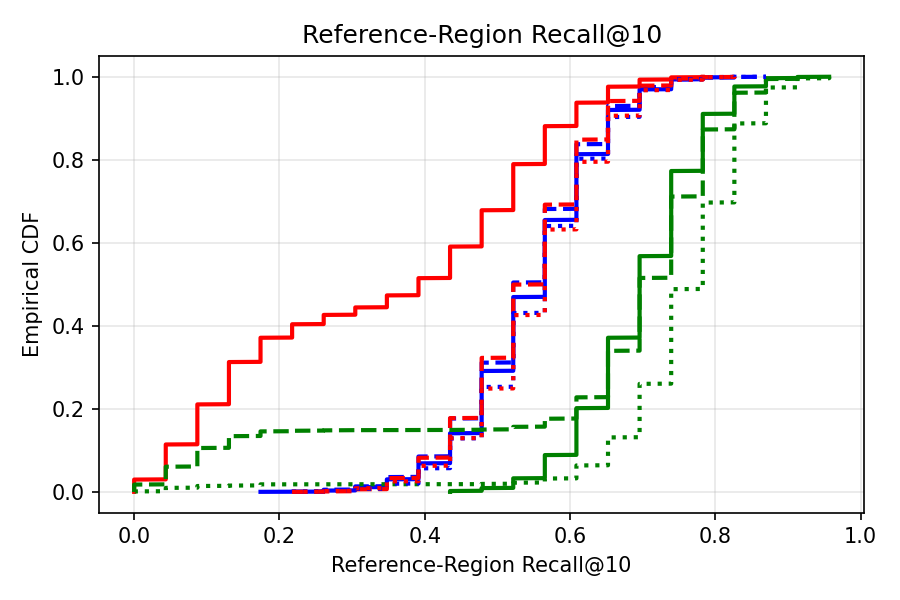}\\[-0.5em]
        {\footnotesize \textbf{(f)} Reference-region recall@10 CDF}
    \end{minipage}

    \caption{
    \textbf{Additional attention-alignment CDF statistics.}
    The CDF plots provide distributional views of attention stability under different gradient-routing comparisons.
    Relative to the direct-fusion baseline, Action QFormer generally exhibits smaller spatial shifts, stronger reference-pattern similarity, higher reference-region mass, lower non-reference-region mass, and stronger overlap with high-response reference regions.
    These statistics provide supporting evidence for the attention-stability trend observed in the joint reference-region-mass and center-shift analysis.
    }
    \label{fig:app_attention_supporting_cdf}
    \vspace{-0.5em}
\end{figure*}

\subsubsection{Additional Attention Statistics}
\label{app:attention_stats}

To complement the joint reference-region-mass and center-shift analysis, we also report auxiliary attention-alignment statistics.
These statistics provide additional views of how closely an action-loss-exposed attention map remains aligned with its action-update-blocked reference.
Specifically, we measure global heatmap correlation, top-region intersection-over-union, top-region recall with respect to the reference, and attention mass outside the reference region, as reported in Figs.~\ref{fig:app_attention_supporting_scatter} and~\ref{fig:app_attention_supporting_cdf}.

\subsubsection{Distributional Attention Statistics}
\label{app:attention_cdf}

Fig.~\ref{fig:app_attention_supporting_cdf} provides complementary distributional views of attention stability across the full set of phrase-level comparisons.
Unlike the joint scatter plots, which expose relationships between pairs of attention metrics, the cumulative distributions summarize how frequently different levels of alignment and spatial drift occur within each interface and gradient-routing setting.

Across the auxiliary metrics, Action QFormer generally shifts the distributions toward smaller center shifts, stronger reference-pattern similarity, higher reference-region mass, lower non-reference-region mass, and greater overlap with high-response regions in the action-update-blocked reference.
These differences are most pronounced in the full-update comparison, consistent with the joint reference-region-mass and center-shift analysis in Fig.~\ref{fig:app_attention_joint_refmass_shift}.
The remaining gradient-routing settings provide context for how the distribution of attention changes varies with action-loss exposure.

\subsubsection{Additional Attention-Focus Examples}
\label{app:additional_attention_focus_cases}

\begin{figure*}[!t]
    \centering

    \begin{minipage}[t]{0.48\textwidth}
        \centering
        \includegraphics[
            width=\linewidth,
            trim=0 205pt 0 70pt,
            clip
        ]{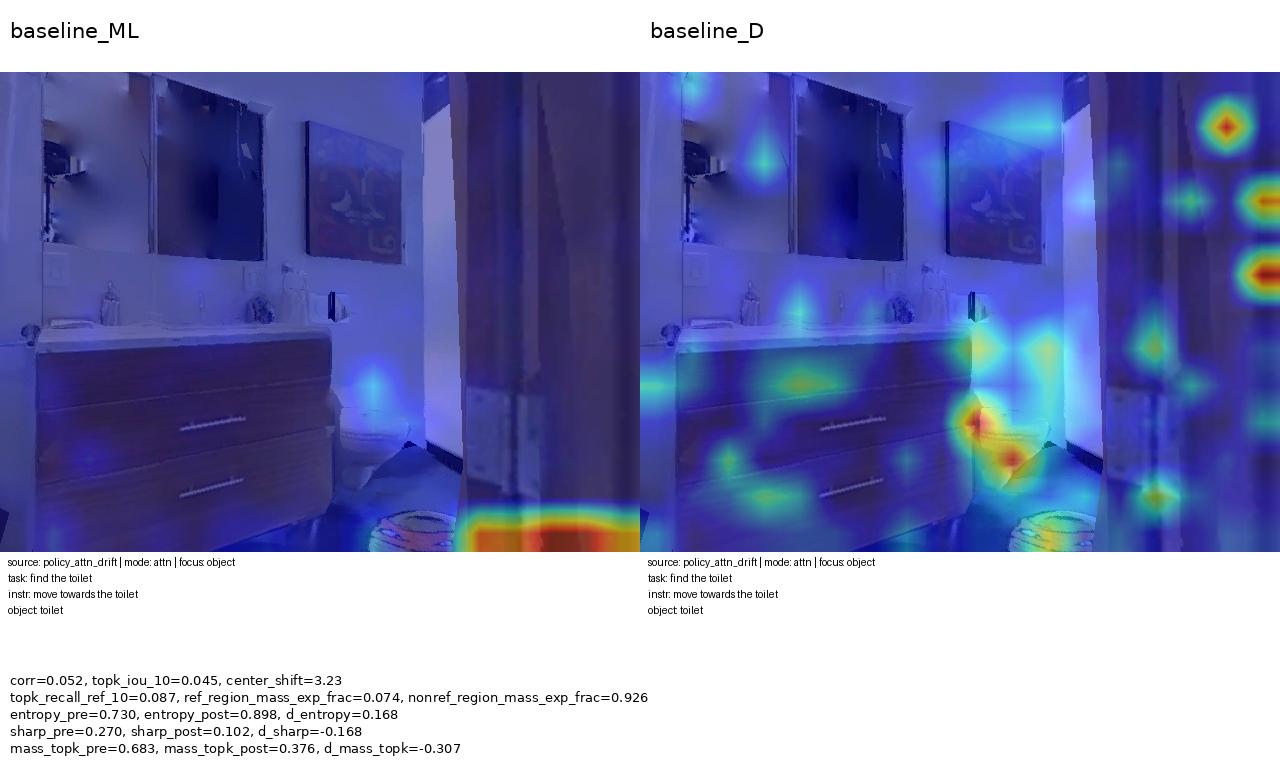}
    \end{minipage}
    \hfill
    \begin{minipage}[t]{0.48\textwidth}
        \centering
        \includegraphics[
            width=\linewidth,
            trim=0 205pt 0 70pt,
            clip
        ]{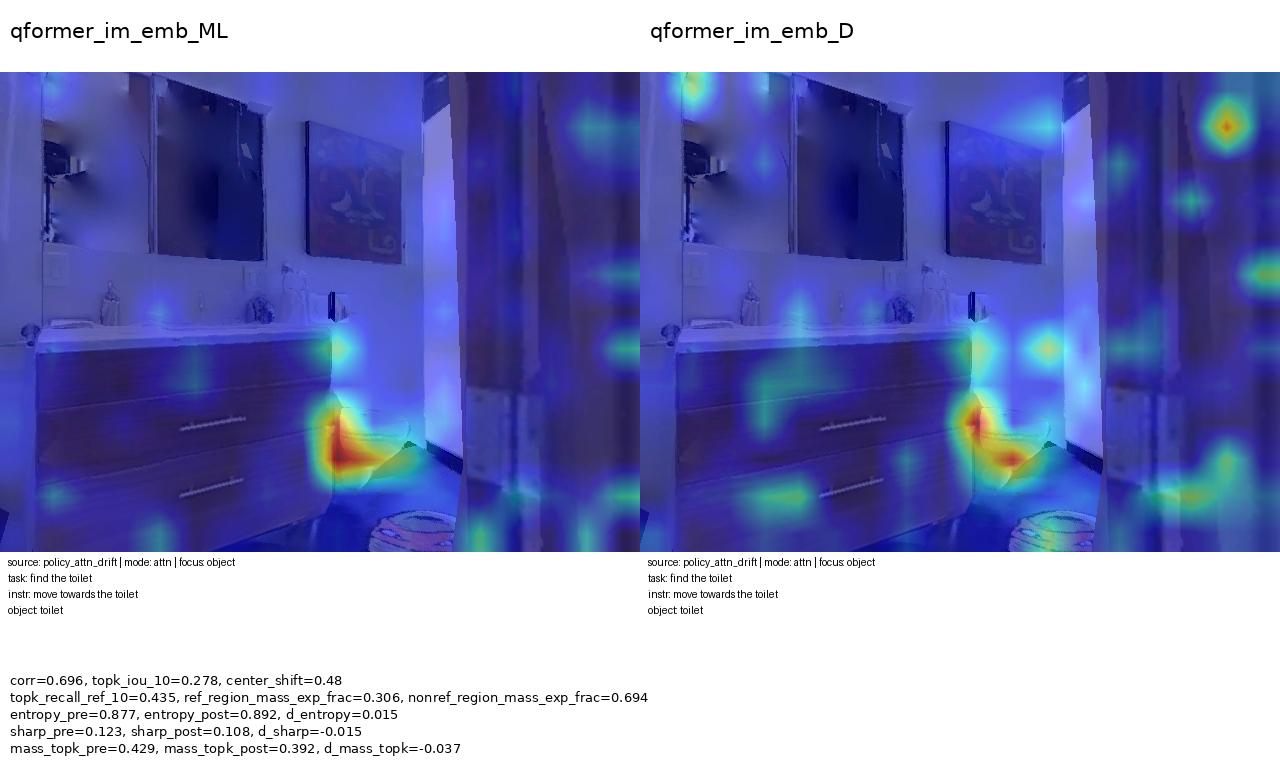}
    \end{minipage}\\[-0.5em]
    {\footnotesize \textbf{(a) Toilet: baseline attention collapse.}}\\
    \vspace{0.3em}

    \begin{minipage}[t]{0.48\textwidth}
        \centering
        \includegraphics[
            width=\linewidth,
            trim=0 205pt 0 70pt,
            clip
        ]{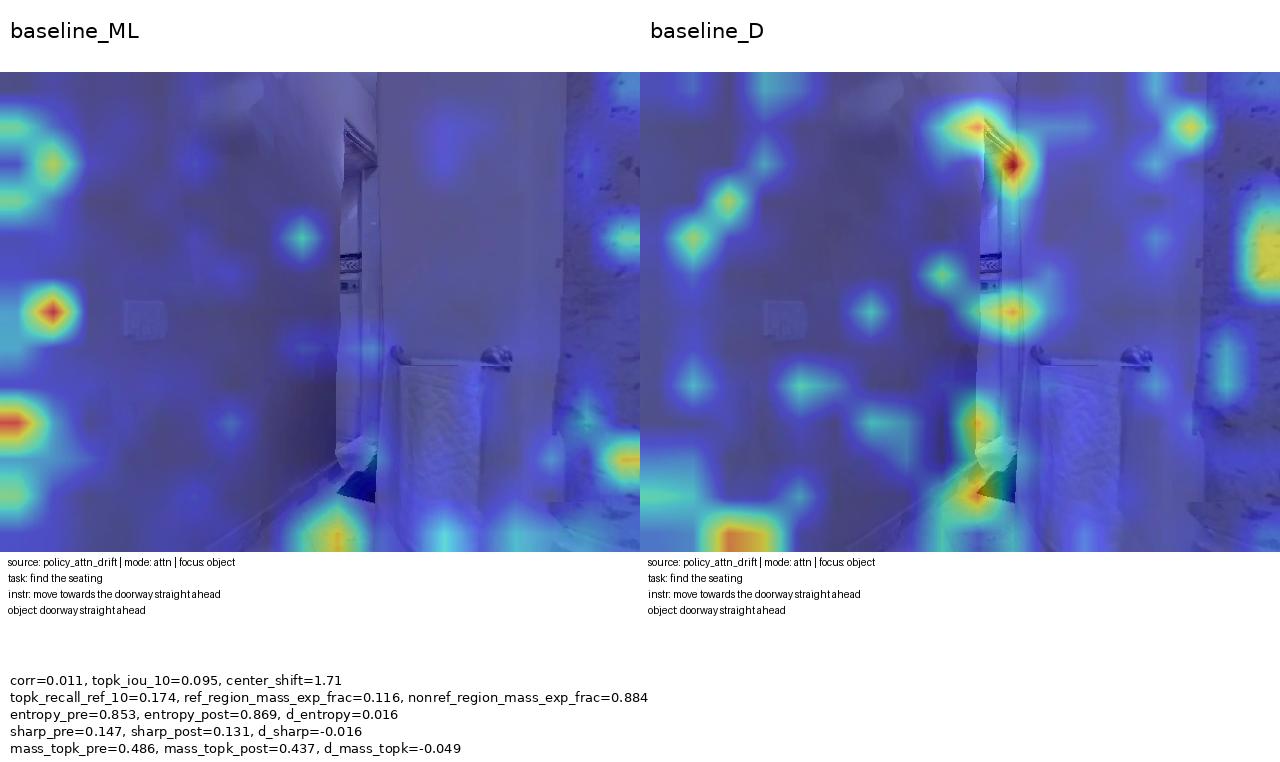}
    \end{minipage}
    \hfill
    \begin{minipage}[t]{0.48\textwidth}
        \centering
        \includegraphics[
            width=\linewidth,
            trim=0 205pt 0 70pt,
            clip
        ]{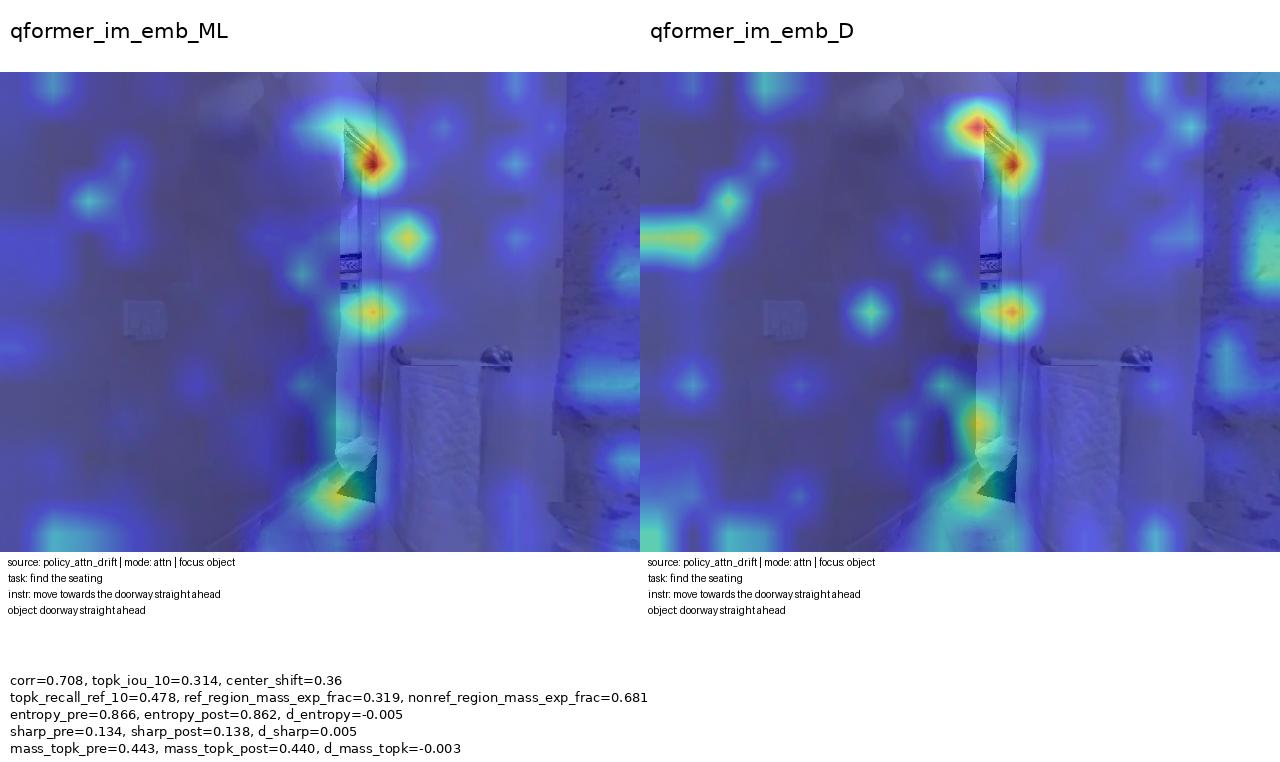}
    \end{minipage}\\[-0.5em]
    {\footnotesize \textbf{(b) Doorway straight ahead: baseline attention drift.}}\\
    \vspace{0.3em}

    \begin{minipage}[t]{0.48\textwidth}
        \centering
        \includegraphics[
            width=\linewidth,
            trim=0 205pt 0 70pt,
            clip
        ]{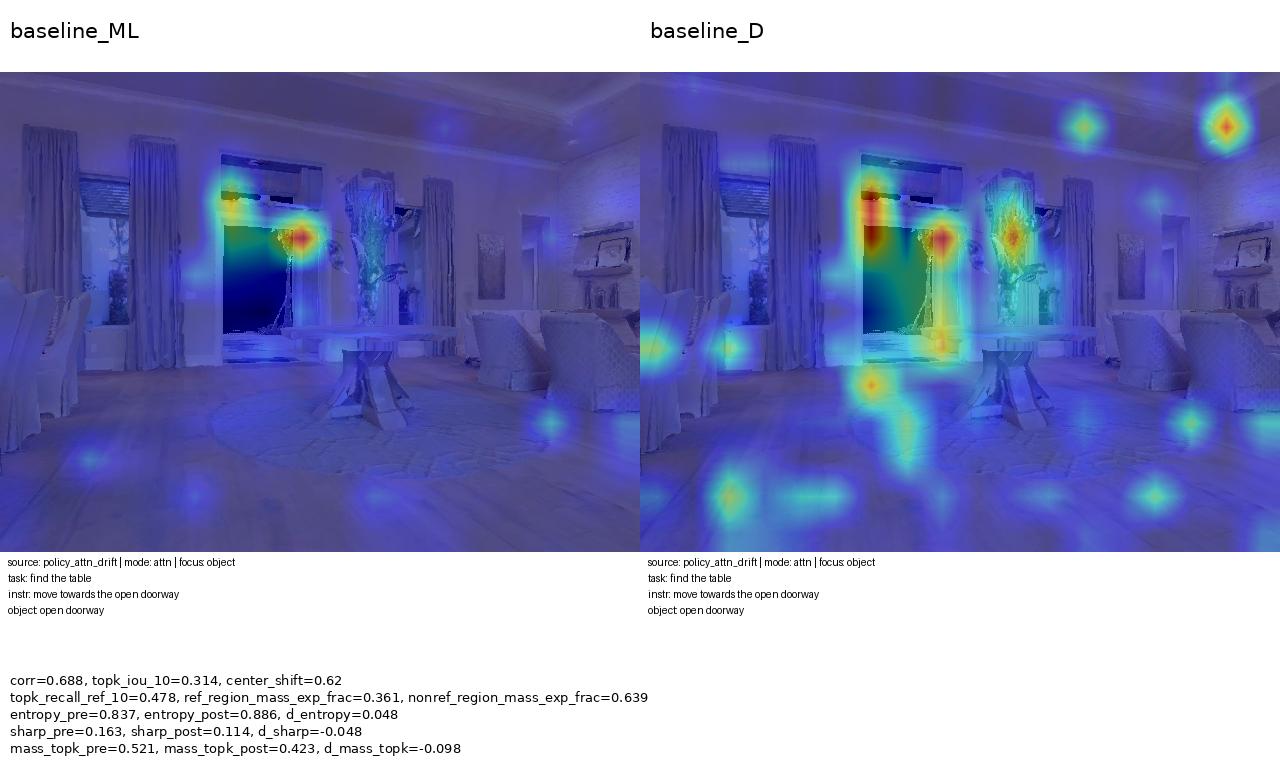}
    \end{minipage}
    \hfill
    \begin{minipage}[t]{0.48\textwidth}
        \centering
        \includegraphics[
            width=\linewidth,
            trim=0 205pt 0 70pt,
            clip
        ]{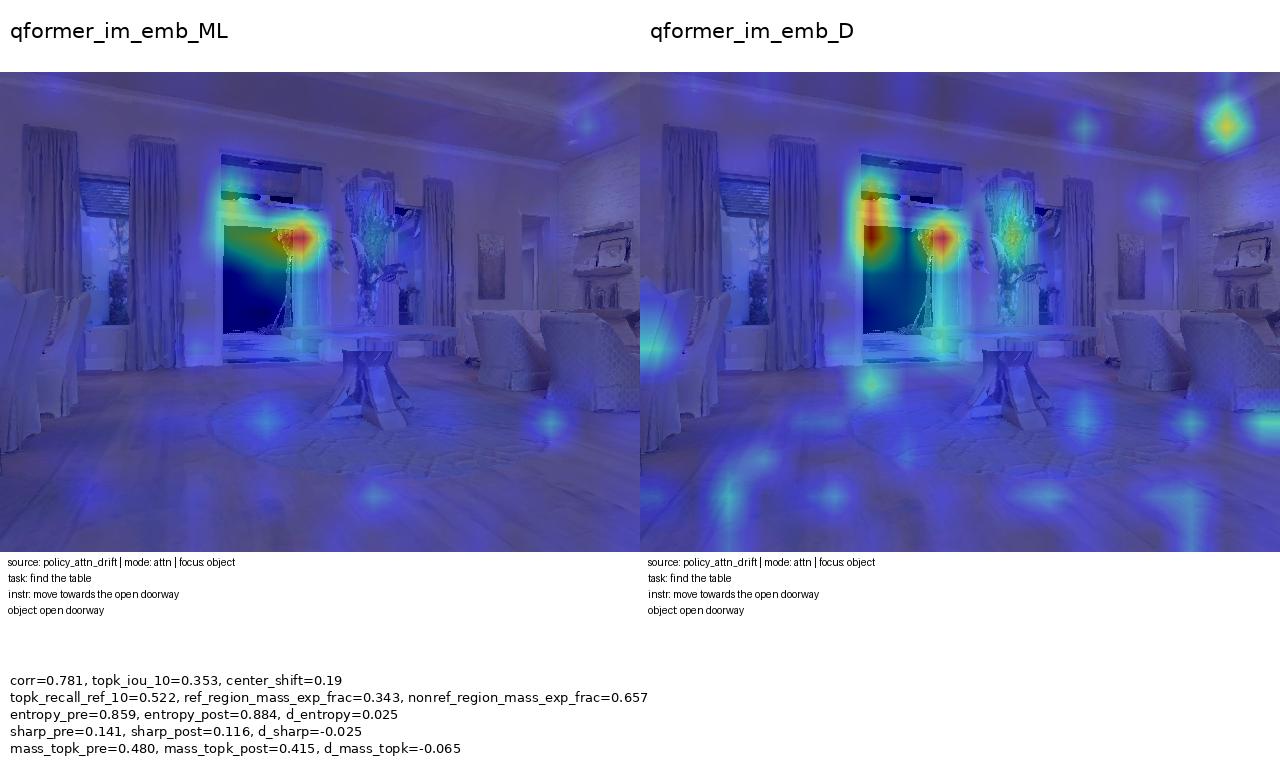}
    \end{minipage}\\[-0.5em]
    {\footnotesize \textbf{(c) Open doorway: partially aligned attention in both interfaces.}}\\

    \vspace{-0.5em}
    \caption{
    \textbf{Additional representative instruction-to-visual attention examples.}
    Each case compares the full-update model with its action-update-blocked reference for the direct-fusion baseline and Action QFormer.
    \textbf{(a)} The toilet case shows severe baseline attention collapse toward irrelevant visual regions, whereas Action QFormer preserves stronger attention around the target object.
    \textbf{(b)} The doorway-straight-ahead case shows baseline attention drifting away from the relevant region, whereas Action QFormer remains more closely aligned with the reference attention pattern.
    \textbf{(c)} The open-doorway case shows that the baseline is not uniformly collapsed:
    both interfaces retain attention toward the relevant doorway region, with Action QFormer exhibiting stronger reference alignment.
    Together, these examples complement the distributional attention-stability results by illustrating several qualitative regimes of attention change.
    }
    \label{fig:app_attention_additional_focus_cases}
    \vspace{-1.0em}
\end{figure*}

We also include representative attention-focus examples in Fig.~\ref{fig:app_attention_additional_focus_cases}.
These examples provide qualitative illustrations that complement the staircase example in the main text.
They cover three regimes:
severe baseline attention collapse,
baseline attention drift away from the relevant region,
and a partially aligned case in which both interfaces retain task-relevant attention.
Because these examples are selectively chosen, they should be interpreted as illustrations of possible attention behaviors rather than as distributional evidence.

\subsubsection{Additional Per-Head Attention-Focus Visualization}
\label{app:additional_per_head_attention_focus}

\begin{figure*}[!t]
    \centering

    \includegraphics[
        width=0.485\textwidth,
        trim=0 205pt 0 70pt,
        clip
    ]{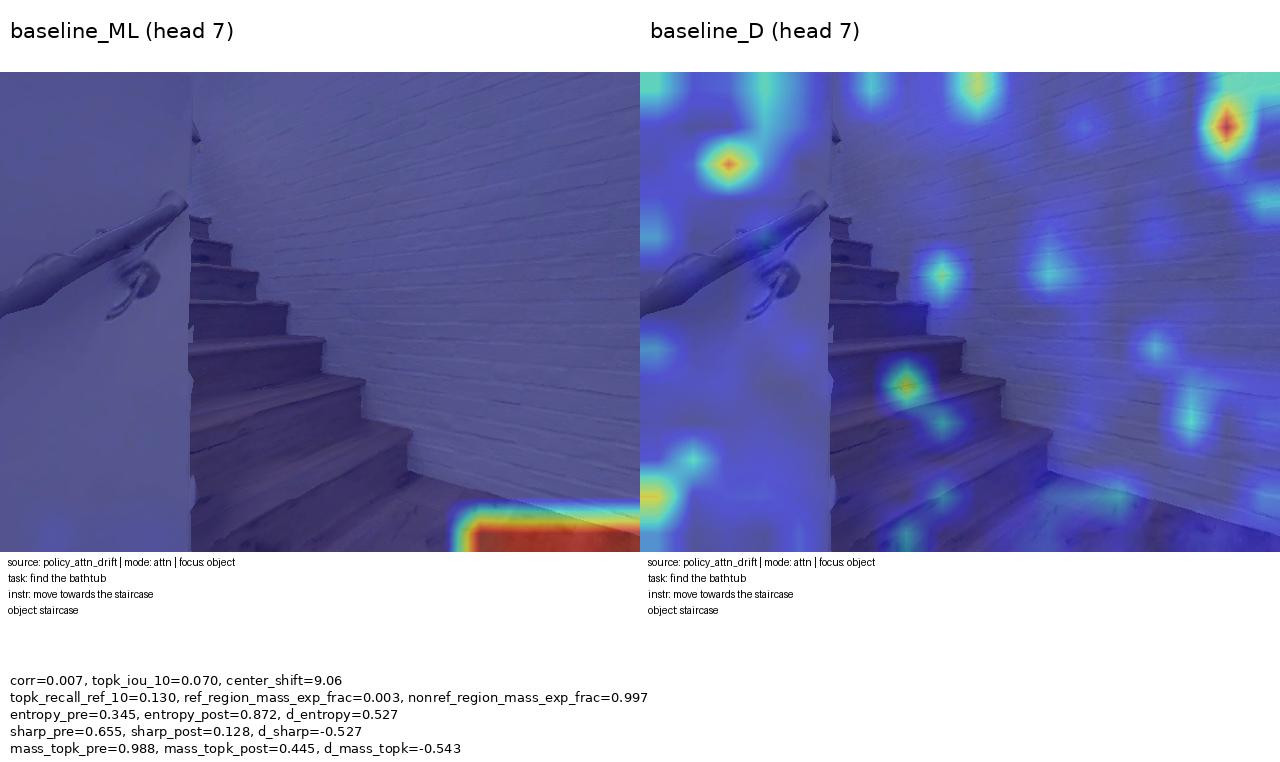}
    \hfill
    \includegraphics[
        width=0.485\textwidth,
        trim=0 205pt 0 70pt,
        clip
    ]{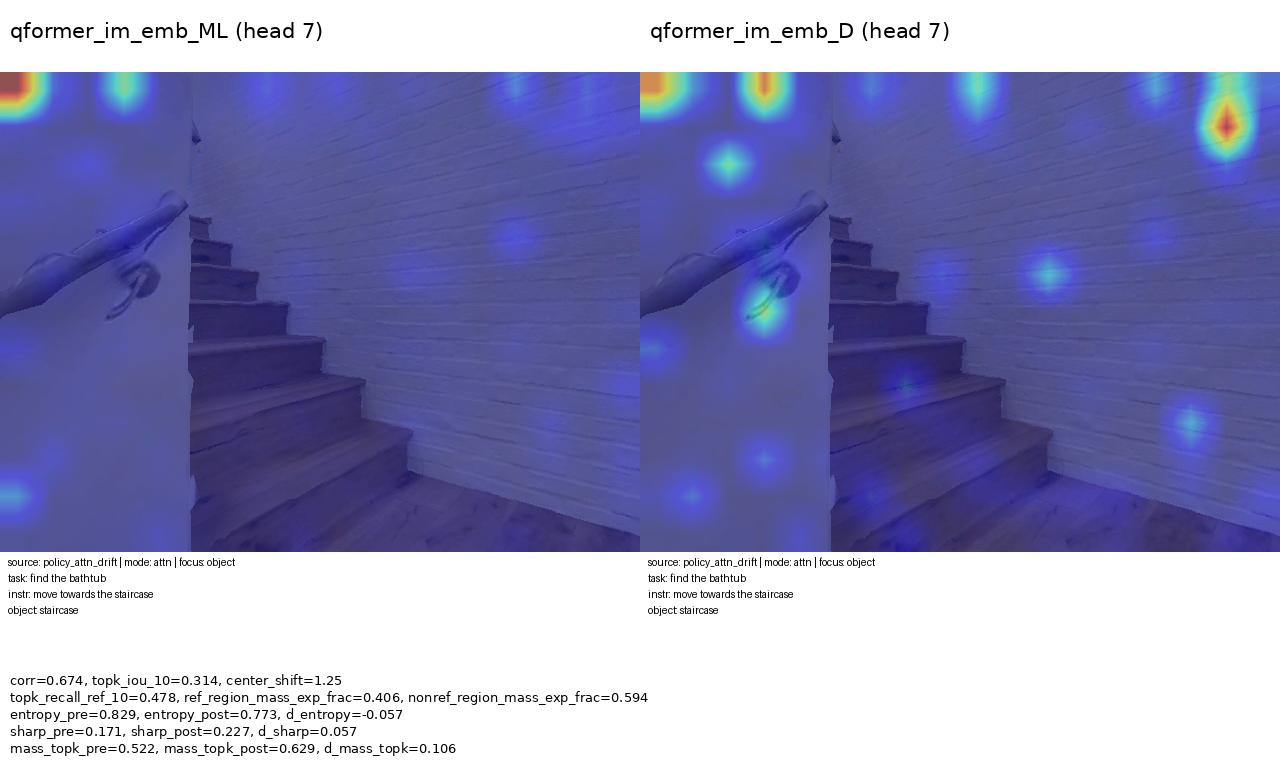}\\[-0.45em]
    {\scriptsize \textbf{(a) Head 7.} Baseline vs.\ Action QFormer}

    \vspace{0.15em}

    \includegraphics[
        width=0.485\textwidth,
        trim=0 205pt 0 70pt,
        clip
    ]{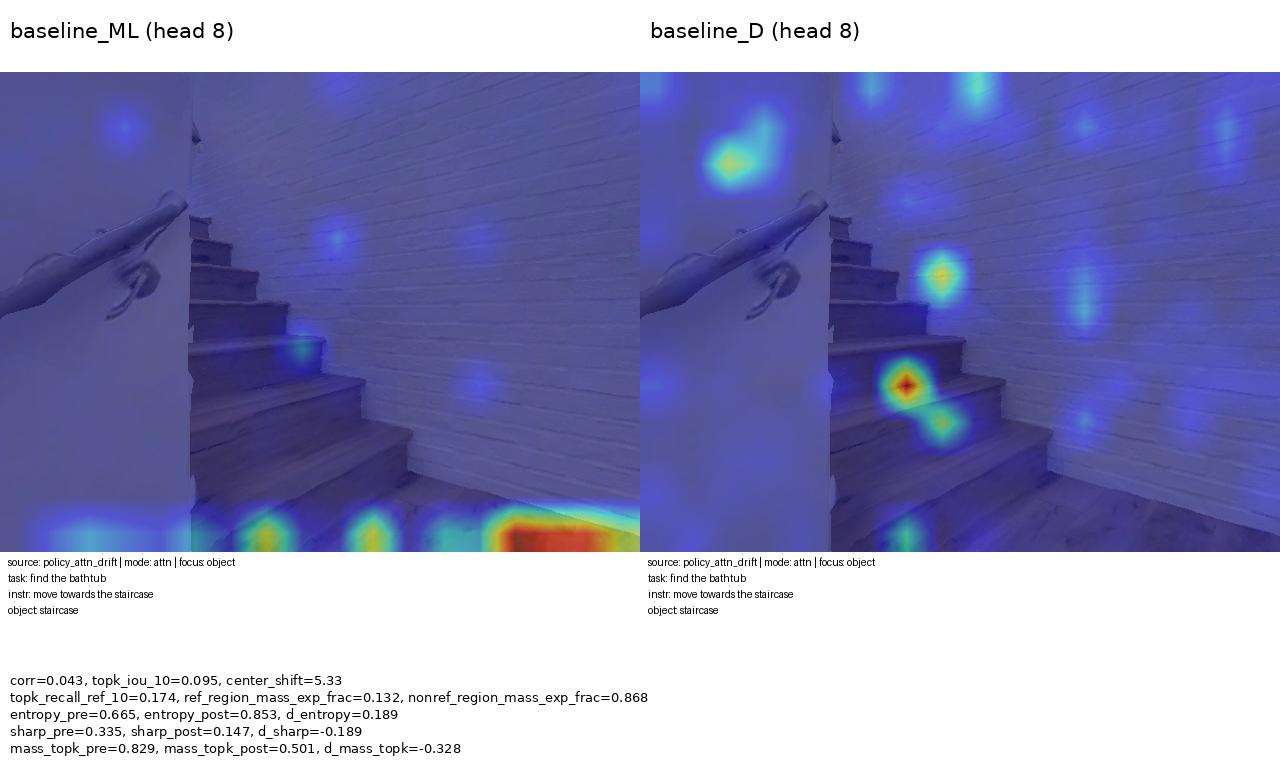}
    \hfill
    \includegraphics[
        width=0.485\textwidth,
        trim=0 205pt 0 70pt,
        clip
    ]{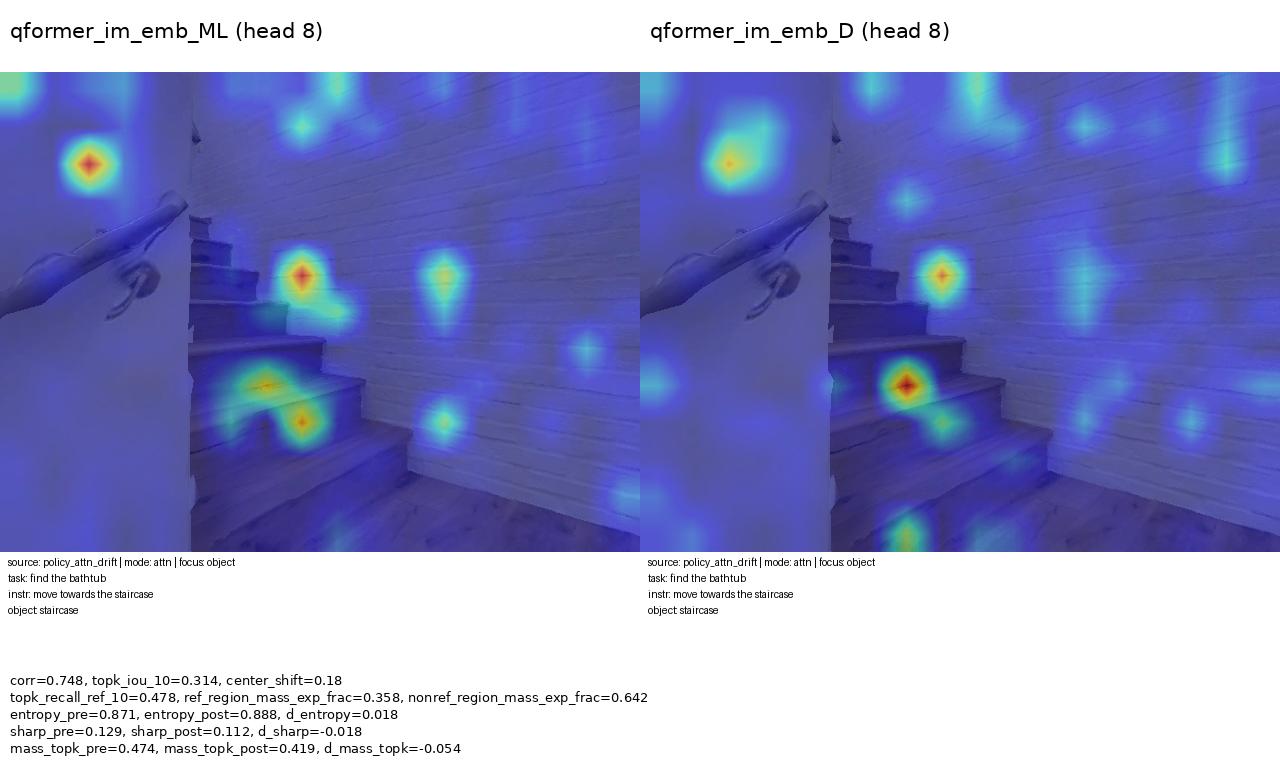}\\[-0.45em]
    {\scriptsize \textbf{(b) Head 8.} Baseline vs.\ Action QFormer}

    \vspace{0.15em}

    \includegraphics[
        width=0.485\textwidth,
        trim=0 205pt 0 70pt,
        clip
    ]{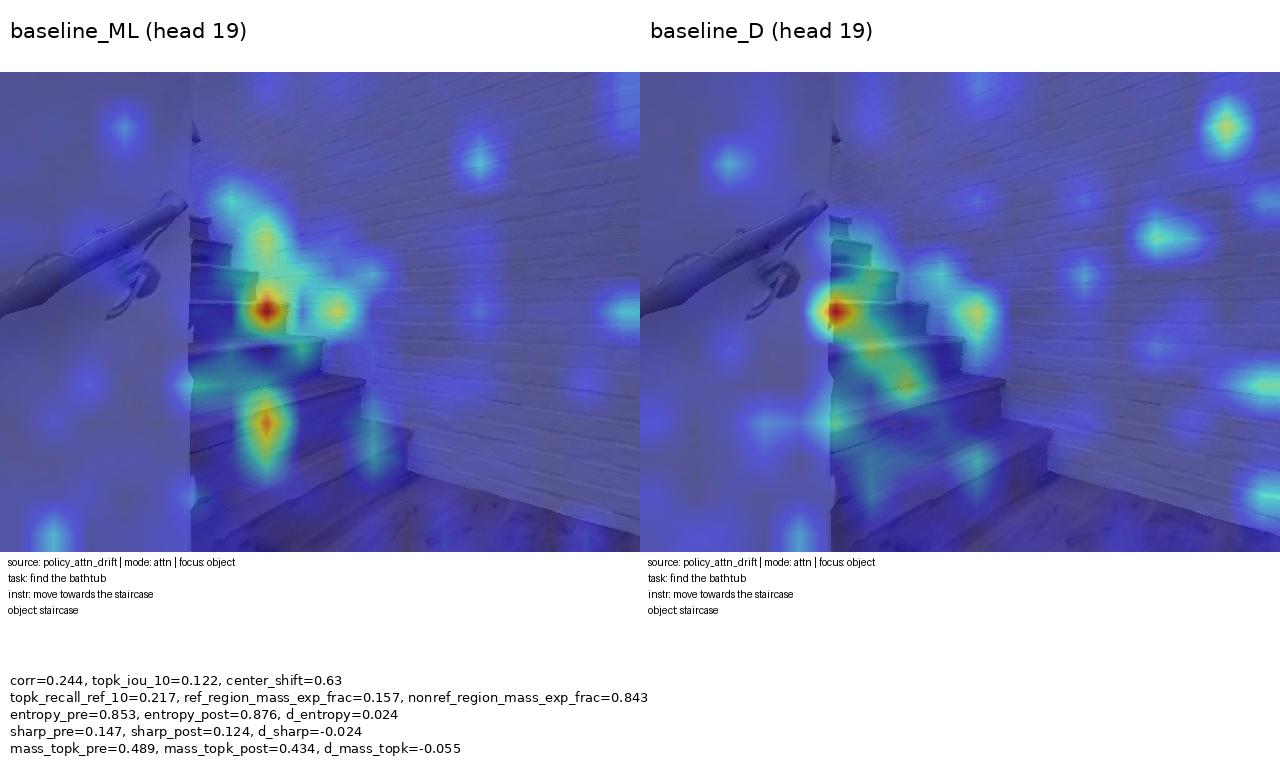}
    \hfill
    \includegraphics[
        width=0.485\textwidth,
        trim=0 205pt 0 70pt,
        clip
    ]{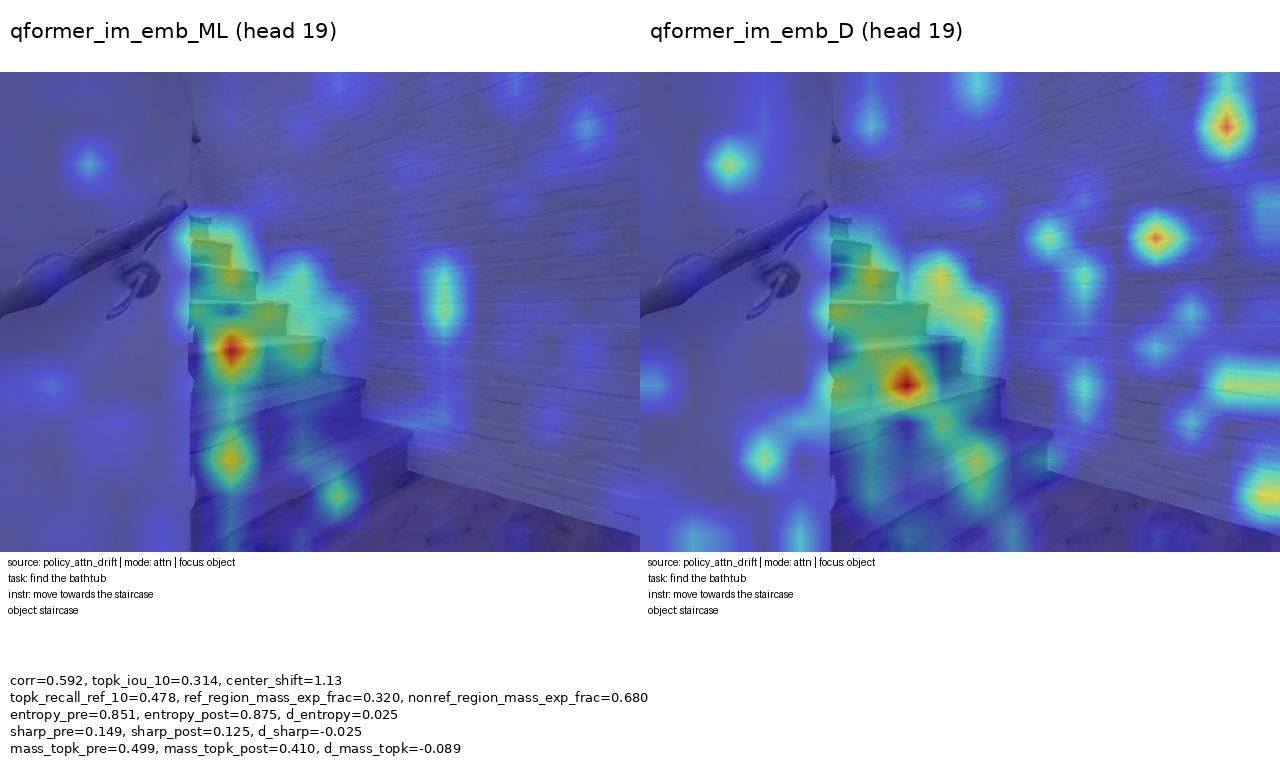}\\[-0.45em]
    {\scriptsize \textbf{(c) Head 19.} Baseline vs.\ Action QFormer}

    \vspace{0.15em}

    \includegraphics[
        width=0.485\textwidth,
        trim=0 205pt 0 70pt,
        clip
    ]{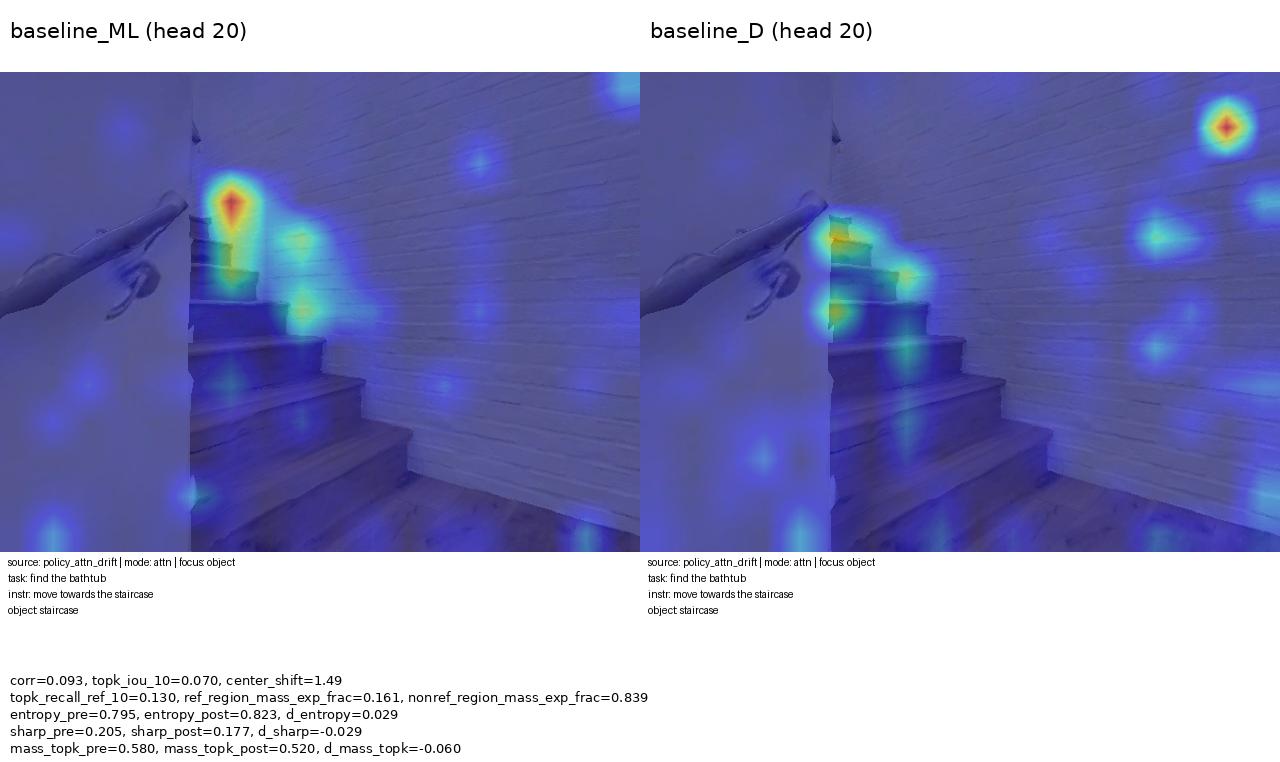}
    \hfill
    \includegraphics[
        width=0.485\textwidth,
        trim=0 205pt 0 70pt,
        clip
    ]{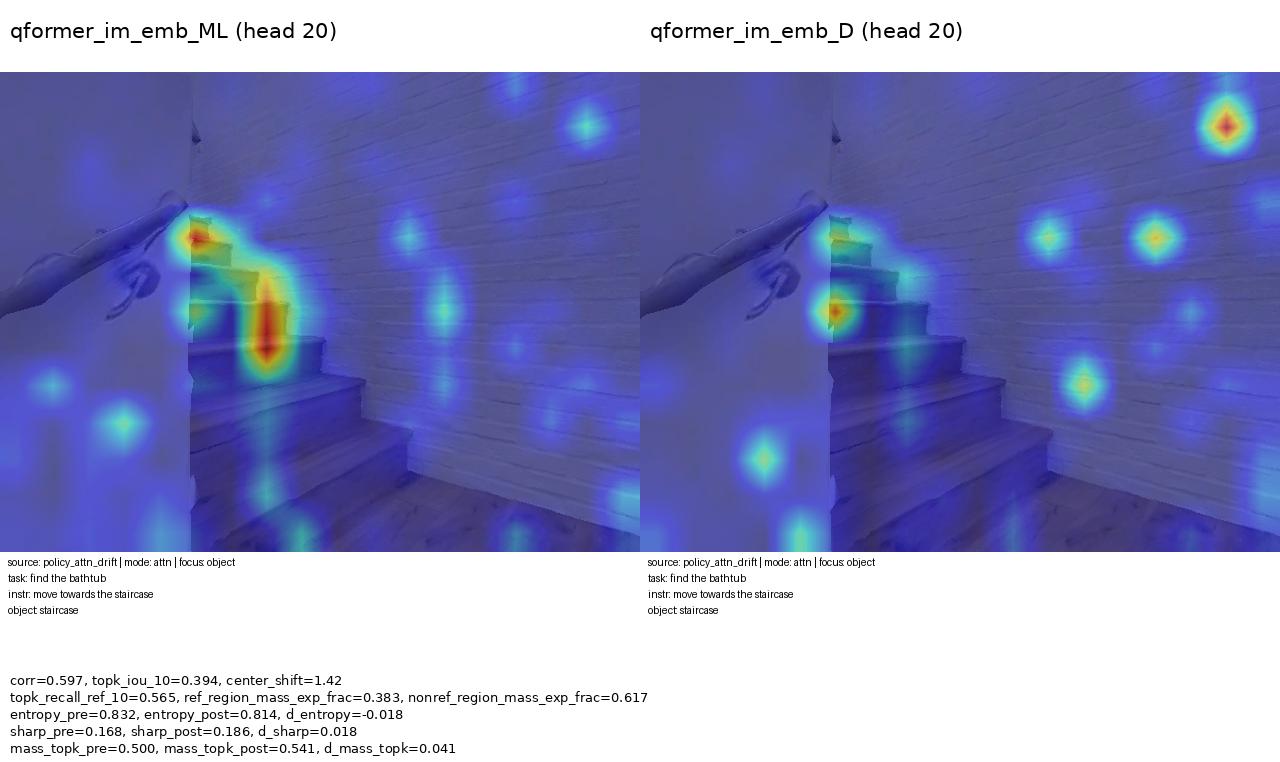}\\[-0.45em]
    {\scriptsize \textbf{(d) Head 20.} Baseline vs.\ Action QFormer}

    \vspace{-0.5em}
    \caption{
    \textbf{Selected per-head instruction-to-visual attention changes for the staircase case in Fig.~\ref{fig:attention_focus_main}.}
    Each row shows one attention head for the instruction \emph{move towards the staircase}; direct-fusion baseline results are shown on the left and Action QFormer results on the right.
    Within each panel, the full-update model is compared with its action-update-blocked reference.
    Heads 7 and 8 show severe baseline attention shifts toward irrelevant visual-token regions, whereas Action QFormer retains more stable target-oriented attention.
    Heads 19 and 20 show that the baseline is not uniformly collapsed, as selected heads in both interfaces continue to concentrate attention toward the target object.
    }
    \label{fig:app_attention_per_head}
    \vspace{-1.0em}
\end{figure*}

The per-head visualization in Fig.~\ref{fig:app_attention_per_head} shows that the all-head average can hide substantial head-level heterogeneity.
Some baseline heads exhibit constructive adaptation by concentrating attention near the target object, whereas other heads shift toward irrelevant visual-token regions.
For the selected heads exhibiting severe baseline shifts, Action QFormer maintains a more stable target-oriented attention pattern.
These examples therefore illustrate both that the baseline is not uniformly collapsed across all heads and that Action QFormer can reduce severe head-level shifts while preserving task-relevant attention.
We treat this visualization as qualitative evidence of head-level heterogeneity rather than as a general claim about specialization across all attention heads.

\end{document}